%% file: arxiv.tex
\documentclass[10pt,twocolumn,letterpaper]{article}

\usepackage{etoolbox}
\newtoggle{main}
\settoggle{main}{true}

\usepackage[pagenumbers]{cvpr}

\usepackage[table,dvipsnames]{xcolor}
\usepackage{tikz,pgfplots}
\usepackage{array}
\usepackage{arrayjobx}
\usepackage{graphicx}
\usepackage{caption}
\usepackage{subcaption}
\usepackage{booktabs}
\usepackage{multirow}
\usepackage{gensymb}
\usepackage{pifont}
\usepackage{enumitem}
\usepackage{xspace}
\usepackage{adjustbox}
\usepackage{makecell}
\usepackage[normalem]{ulem}
\usepackage{wrapfig}
\usepackage{appendix}
\usepackage{etoc}
\usepackage[accsupp]{axessibility}

\usepackage{epigraph}

\input{tex/math_commands}

\definecolor{cvprblue}{rgb}{0.21,0.49,0.74}
\usepackage[pagebackref,breaklinks,colorlinks,citecolor=cvprblue]{hyperref}

\begin{document}

\title{\textbf{DUNE}: Distilling a Universal Encoder from Heterogeneous 2D and 3D Teachers}

\author{
\renewcommand{\arraystretch}{1.2}
\begin{tabular}{
  >{\centering\arraybackslash}p{4cm}
  >{\centering\arraybackslash}p{4cm}
  >{\centering\arraybackslash}p{4cm}
}
Mert Bülent Sarıyıldız & Philippe Weinzaepfel & Thomas Lucas \\
Pau de Jorge & Diane Larlus & Yannis Kalantidis \\
\multicolumn{3}{c}{NAVER LABS Europe} \\
\multicolumn{3}{c}{\url{https://europe.naverlabs.com/dune}} \\
\end{tabular}
}

\maketitle

\etocdepthtag.toc{mtchapter}
\etocsettagdepth{mtappendix}{none}

\input{tex/abbrev}

\input{tex/00_abstract}
\input{tex/01_introduction}
\input{tex/02_related}
\input{tex/03_htd}
\input{tex/04_method}
\input{tex/05_experiments}

\input{tex/07_conclusions}

\myparagraph{Acknowledgements.}
We thank the 3DH and GeoDe teams at NAVER LABS Europe who created \hmr and \master, respectively, for sharing their dataset, training and evaluation pipelines.
We also thank César de Souza for developing a real-time multi-task demo interface for \ours, as well as for naming our model.

{\small
\bibliographystyle{ieeenat_fullname}
\bibliography{main}
}

\clearpage
\newpage
\appendix

\etocdepthtag.toc{mtappendix}
\etocsettagdepth{mtappendix}{subsection}
\etocsettagdepth{mtchapter}{none}

{
    \hypersetup{linkcolor=black}
    \twocolumn[
    \begin{center}
        \tableofcontents
        \vspace{0.5cm}
    \end{center}
    ]
}

\input{tex/99_supp}

\end{document}

%% file: tex/math_commands.tex
\usepackage{amsmath,amsfonts,bm}

\def\eqref#1{equation~\ref{#1}}

\def\1{\bm{1}}

\def\vs{{\bm{s}}}
\def\vt{{\bm{t}}}

\DeclareMathAlphabet{\mathsfit}{\encodingdefault}{\sfdefault}{m}{sl}
\SetMathAlphabet{\mathsfit}{bold}{\encodingdefault}{\sfdefault}{bx}{n}

\def\gD{{\mathcal{D}}}

\def\gI{{\mathcal{I}}}

\def\gL{{\mathcal{L}}}

\def\gT{{\mathcal{T}}}

\def\gZ{{\mathcal{Z}}}

%% file: tex/abbrev.tex
\newcounter{rownumbers}
\newcounter{rownumbersft}
\renewcommand\rownum{{\color{darkgray} \stepcounter{rownumbers}\arabic{rownumbers}.}}
\newcommand\rownumft{{\color{darkgray} \stepcounter{rownumbersft}\arabic{rownumbersft}.}}

\crefname{section}{Sec.}{Secs.}
\Crefname{section}{Sec.}{Secs.}
\Crefname{table}{Tab.}{Tabs.}
\crefname{table}{Tab.}{Tabs.}
\Crefname{figure}{Fig.}{Figs.}
\crefname{figure}{Fig.}{Figs.}
\Crefname{equation}{Eq.}{Eqs.}
\crefname{equation}{Eq.}{Eqs.}
\crefname{appendix}{Appendix}{Appendices}

\makeatletter
\newcommand{\CapitalizeFirst}[1]{
  \expandafter\@CapitalizeFirst\expandafter{#1}
}
\def\@CapitalizeFirst#1{\MakeUppercase{\@car#1\@nil}\@cdr#1\@nil}
\makeatother

\newcommand{\ours}{DUNE\xspace}
\newcommand{\fbest}[1]{{\color{YellowOrange}{#1}}}

\newcommand{\ltnorm}[1]{\left\|#1\right\|_2}

\newcommand{\tcell}[2]{\begin{tabular}[c]{@{}c@{}} #1 \\ #2 \end{tabular}}

\newcommand{\tone}{\cT}

\newcommand{\done}[1]{{\color{OliveGreen}{#1}}}
\newcommand{\running}[1]{{\color{orange}{#1}}}
\newcommand{\torun}[1]{{\color{BrickRed}{#1}}}
\newcommand{\queued}[1]{{\color{Dandelion}{#1}}}

\makeatletter
\DeclareRobustCommand\onedot{\futurelet\@let@token\@onedot}
\def\@onedot{\ifx\@let@token.\else.\null\fi\xspace}
\def\eg{\emph{e.g}\onedot} \def\Eg{\emph{E.g}\onedot}
\def\ie{\emph{i.e}\onedot} \def\Ie{\emph{I.e}\onedot}
\def\cf{\emph{cf}\onedot} \def\Cf{\emph{Cf}\onedot}
\def\etc{\emph{etc}\onedot} \def\vs{\emph{vs}\onedot}
\def\wrt{w.r.t\onedot} \def\dof{d.o.f\onedot}
\def\etal{\emph{et al}\onedot}
\def\st{\emph{s.t}\onedot}
\makeatother

\renewcommand{\paragraph}[1]{\vspace{1pt}\noindent\textbf{#1}}
\newcommand{\myparagraph}[1]{\paragraph{#1}}

\definecolor{tabdefault}{gray}{0.8}
\definecolor{lightgray}{gray}{0.93}
\newcommand{\insight}[1]{
\par\noindent\colorbox{lightgray}{
\begin{minipage}{0.97\linewidth}\vspace{1pt} $\blacktriangleright$ \emph{#1}\end{minipage}}
}

\newcommand{\nodotparagraph}[1]{\noindent\textbf{#1}}

\newcommand{\cmark}{\ding{52}}
\newcommand{\xmark}{\ding{55}}

\newcommand{\supp}{supplementary material\xspace}

\newcommand{\dino}{DINO-v2\xspace}
\newcommand{\hmr}{Multi-HMR\xspace}
\newcommand{\master}{MASt3R\xspace}
\newcommand{\amradio}{AM-RADIO-v2.5\xspace}

\newcommand{\tdrop}{{\em tdrop}\xspace}
\newcommand{\dpshort}{LP\xspace}
\newcommand{\dplong}{ladder of projectors\xspace}
\newcommand{\Dplong}{Ladder of Projectors\xspace}

\definecolor{tabsecond}{rgb}{0.8, 1, 0.8}
\definecolor{tabthird}{rgb}{0.88, 1, 0.88}
\definecolor{tabfirst}{rgb}{0.5, 1, 0.5}
\definecolor{tabtop}{rgb}{0.2, 1, 0.2}
\definecolor{tablast}{rgb}{1, 0.5, 0.5}
\definecolor{tablastest}{rgb}{1, 0.2, 0.2}
\definecolor{tabsecondlast}{rgb}{1, 0.8, 0.8}

\newcommand{\gradinggrad}{35}
\newcommand{\ok}[1]{{\cellcolor{tabthird!\gradinggrad}{#1}}}
\newcommand{\good}[1]{{\cellcolor{tabsecond!\gradinggrad}{#1}}}
\newcommand{\better}[1]{{\cellcolor{tabfirst!\gradinggrad}{#1}}}
\newcommand{\best}[1]{{\cellcolor{tabtop!\gradinggrad}{#1}}}
\newcommand{\degradesmost}[1]{{\cellcolor{tablastest!\gradinggrad}{#1}}}
\newcommand{\degrades}[1]{{\cellcolor{tablast!\gradinggrad}{#1}}}
\newcommand{\degradesless}[1]{{\cellcolor{tabsecondlast!\gradinggrad}{#1}}}

\newcommand{\bad}[1]{{\color{black!30!BrickRed}{#1}}}

\newcommand{\gain}[1]{\textbf{\color{OliveGreen}{{$\uparrow$}{#1}}}}
\newcommand{\loss}[1]{\textbf{\color{BrickRed}{{$\downarrow$}{#1}}}}

\definecolor{teal}{RGB}{41,120,108}

\abovedisplayskip=3pt
\belowdisplayskip=3pt

%% file: tex/00_abstract.tex
\begin{abstract}
\noindent Recent multi-teacher distillation methods have unified the encoders of multiple foundation models into a single encoder, achieving competitive performance on core vision tasks like classification, segmentation, and depth estimation.
This led us to ask: Could similar success be achieved when the pool of teachers also includes vision models specialized in diverse tasks across both 2D and 3D perception?
In this paper, we define and investigate the problem of \textit{heterogeneous teacher distillation}, or co-distillation—a challenging multi-teacher distillation scenario where teacher models vary significantly in both (a) their design objectives and (b) the data they were trained on.
We explore data-sharing strategies and teacher-specific encoding, and introduce \textbf{\ours}, a single encoder excelling in 2D vision, 3D understanding, and 3D human perception.
Our model achieves performance comparable to that of its larger teachers, sometimes even outperforming them, on their respective tasks.
Notably, \ours surpasses MASt3R in Map-free Visual Relocalization with a much smaller encoder.
\end{abstract}

%% file: tex/01_introduction.tex
\section{Introduction}
\label{sec:introduction}

Computer vision has seen the rise of foundation models~\cite{bommasani2021opportunities} such as \dino~\cite{oquab2024dinov2} or SAM~\cite{wang2023sam}.
Trained on massive web-crawled datasets, they provide representations useful for multiple downstream tasks.
A recent body of work including AM-RADIO~\cite{ranzinger2024radio}, Theia~\cite{shang2024theia} or UNIC~\cite{sariyildiz2024unic} has
successfully unified
the encoders of several of these foundation models into a single compact encoder via \textit{multi-teacher distillation}.
Although an impressive feat,
this line of research has so far only
distilled models all trained on data of a similar nature, composed of ``generic'' web-crawled images. In fact,
in all recent works~\cite{ranzinger2024radio, sariyildiz2024unic, lu2024swiss}, distilling using the ImageNet-1K~\cite{deng2009imagenet} dataset suffices to
match teacher performance for tasks like classification, segmentation and monocular depth.

\newpage
In this work we study \textit{heterogeneous teacher distillation}, or \textit{co-distillation}.\footnote{In chemistry, co-distillation refers to distillation performed on mixtures in which the compounds are not miscible.}
We consider a set of teacher models as \textit{heterogeneous} if
they vary
significantly with respect to both (a) the \textit{tasks} these teachers are trained for
and (b) the visual domains of their training data.
We seek to answer the following question:
\textit{Can we train
a single encoder that excels at widely diverse tasks by distilling from
state-of-the-art heterogeneous models?}

\input{tex/float/fig_teaser}

To that end, we distinguish between task-agnostic teachers, aimed at producing representations that generalize across several tasks, and specialized teachers that achieve state-of-the-art performance on one specific task.
Given the heterogeneity in training data across teachers, we examine which data should be used for distillation.
Finally, we study the use of projectors,
modules used in multi-teacher distillation to ensure compatibility across teachers and capture teacher-specific information. More specifically we question the projector design
when distilling
heterogeneous teachers with abundant specialized information.

\looseness=-1
To evaluate our framework, we
choose
three state-of-the-art models.
Two are highly task-specific: \master~\cite{mast3r} solves 3D scene reconstruction and matching, while \hmr~\cite{multihmr} solves 3D human perception.
As a third teacher, we add \dino~\cite{oquab2024dinov2}, a popular visual foundation model that generalizes to various visual downstream tasks {including} semantic segmentation or monocular depth estimation.
This leads to \textbf{\ours}, obtained by \textbf{D}istilling a \textbf{UN}iversal \textbf{E}ncoder from heterogeneous 2D and 3D Teachers.

\input{tex/float/fig_pca_mini}

The selected teachers vary with respect to both the tasks they are tackling and the datasets they were trained on.
Their heterogeneity is clearly visible when inspecting the patch features they produce.
In~\cref{fig:pca_mini}, we plot the top three components (obtained via principal component analysis)
for the features of each teacher, as well as
the features of our co-distilled encoder, \ours.
We observe that each teacher has distinct and complementary features.
We also see that our model captures properties present across all three teachers.

Our study yields several insights into the selection of distillation data and the impact of projector design for co-distillation.
More importantly, it results in a powerful universal encoder that matches top models in binocular 3D reconstruction, 3D human perception, and classical 2D vision tasks while retaining much of \dino's generalization performance.
Additionally, it sets a new state-of-the-art on the Map-free Visual Relocalization Challenge\footnote{\url{https://research.nianticlabs.com/mapfree-reloc-benchmark/leaderboard}}  using a ViT-Base encoder.

\paragraph{Contributions.}
{(a)} We define the problem of heterogeneous teacher distillation, where a single model is distilled from teacher models that vary significantly in training tasks and image domains.
{(b)} We investigate suitable distillation strategies for this setting, focusing on distillation data and projector design.
{(c)} We introduce \textbf{DUNE}, a strong ViT-Base encoder capable of excelling in 3D scene understanding, 3D human perception, and 2D vision tasks.

%% file: tex/float/fig_teaser.tex
\begin{figure}[t]
\centering
\adjustbox{max width=.95\linewidth}{
\includegraphics[]{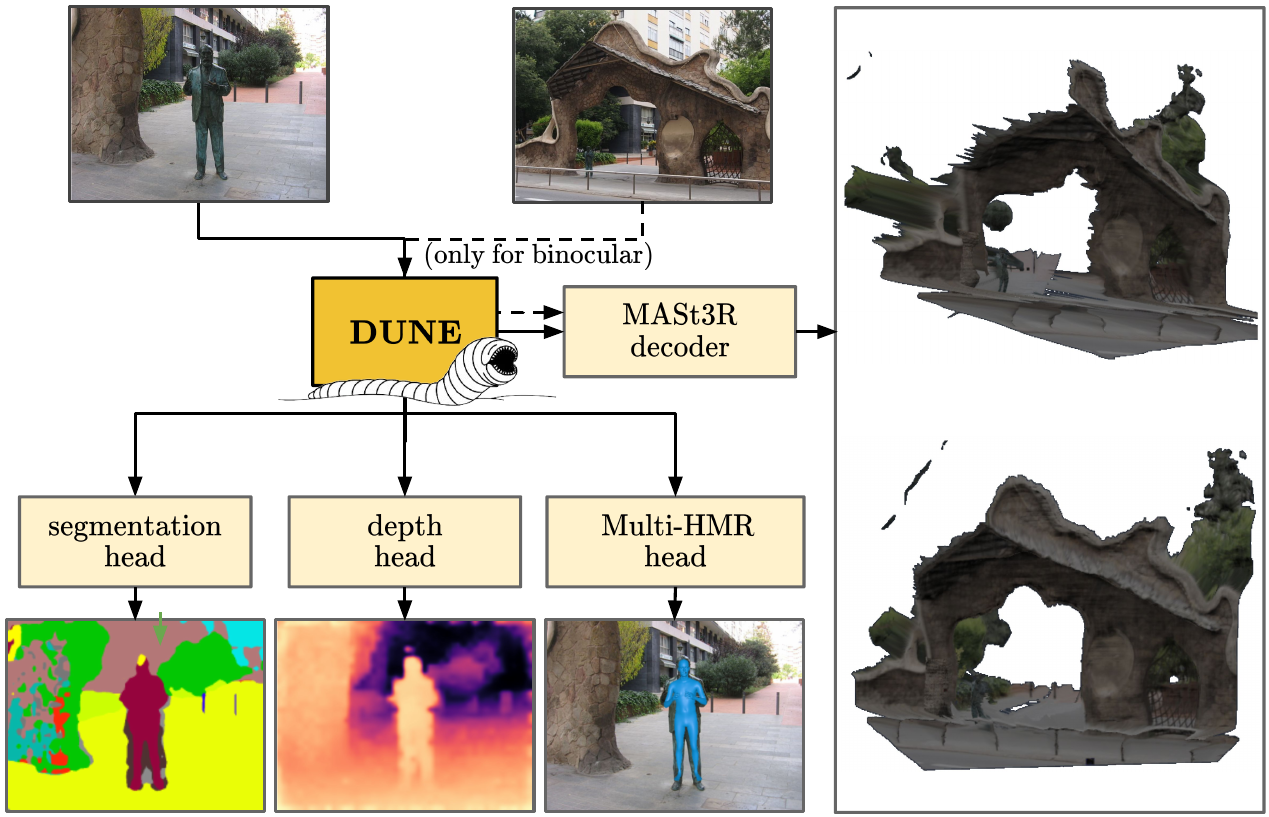}
}
\vspace{-0.1cm}
\caption{
\textbf{\ours} is a universal encoder for 2D and 3D tasks distilled from heterogeneous teachers. It enables multi-task inference with a single encoder. Teachers are \dino~\cite{oquab2024dinov2}, \master~\cite{mast3r}, and \hmr~\cite{multihmr} (see Fig.~\ref{fig:overview} for distillation details).
}
\label{fig:teaser}
\end{figure}

%% file: tex/float/fig_pca_mini.tex
\begin{figure}[t]
\centering
\adjustbox{max width=\linewidth}{
\begin{tabular}{c@{\hspace{20pt}}c@{\hspace{3pt}}c@{\hspace{2pt}}c@{\hspace{20pt}}c}
   \textbf{Input}   & \textbf{\dino} & \textbf{\master} & \textbf{\hmr}        & \textbf{\ours} \\
\includegraphics[width=1.75cm]{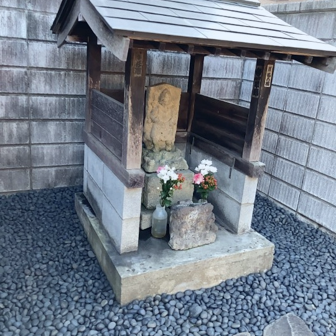} &
    \includegraphics[width=1.75cm]{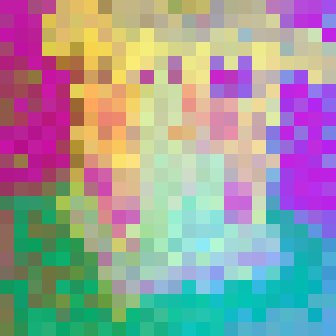} &
    \includegraphics[width=1.75cm]{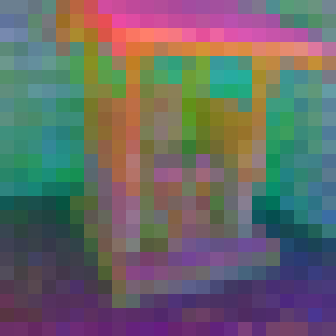} &
    \includegraphics[width=1.75cm]{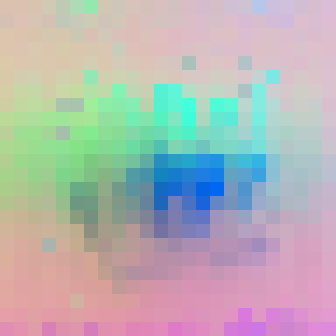} &
    \includegraphics[width=1.75cm]{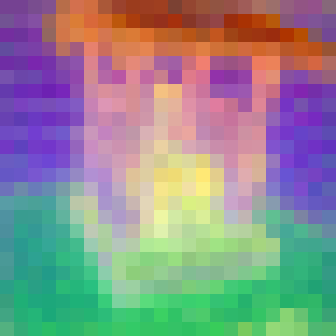} \\
\includegraphics[width=1.75cm]{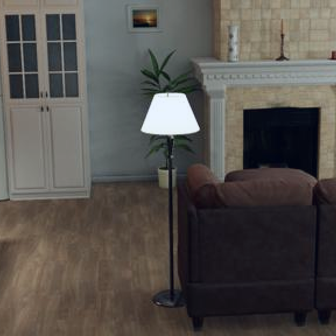} &
    \includegraphics[width=1.75cm]{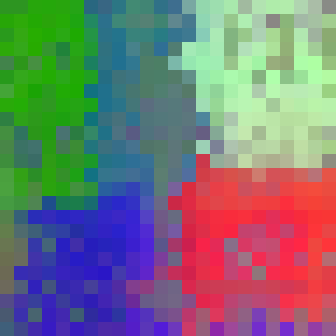} &
    \includegraphics[width=1.75cm]{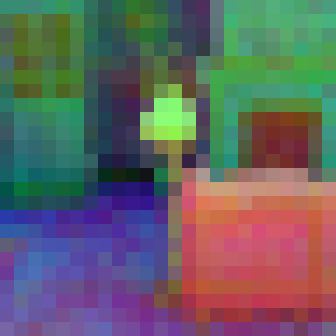} &
    \includegraphics[width=1.75cm]{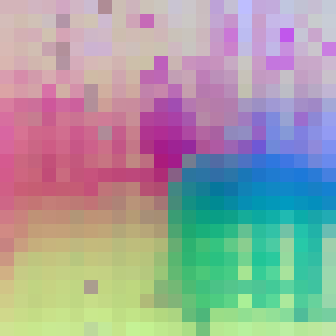} &
    \includegraphics[width=1.75cm]{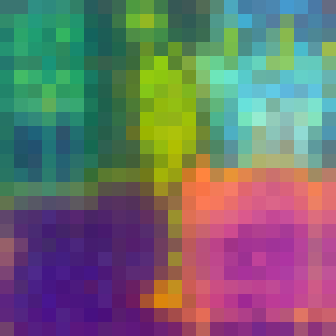} \\
\includegraphics[width=1.75cm]{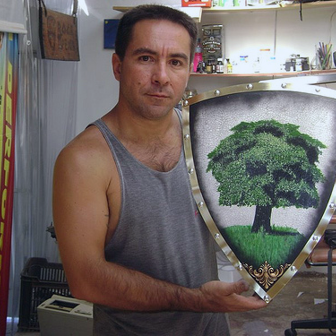} &
    \includegraphics[width=1.75cm]{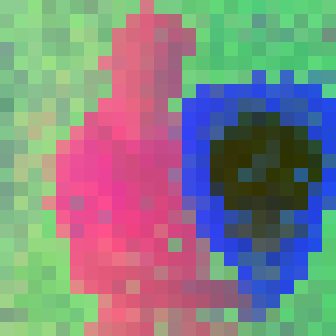} &
    \includegraphics[width=1.75cm]{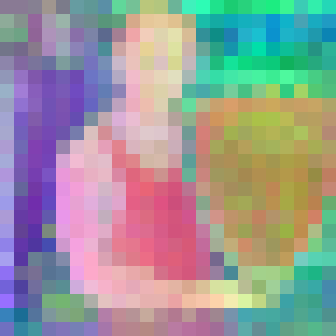} &
    \includegraphics[width=1.75cm]{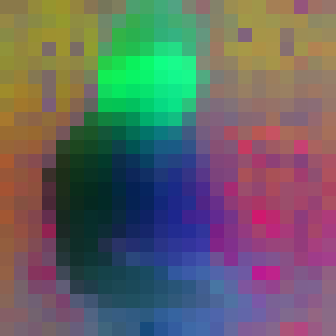} &
    \includegraphics[width=1.75cm]{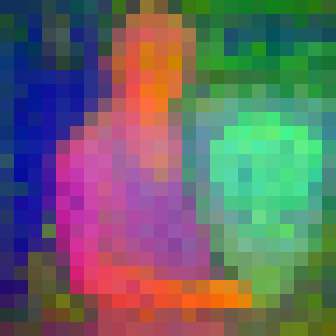} \\
\end{tabular}
}
\vspace{-0.3cm}
\caption{
    {\bf PCA visualization of encoder outputs.}
    Given an image, we extract patch embeddings from the encoders of the teacher models and our student, and reduce their dimension to 3 via PCA.
}
\label{fig:pca_mini}
\end{figure}

%% file: tex/02_related.tex
\section{Related work}
\label{sec:related}

\paragraph{Combining multiple models.}
Various approaches to combining models have been explored. One method involves extracting features from each model and either concatenating or fusing them for downstream tasks,
a strategy used in early works~\cite{sun2019vidbert} as well as more recently in \cite{cheng2023can,kar2024brave}.
While effective, this approach is often impractical due to memory
and compute requirements.
Other studies focus on merging models together~\cite{wang2024localizing,sanh2022multitask,jang2025model,dey2024revla,davari2023model,stoica2023zipit}, typically under the assumption that models have the same architecture and size.

\paragraph{Multi-teacher distillation} is another way to combine multiple models into one. It replaces the encoder of each teacher by a single student encoder.
This encoder is obtained by distilling the outputs of the teacher encoders on well-chosen data.
This offers great flexibility: there are a-priori no constraints on the student architecture or size.
Recent examples include generic methods like AM-RADIO~\cite{ranzinger2024radio,heinrich2024radioamplifiedimprovedbaselines}, UNIC~\cite{sariyildiz2024unic}, PHI-S~\cite{ranzinger2024phi}, UNIT~\cite{zhu2024unit},
robotics-specific models
such as Theia~\cite{shang2024theia},
and methods using sequential distillation such as in~\cite{roth2024fantastic}.
By distilling from three or more teachers, these teacher-agnostic strategies sometimes even yield students that outperform their teachers on some tasks~\cite{ranzinger2024radio,sariyildiz2024unic}.
However, all of these methods focus on distilling teachers of the same nature, foundation models trained on massive web-crawled image collections with self- or weak supervision.
Common choices include \dino~\cite{oquab2024dinov2}, CLIP~\cite{radford2021clip}, and SAM~\cite{wang2023sam}. The students obtained are then
applied to
common benchmark tasks like classification, segmentation, or monocular depth.

In this paper, we explore distillation from heterogeneous teachers, \ie trained on distinct domains and solving diverse tasks. Specifically, we distill from \master~\cite{mast3r} for binocular 3D tasks, \hmr~\cite{multihmr} for human mesh recovery, and \dino~\cite{oquab2024dinov2} for 2D tasks like monocular depth and semantic segmentation. This results in an encoder that excels across many of these tasks out of the box.

\paragraph{Training using heterogeneous data domains}.
All aforementioned multi-teacher distillation methods have another thing in common: they distill only from generic data
suitable to all teachers: DataComb-1B~\cite{gadre2024datacomp} for AM-RADIO, ImageNet-1K~\cite{deng2009imagenet} for UNIC~\cite{sariyildiz2024unic} and Theia~\cite{shang2024theia}.
Heterogeneous data has recently been used for distillation of different datasets for classification~\cite{ypsilantis2024udon,iordache2024multi} or domain adaptation~\cite{tang2024direct}.
To the best of our knowledge, no existing work has looked into distillation using data that contains natural images as well as synthetic data from 3D engines, CAD models, simulators, and rendered from structure from-motion reconstructions.
Our
teachers' training data even vary in terms of content, from
small groups of people for human mesh recovery to empty indoor rooms and outdoor buildings for 3D.
We  hypothesize that leveraging such data is necessary to accurately distill highly specialized teacher models such as \master~\cite{mast3r} and \hmr~\cite{multihmr}, and attempt to do so in this paper.

\paragraph{Relation to multi-task learning.}
As in multi-task training, our heterogeneous teacher distillation tackles specialized tasks such as segmentation, human pose estimation, and 3D reconstruction with a single encoder. However, it also preserves the generic nature of some teachers, ensuring strong generalization. Unlike
multi-task training, distillation relies on teacher outputs instead of ground-truth labels and does not require access to the original training data.

\paragraph{Relation to 3D-to-2D distillation.} Prior work~\cite{hou2021pri3d,yue2024fit3d} has used 3D-aware models to improve results on 2D tasks. While these methods are related, we show that DUNE excels at \textit{both} 2D and 3D.

%% file: tex/03_htd.tex
\section{Problem definition and challenges}
\label{sec:method}

First, we describe the problem of \textbf{heterogeneous teacher distillation} or \textbf{co-distillation},
a challenging multi-teacher distillation setup where teacher models significantly vary with respect to
(a) the goals behind their design, and (b) the data they were trained on.
We would like to jointly distill from
a set of teachers that satisfy
the following
properties:
\begin{enumerate}
    \item
    \textbf{The teachers cover an heterogeneous set of tasks.}
    We want to jointly distill teachers that vary in the \textit{design objectives} they are trained on. This includes
    \textit{task-agnostic} teachers (models with strong generalization properties, typically self-supervised and trained on pretext tasks)
    together with
    \textit{specialized} models tailored to
    specific tasks.
    \item  \textbf{Their individual training sets consist of heterogeneous data.}
    We want to jointly distill teachers trained on huge generic image datasets crawled from the web, together with teachers trained on highly curated and potentially carefully annotated datasets composed of natural or synthetic images.
\end{enumerate}

\noindent This leads to the
question driving our research: \textit{Can we distill from such an heterogeneous set of teachers
and get a universal visual encoder that retain strong generalization abilities and at the same time} excels at multiple diverse tasks?

\subsection{Task-agnostic {\bf \vs} specialized teachers}
\label{sec:het_tasks}

Following our problem definition,
we differentiate between teacher models trained on proxy tasks that capture inherent image priors and those specializing in
a specific task or set of tasks.
While the former might consist solely of a visual encoder and be referred to as a foundation model, the latter typically include task-specific decoder heads and are trained using
task-specific data and supervision.
Our goal is not only to match the performance of specialized
models, but also retain
the generalization capability of the representations learned from the task-agnostic teachers.

\noindent \textbf{Task-agnostic teachers}
refers to models trained on proxy tasks such as self-supervised objectives like context prediction or photometric invariance.
They aim to capture broadly used visual priors that are useful on a wide range of tasks.
This includes
\dino~\cite{oquab2024dinov2} that we consider as one of the teachers for \ours.

Such models are typically evaluated on tasks
for which the encoder representations can be used directly, such as
$k$-NN or zero-shot~\cite{radford2021clip} classification,
or by training linear classifiers for various classification tasks at the image or pixel level (\eg, semantic segmentation or monocular depth estimation using a linear head).
Notably, these models are recognized for the \textit{generalization} strength of their representations. Those have been shown to be beneficial to
a wide range of tasks~\cite{wysoczanska2024clip,yang2024depth}. However, on their own, they usually underperform compared to specialized models trained with supervised or
privileged information.

\looseness=-1
\paragraph{Specialized teachers}
\footnote{We use \textit{specialized} rather than \textit{task-specialized} to account for teachers solving multiple tasks within an area (human understanding, 3D vision).} focus on specific perception domains, such as 3D for \master~\cite{mast3r} or human pose understanding for \hmr~\cite{multihmr}.
They are typically trained with weak or strong supervision, and, although all use ViT encoders, they leverage a domain-specific parameterization and require varying levels and types of annotation.
The encoders of specialized teachers may differ in the \textit{nature} of the information they encode (SMPL body model~\cite{smpl} parameters \vs dense matches) and \textit{how} they encode it (\eg, \hmr captures the full 3D human pose in a single patch token).

\looseness=-1
\paragraph{Generalization \vs specificity trade-off.}
The distinction above highlights the trade-off between generalization to novel tasks, afforded by task-agnostic teachers, and performance on specialized tasks coming from specialized teachers.
By incorporating this distinction into our formulation, we can interpret the proposed distillation setup in two ways:
(a)~as a means to enhance the performance of specialized teachers on certain novel tasks by leveraging task-agnostic models, or
(b)~as a way to improve the performance of self-supervised foundation models on the set of specialized tasks.

\subsection{Distillation using heterogeneous data}
\label{sec:het_data}

The diversity of teacher training domains has not been a significant concern for existing multi-teacher distillation approaches~\cite{ranzinger2024radio, sariyildiz2024unic, lu2024swiss}. Foundation models like \dino~\cite{oquab2024dinov2}, SAM~\cite{wang2023sam}, and CLIP~\cite{radford2021clip, ilharco2021openclip} are all trained on data of a similar nature, \ie ``generic'' datasets like LAION~\cite{schuhmann2022laion} or DataComp1B~\cite{gadre2024datacomp},
and recent works~\cite{ranzinger2024radio, sariyildiz2024unic, lu2024swiss} show that even using ImageNet-1K~\cite{deng2009imagenet} is enough to match teacher-level performance with a unified encoder.

\looseness=-1
One of the main challenges of co-distillation
is the teachers' training data that spans \textit{multiple visual domains}.
Alongside visual encoders trained on natural images, \ie \dino~\cite{oquab2024dinov2}, we aim to jointly distill encoders trained on diverse types of data (\eg synthetic data from 3D engines, CAD models, simulators, \etc)
and with diverse content (\eg data focused on small groups of people in the case of \hmr~\cite{multihmr}, or empty indoor rooms and outdoor buildings for \master~\cite{mast3r}).
 When jointly distilling teachers trained on such heterogeneous data, the choice of data for distillation becomes less straightforward: \textit{Do we need to incorporate data from all teacher domains, or is generic data sufficient?}

%% file: tex/04_method.tex
\section{Framework for co-distillation}
\label{sec:setup}

\input{tex/float/fig_overview}

In this section, we introduce our co-distillation framework for heterogeneous teacher distillation, as shown in~\Cref{fig:overview}. We begin with the fundamentals of multi-teacher distillation, then discuss teacher-specific projectors and
our
choices
towards an efficient inference.

\subsection{Background on multi-teacher distillation}
We focus on visual encoders based on the ViT~\cite{dosovitskiy2021an} architecture.
These models take an image $x \in \gI$ as input and produce a set $Z \in \gZ$ of feature vectors, where $\gZ \in \mathbb{R}^{(HW+1) \times d}$.
This feature set includes $HW$ features for the $H \times W$ patches, along with an optional global feature corresponding to a CLS token.
We assume each feature vector has a dimensionality of $d$.
These feature vectors are typically used as an input to one or more decoder heads to tackle specific computer vision tasks.

Let $\gT = \{ \gT_1, \dots, \gT_N \}$ represent the set of $N$ teachers we want to distill, each parametrized by an encoder $t_i(x)$.
Our objective is to learn the parameters of a student model $f$ that produces outputs closely aligned with those of \textit{all} teachers simultaneously.
We train this student encoder by applying a distillation loss on both the global and patch token features.
Let $f(x): \gI \rightarrow \gZ$ denote the student’s encoder, and $h_i$ represent a teacher-specific projector for each teacher $\gT_i, i=1,..,N$.
We minimize the cosine-similarity and smooth-$\ell_1$ losses combined among all teachers:

\begin{equation}
    \mathcal{L}_{\text{distil}} = \sum_{i=1}^N \mathcal{L}_{cos} \big(f_i(x), t_i(x)\big) + \mathcal{L}_{s\ell_1} \big(f_i(x), t_i(x)\big),
\end{equation}
where $f_i = h_i(f(x))$, and $\mathcal{L}_{cos}$ and $\mathcal{L}_{s\ell1}$ denote the cosine and smooth-$\ell_1$ losses, respectively, as defined in~\cite{sariyildiz2024unic}.

The process described above imposes no restriction on the types of teachers used and,
with minor modifications,
forms the foundation of several recent works~\cite{ranzinger2024radio, sariyildiz2024unic, shang2024theia}.

\subsection{Teacher-specific projector design}
\label{sec:projector_design}

Existing multi-teacher distillation approaches~\cite{ranzinger2024radio,sariyildiz2024unic, lu2024swiss} address the issue of teacher-specific (or ``complementary'') information derived from the different training objectives of teachers using teacher-specific modules also known as \textit{projectors}.
In our problem formulation (\Cref{sec:method}),
such modules become even more important:
The patch representations from the encoder and their interactions can differ significantly between teachers.
The \hmr model~\cite{multihmr}, for example, requires pose information for the whole human body to be captured in the representation of the patch where the
head and nose are detected,
since the corresponding decoder only takes such patches as input.

A simple projector design is a
two-layer MLP appended to the top of the encoder for each teacher~\cite{ranzinger2024radio}.
More complex designs have been explored
~\cite{sariyildiz2024unic,lu2024swiss, shang2024theia}.
UNIC~\cite{sariyildiz2024unic} attaches multiple additional MLPs to intermediate layers of the student. Its ``ladder of projectors'' (LP)  improves information flow to the teacher-specific parameters and was shown to help improve distillation across teachers and tasks.

For all the cases mentioned above, projectors operate \textit{per-patch}, \ie teacher-specific parameters cannot explicitly capture patch feature interactions.
This requires inter-patch interactions specific to each teacher to be embedded in the attention layers of the shared encoder.
In other words, it is the shared encoder that has to model all patch interactions relevant for any teacher.
This
motivated us to introduce attention-based projectors that we define next.

\myparagraph{Transformer projectors.}
For projectors to model interactions across patches,
a sensible and efficient design would be an attention-based projector
composed of a single transformer~\cite{vaswani2017attention} block:
\begin{align}
    a &= f(x) + \text{SA}(\text{LN}(f(x))), \\
    m &= a + \text{MLP}(\text{LN}(a)), \\
    h &= \text{Linear}(m),
\end{align}
where LN denotes
layer normalization, SA a multi-head self-attention layer, MLP a two-layer perceptron and Linear a fully-connected layer.
We refer to this projector design as a \textbf{transformer projector} or \textbf{TP}.
\Cref{sec:experiments} compares it
with more standard projectors.

\subsection{Co-distillation with heterogeneous data}
\label{sec:co-het}
In co-distillation, the optimal distribution of data to distill from is not trivial:
Data associated with a specific teacher can be irrelevant or even harmful to others.
It therefore makes sense to control which data gets forwarded to each teacher-specific projector.

Let $\gD_i$ denote the data associated with teacher $\gT_i \in \gT$ with $i=1..N$. We assume that $\gD_i \cap \gD_j = \emptyset$ for all specialized teachers $i,j$, \ie the specialized teachers are all trained on different datasets. We also assume that all task-agnostic teachers are trained with the same generic data
$\gD_g$. Let $h_i$ denote the projector associated with teacher $i$.
We explore three simple ways of sharing datasets across teachers:
\begin{itemize}
    \item \textbf{No data sharing}: Projector $h_i$ receives only $\gD_i$, \ie the data associated with its teacher $\gT_i$.
    \item \textbf{Full data sharing}: Projector $h_i$ receives \textit{all} data,
    \ie $\cup \gD_i, i=1..N $, as well as generic data $\gD_g$.
    \item \textbf{Generic data sharing}: Projector $h_i$ receives only $\gD_i$ (the data associated with $\gT_i$) and generic data $\gD_g$.
\end{itemize}
In the next section, we evaluate these different ways of sharing data across teachers and study their impact on area-specific tasks as well as representation generalization tasks.

\subsection{Fine-tuning task heads for efficient inference}
\label{sec:inference}

Most related works~\cite{ranzinger2024radio,heinrich2024radioamplifiedimprovedbaselines,shang2024theia,lu2024swiss} require the teacher-specific projectors learned during training to be used during inference.
This allows for a plug-and-play reuse of task-specific decoders, yet this results in more parameters, not only for the student encoder itself but also for all the teacher-specific projectors.
The number of those additional parameters scales linearly with the number of teachers.

Projectors would become irrelevant if the decoder modules were jointly fine-tuned for the tasks to solve, once the student is trained.
Such an approach offers several advantages:
It introduces no additional modules during inference, it keeps the encoder size and memory constant regardless of the number of teachers, and, most importantly, it does not impact inference time.
For \ours, we opt for that second option and fine-tune the different heads and decoders, a one-time cost that enables a more efficient inference.

%% file: tex/float/fig_overview.tex
\begin{figure*}[t]
\centering
\adjustbox{max width=\linewidth}{
\includegraphics[]{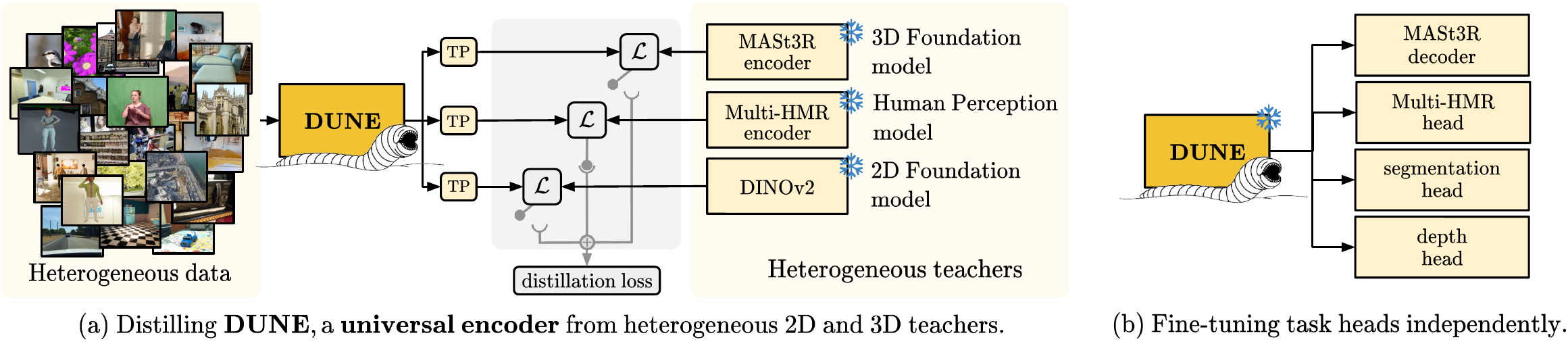}
} \\[-0.2cm]
\caption{
{\bf Overview of the \ours encoder training process.}  (a) \ours is trained via distillation from \textit{heterogeneous} teachers across 2D vision, 3D vision, and 3D human perception, leveraging diverse data from multiple visual domains. {We use \textit{teacher dropping} regularization from~\cite{sariyildiz2024unic}. (b) Task-specific heads are then fine-tuned independently for each task, with the \ours encoder kept frozen.
}}
\label{fig:overview}
\end{figure*}

%% file: tex/05_experiments.tex
\section{Experiments}
\label{sec:experiments}

In this section, we experimentally evaluate the co-distillation framework from~\cref{sec:setup}.
After introducing the evaluation protocol (\cref{sec:protocol}), we experimentally validate the
projector design, the use of heterogeneous data and data sharing across projectors, and we present results on five
very different
tasks (\cref{sec:results}).
We then present some feature and loss analyses, as well as some qualitative results comparing the teacher and student model outputs (\cref{sec:analysis}).

\subsection{Evaluation protocol}
\label{sec:protocol}

\paragraph{Teachers.}
We select a representative set of heterogeneous teachers for our experimental validation.
The first teacher is \dino~\cite{oquab2024dinov2} with registers~\cite{darcet2024register}, a self-supervised task-agnostic model
whose strong representations have been used in many computer vision tasks, and a common teacher choice in
multi-teacher distillation~\cite{ranzinger2024radio,sariyildiz2024unic}.
Then, we select two state-of-the-art domain-specialized models: \hmr~\cite{multihmr} a model for human mesh recovery, the winner of the Robin Challenge at CVPR'24,\footnote{\url{https://rhobin-challenge.github.io/}} and \master~\cite{mast3r} a 3D foundation model, the winner of the Map-free Visual Relocalization at ECCV'24.\footnote{\url{https://nianticlabs.github.io/map-free-workshop/2024/}}
The selected teachers are designed to span across both 2D and 3D tasks, with the two 3D-oriented teachers differing in the \textit{nature} of the 3D information they encode (SMPL parameters \vs dense matches) and \textit{how} they encode it (\eg Multi-HMR captures the full 3D human pose in a single patch token).
We use the publicly available ViT-Large models for all teachers.

\looseness=-1
\paragraph{Datasets.}
We use 19 publicly available datasets from the training sets of the three teachers as distillation data, leading to around 20.7M images in total:
ImageNet-19K~\cite{deng2009imagenet} (2021 release~\cite{yang2020fda}), Mapillary~\cite{mapillary} and Google Landmarks v2~\cite{google-landmarks-v2} from \dino,
AGORA~\cite{agora}, BEDLAM~\cite{bedlam}, UBody~\cite{ubody,how2sign} and CUFFS~\cite{multihmr} from \hmr;
Habitat~\cite{habitat19iccv},
ARKitScenes~\cite{arkitscenes},
Blended MVS~\cite{blendedMVS},
MegaDepth~\cite{megadepth},
ScanNet++~\cite{scannet++},
CO3D-v2~\cite{co3d},
Map-free~\cite{mapfree},
WildRgb~\cite{wildrgb},
VirtualKitti~\cite{vkitti2},
Unreal4K~\cite{unreal4k},
TartanAir~\cite{tartanair2020iros} and
DL3DV~\cite{ling2024dl3dv}
from \master.
We only use the images of those datasets
and discard all annotations.

\paragraph{Implementation details.}
During distillation, our student is composed of a
ViT-Base encoder and three projector heads, one for each teacher.
In our initial experiments, we used the Ladder of Projectors (LP) design from UNIC~\cite{sariyildiz2024unic}, which was shown beneficial for dense prediction tasks.
We also evaluate the
Transformer Projector we introduced
in \cref{sec:projector_design}.
Unless otherwise specified,
\textit{projectors are discarded} after distillation and are not used for evaluations or inference.
By default, we fix the compute budget for all the distillation
variants to process $100 \times 1,281,167$ images, which amounts to 100-epoch training on ImageNet-1K~\cite{russakovsky2015ilsvrc}.
We follow the data augmentation and optimization practices of~\cite{sariyildiz2024unic}, and also use their teacher dropping regularization.
Unless otherwise stated, we distill our encoders at image resolution $336\times336$.
We further distill some of our models at resolution $448\times448$ for a couple of additional epochs.

\paragraph{Evaluation tasks and metrics.}
We evaluate our encoder on the tasks that our selected teachers excel at.
For domain-specialized teachers, we choose tasks for which they are the state of the art: multi-person human mesh recovery for \hmr and map-free visual relocalization for \master.
For \hmr, we report results on the BEDLAM validation set, using F1-Score to evaluate detection performance and PA-PVE to measure mesh reconstruction errors.
For \master, we report the Area under the Curve (AUC) for samples with Virtual Correspondence Reprojection Error (VCRE) below 90px on the validation set of the Map-free Visual Relocalization dataset~\cite{mapfree}. We also provide results for multi-view depth estimation and multi-view camera pose regression in the supplementary material.
We evaluate the generalization performance of our encoder on semantic segmentation with ADE20K~\cite{zhou2019semantic} (mIoU) and depth estimation on NYUdv2~\cite{nyu} (RMSE), following related work~\cite{ranzinger2024radio,sariyildiz2024unic}.
{We also report comparisons to 3D-to-2D distillation methods~\cite{hou2021pri3d,yue2024fit3d} as well as segmentation results on Cityscapes~\cite{cityscapes}, NYUdv2~\cite{nyu} and ScanNet~\cite{scannet} in the supplementary material.
Further details on the evaluation protocol are provided there too.

\paragraph{Fine-tuning task heads.}
For specialized tasks, we attach the corresponding teacher’s decoder module to the frozen encoder and fine-tune the decoder for that task.
For semantic segmentation and depth estimation, we train a linear head from scratch, appending it to the frozen encoder, as in~\cite{oquab2024dinov2}.
For segmentation, re-using the frozen transformer projector for the \dino teacher before training the linear layer significantly improved performance.
For this case only, we report results \textit{using} the projector, similar to~\cite{ranzinger2024radio}.
We compare results with and without projectors in the supplementary material.

\input{tex/float/tab_tp_slp_ablation}

\subsection{Results}
\label{sec:results}

\paragraph{Is ``generic'' data enough?}
We start our experiments by checking whether or not a large, generic dataset like ImageNet
is enough for distilling heterogeneous teachers.
We train two student models using a sparsely-connected Ladder of Projectors (LP), \ie a multi-layer Perceptron attached after every 3 encoder blocks. The first model only uses ImageNet-19K as distillation data whereas the second one uses all distillation datasets mentioned above. From
{the results presented in~\cref{tab:tp_vs_slp},}
we observe that using
additional specific data
improves the performance of the students on all the tasks by a decent margin.

\input{tex/float/tab_data_sharing}

\paragraph{Transformer Projectors.}
Looking at~\cref{fig:pca_mini}, we notice that feature similarity
patterns significantly vary from one teacher's encoder to another.
Visualizing the attention maps from the final layer encoder of each teacher (provided in the supplementary material) also shows clear differences across models: \master produces highly localized attentions while the \dino attention span is much larger.
Attention maps for \hmr seem more focused to the persons' head, ignoring the remaining content in many cases.

\looseness=-1
We argue that capturing feature similarities of such different {spatial} extents would be easier
{using the Transformer projector (TP) presented in~\cref{sec:projector_design}.}
To test this hypothesis, we replace the LP-based projectors attached to multiple layers of the student encoder with {a single} TP projector {after the last layer of the encoder}.
We compare~\cref{tab:tp_vs_slp} the LP and TP designs,
{as well as the simple MLP projector (SP) used in~\cite{ranzinger2024radio} (row 3).
We observe that TP outperforms both LP and SP across all tasks.}

\input{tex/float/tab_ours_vs_baselines_teachers}

\paragraph{Sharing data across teacher projectors.}
{In~\cref{sec:co-het}}, we have presented three different ways of sharing data
across teachers.
In~\cref{tab:data_sharing}, we evaluate all three and see that sharing all data among teachers generally yields the best performance.
It suggests that the domain gap between the datasets might not be an issue and teachers still produce useful information for out-of-domain images.
Interestingly, sharing only generic data is the best for semantic segmentation.
This suggests that
semantic information is better preserved by the encoder when ImageNet-19K is shared by \master and \hmr.

\paragraph{Comparing to state-of-the-art multi-teacher distillation.}
In the middle part of~\cref{tab:ours_vs_baselines}, we compare our distilled encoder to the strongest comparable ViT-Base encoders available, \ie \dino~\cite{oquab2024dinov2} and \amradio~\cite{heinrich2024radioamplifiedimprovedbaselines}.
For both models, we fine-tune the decoder heads for each task using the same procedure as for our encoder.
\ours outperforms both models on all evaluations except semantic segmentation, where \amradio achieves the best results by a significant margin.
This is expected, as \amradio is distilled from two semantically rich teachers, CLIP and OpenCLIP, along with the strong segmentation model SAM~\cite{wang2023sam}.

\paragraph{Improving Map-free Visual Relocalization.}
In~\cref{tab:mapfree}, we report results from the official leaderboard of the Map-free visual relocalization dataset.\footnote{\url{https://research.nianticlabs.com/mapfree-reloc-benchmark}}
On this benchmark, \master~\cite{mast3r}
{was shown to significantly outperform} the state of the art~\cite{loftr,dust3r,mickey}, where all recent approaches build upon a ViT-Large backbone.
{Surprisingly, when replacing the ViT-Large encoder of \master by the frozen ViT-Base encoder of our student model, and then fine-tuning the \master decoder},
we obtain even better performance than \master
despite using a significantly smaller encoder.

\input{tex/float/tab_mapfree}

\subsection{Analysis and visualizations}
\label{sec:analysis}

\looseness=-1
In this section, we provide some analysis
to better understand the distillation process and feature spaces
learned by co-distillation.
We then provide qualitative results
on the tasks of the two specialized teachers.

\paragraph{Feature analysis.}
In \cref{fig:explained_variance}, we plot the cumulative explained variance, \ie the proportion of the dataset’s variance that is cumulatively explained by each additional PCA component.
We plot curves for three representative datasets, comparing the teacher encoder features to the corresponding student projector features.
Interestingly, we observe that rather than seeing a significant change across datasets, the most consistent difference in terms of feature compactness is across teachers.
The \hmr teacher needs consistently fewer PCA components to explain its feature variance while \dino needs the most.
We argue that this could be because the \hmr model has a more specialized task and training data, while \dino has been trained with a more diverse dataset aiming to be a versatile encoder.
\master seems to be somewhere in the middle as a versatile encoder specialized in 3D tasks.

In the same figure, we also plot explained variance
on the three datasets for the features of our learned encoder after the teacher-specific heads and observe that
{they follow the ranking of the corresponding teacher features.}
However, student representations are consistently more compact than the corresponding teachers.

\paragraph{Correlation of loss updates.}
In \cref{fig:teacher_alignment}, we plot how often loss updates are correlated for three pairs of teachers, \ie we look at the change in loss magnitudes after each weight update, and measure the correlation between the loss fluctuations for each teacher.
If the teachers are well aligned, we expect a strong positive correlation (\eg, minimizing the loss for \hmr also minimizes the loss for \dino), while a low correlation would indicate the teacher feedback is less aligned and training might be more unstable.

The first four bars represent the correlation for two training data choices (ImageNet-19K or all 19 datasets) and two projector designs.
We observe that with LP, teachers are always less aligned compared to TP, regardless of whether training is performed only on ImageNet-19K or on all datasets.
This may explain LP's inferior performance, particularly on specialized tasks, as shown in~\cref{tab:tp_vs_slp}.
For TP, we additionally measure teacher alignment across the three data-sharing options presented in~\cref{sec:co-het} (rightmost three bars).
All in all, we observe that using all data for all teachers results in the highest possible correlation on all teacher pairs, something that is further reflected on performance in~\cref{tab:data_sharing}.

\begin{figure}[t]
\vspace{10pt}
    \centering
    \includegraphics[width=1\linewidth]{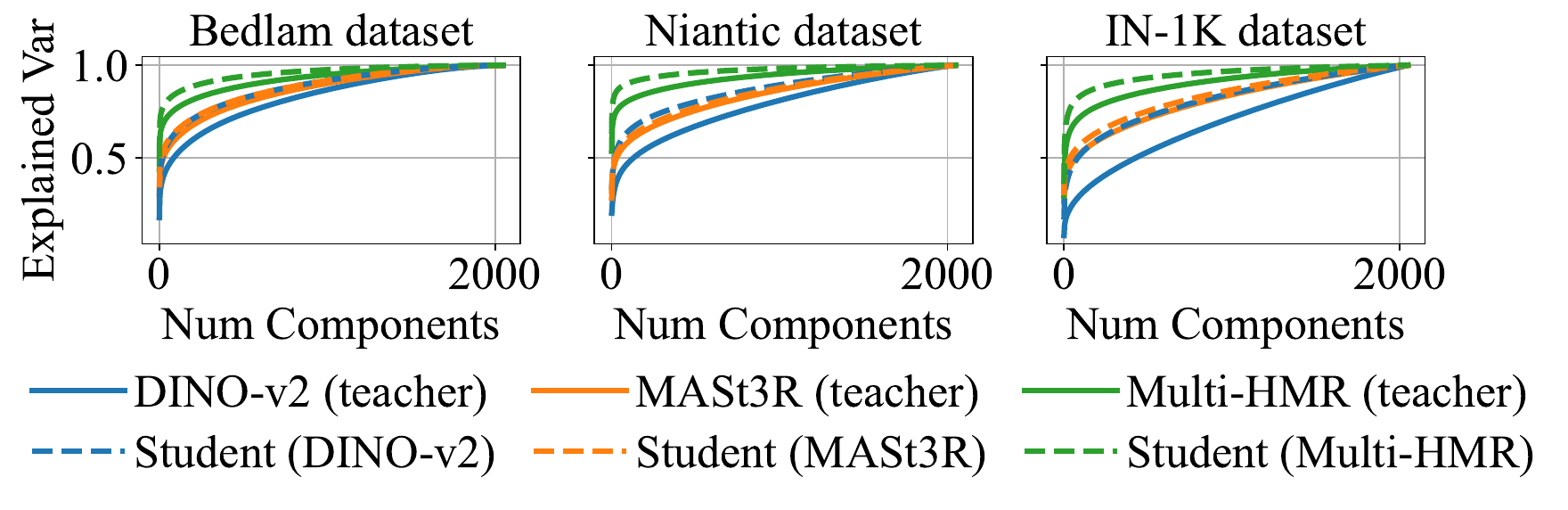} \\
    \vspace{-0.28cm}
    \caption{
            \textbf{
            Cumulative explained variance} computed over features from three representative datasets, for the three teacher encoders (solid lines) and student's projectors (dashed lines).
    \label{fig:explained_variance}}
\end{figure}

\begin{figure}[t!]
\vspace{10pt}
    \centering
    \includegraphics[width=\linewidth] {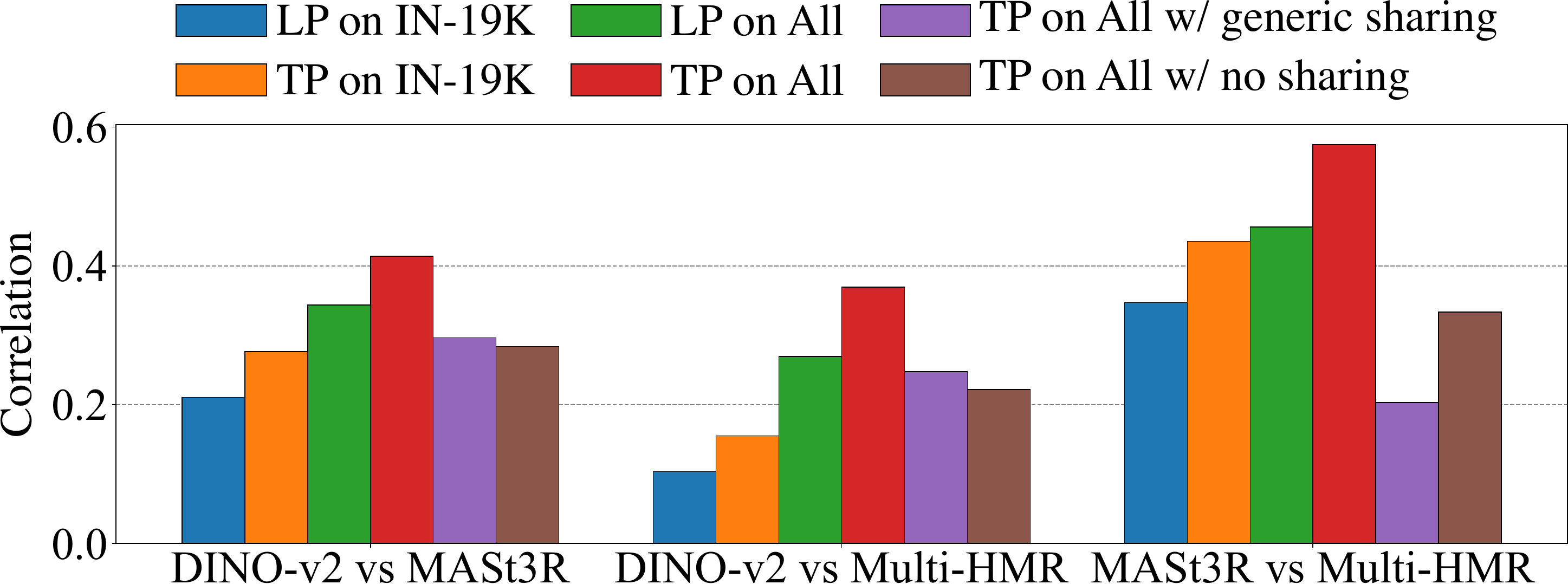} \\
    \vspace{-0.24cm}
    \caption{\textbf{Correlation of loss updates} during training for each pair of teachers when training with different strategies. Training with TP leads to more \textit{alignment} between teachers regardless of the training data. On the other hand, using all data with all teachers seems to be the best data strategy to improve teacher alignment.}
    \label{fig:teacher_alignment}
\end{figure}

\paragraph{Qualitative Results.}
We present side-by-side qualitative results for the specialized tasks our encoder jointly solves: human mesh recovery in~\cref{fig:qualitative_multihmr} and 3D reconstruction in~\cref{fig:qualitative_mast3r}.
\ours, combined with the corresponding task decoder, produces visually similar outputs to those of the \hmr and \master teachers, respectively.

\begin{figure}[t]
    \centering
    \includegraphics[width=.97\linewidth]{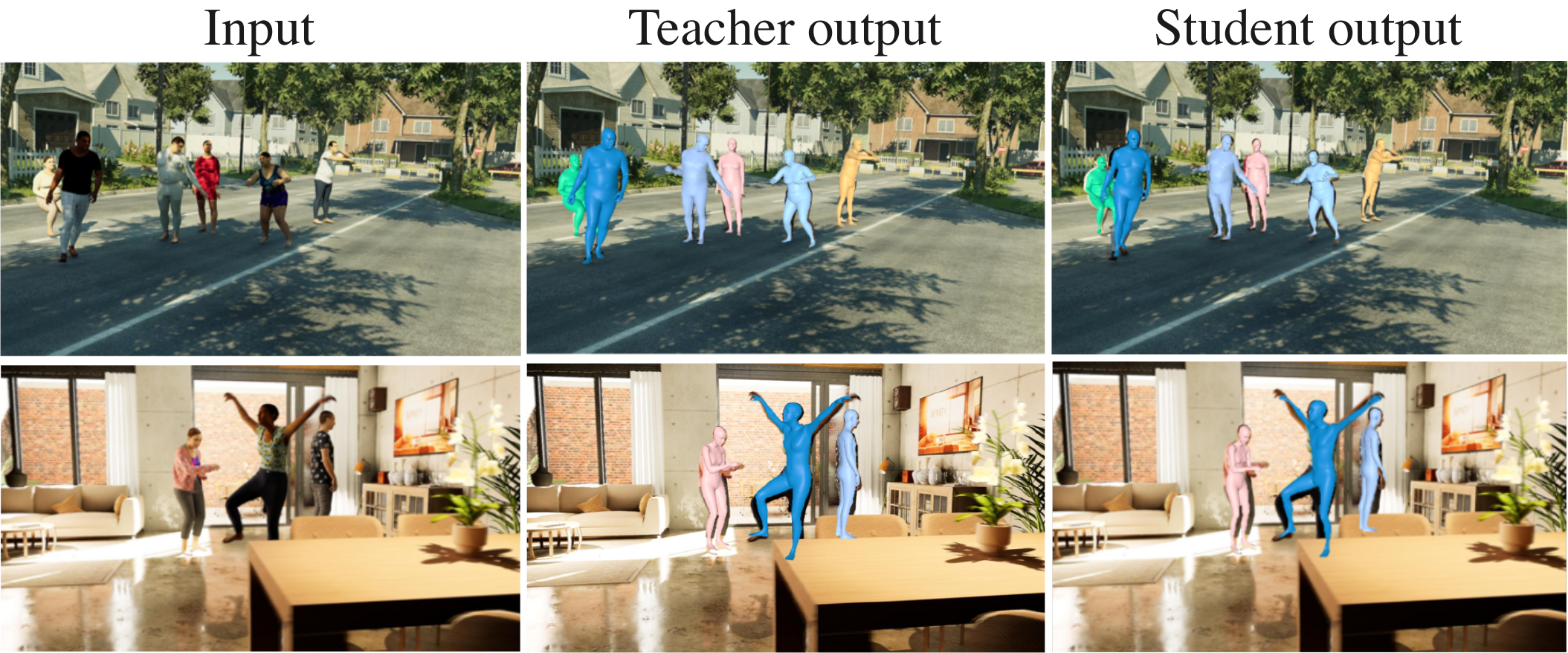}
    \vspace{-0.28cm}
    \caption{\textbf{Qualitative comparison on human mesh recovery} between \hmr {} {(the teacher)} and \ours{} {(the student)}.}
    \label{fig:qualitative_multihmr}
\end{figure}

\begin{figure}[t]
    \centering
    \includegraphics[width=.97\linewidth]{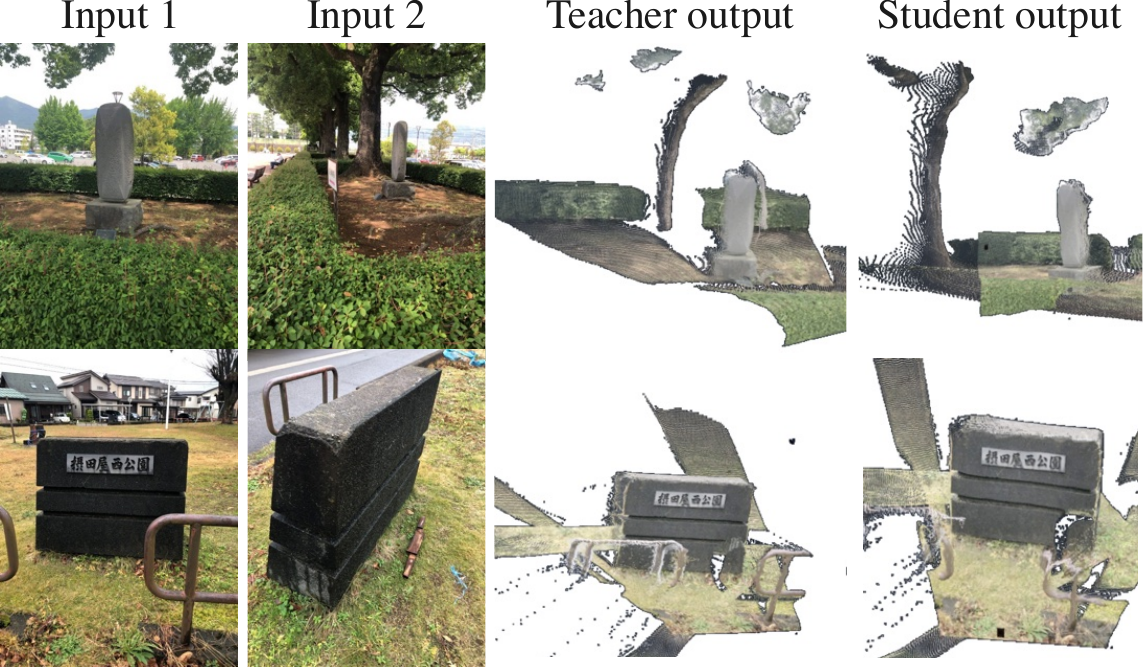}
    \vspace{-0.28cm}
    \caption{\textbf{Qualitative comparison on 3D reconstruction} between \master (the teacher) and \ours (the student).}
    \label{fig:qualitative_mast3r}
\end{figure}

%% file: tex/float/tab_tp_slp_ablation.tex
\begin{table}[t!]
    \centering
    \adjustbox{max width=\linewidth, center}{
        \begin{tabular}{lc@{\hskip 0.3cm}cccccc}
            \toprule
            \textbf{Distil.} & \textbf{Proj.} & \textbf{ADE20K} & \textbf{NYUd} & \textbf{MapFree} & \textbf{BEDLAM} \\
            \textbf{Data} & \textbf{Design} & (mIoU $\uparrow$) & (RMSE $\downarrow$) & (AUC $\uparrow$) & (PA-PVE $\downarrow$) \\
            \midrule
            IN-19K & LP & {42.4} & 0.446 & 91.4 & 83.9 \\
            IN-19K & TP  & {\bf 44.9} & 0.433 & 93.6 & 73.5 \\
            \midrule
            All    & {SP} & {42.3} & 0.413 & 92.2 & 73.1 \\
            All    & LP & {44.7} & 0.384 & 91.5 & 78.2 \\
            All    & TP  & {\bf 44.9} & {\bf 0.377} & {\bf 93.7} & {\bf 68.3} \\
            \bottomrule
        \end{tabular}
    }
    \vspace{-0.2cm}
    \caption{
        {\bf Distillation data and projector design}.
        We distill models using either ImageNet-19K as a generic dataset or all the 19 teacher datasets combined with full data sharing (\cref{sec:co-het}), employing either a Simple Projector (SP)~\cite{ranzinger2024radio}, a Ladder of Projectors (LP)~\cite{sariyildiz2024unic}, or the Transformer Projector (TP) presented in \cref{sec:projector_design}.
    }
    \label{tab:tp_vs_slp}
\end{table}

%% file: tex/float/tab_data_sharing.tex
\begin{table}[t!]
    \centering
    \adjustbox{max width=\linewidth, center}{
        \begin{tabular}{lccccc}
            \toprule
            \textbf{Data}    & \textbf{ADE20K} & \textbf{NYUd} & \textbf{MapFree} & \textbf{BEDLAM}  \\
            \textbf{Sharing} & (mIoU $\uparrow$) & (RMSE $\downarrow$) & (AUC $\uparrow$) & (PA-PVE $\downarrow$) \\
            \midrule
            No data sharing            & {41.6} &	0.426 &	93.2 & 68.7 \\
            Generic data sharing       & {40.1} & 0.416 & 92.7 & 71.7 \\
            Full data sharing          & {\bf 44.9} & {\bf 0.377} & {\bf 93.7} & {\bf 68.3} \\
            \bottomrule
        \end{tabular}
    }
    \vspace{-0.2cm}
    \caption{
            {\bf Data sharing among teachers.}
                We train three student models on all 19 datasets, using the data-sharing strategies outlined in \cref{sec:co-het}:
                \textit{No data sharing}: Teachers do not share any data.
                \textit{Generic data sharing}: ImageNet-19K is shared across all teachers.
                \textit{Full data sharing}: All images are shared among all teachers.
    }
    \label{tab:data_sharing}
\end{table}

%% file: tex/float/tab_ours_vs_baselines_teachers.tex
\begin{table*}[t!]
    \centering
    \adjustbox{max width=\linewidth, center}{
        \begin{tabular}{lcc@{}c@{\hskip 0.6cm}cccccc}
            \toprule
            \multirow{2}{*}{\textbf{Model}} & {\bf Encoder} & \textbf{Training} & {\bf Training}   & \textbf{ADE20k}            & \textbf{NYUd}                & \textbf{BEDLAM}                & \textbf{BEDLAM}                & \textbf{MapFree} \\
                           &   {\bf Arch.}           & {\bf Data}        & {\bf Res.} & (mIoU $\uparrow$) & (RMSE $\downarrow$) & (F1-score $\uparrow$) & (PA-PVE $\downarrow$) & (AUC $\uparrow$) \\
            \midrule
            \multicolumn{9}{l}{\textit{Teacher models}} \\
            ~\dino       & ViT-Large & LVD-142M & 518 & {47.7} & {0.384} & - & - & - \\
            ~\hmr        & ViT-Large & HMR-500K & 672 & - & - & {95} & {36.9} & - \\
            ~\master     & ViT-Large & \master-1.7M & 512 & - & - & - & - & {91.2} \\
            \midrule
            \multicolumn{9}{l}{\textit{State-of-the-art ViT-Base encoders}} \\
            ~\dino       & ViT-Base & LVD-142M & 518 & 47.3 & 0.399 & 86	& 76.5 & 89.6 \\
            ~\amradio    & ViT-Base & DataComp-1B & 512 & {\bf 50.0} & 0.718 & 89 & 83.2 & 93.1 \\
            \midrule
            ~\ours       & ViT-Base & \ours-20.7M & 336 & {44.9} & 0.377 & 91 & 68.3 & 93.7 \\
            ~\ours   & ViT-Base & \ours-20.7M  & 448 & {45.6} & \fbest{\bf 0.358}  & {\bf 94} & {\bf 56.0} & \fbest{\bf 94.7} \\
            \bottomrule
        \end{tabular}
    }
    \vspace{-0.2cm}
    \caption{
            {\bf Performance across 2D vision, 3D human understanding and 3D vison tasks with a \textit{universal} encoder.}
            The top section shows the performance of the teacher models we use, all of size ViT-Large.
            The middle section compares our encoder to two state-of-the-art ViT-Base encoders: \dino and the latest \amradio model.
            \textit{\ours-20.7M} is the heterogeneous collection of 19 public datasets we use for co-distillation.
            \fbest{\bf Colored} results highlight cases where our model outperforms \textit{the best ViT-Large teacher}.
    }
    \label{tab:ours_vs_baselines}
\end{table*}

%% file: tex/float/tab_mapfree.tex
\begin{table*}
    \centering
    \resizebox{\linewidth}{!}{
    \begin{tabular}{lcm{0.1cm}cccccccc}
    \toprule
    \multirow{2}{*}{Method} & \multirow{2}{*}{Encoder} & & \multicolumn{2}{c}{VCRE${<}$45px} & \multicolumn{2}{c}{VCRE${<}$90px} & Median Reproj. & \multicolumn{3}{c}{Relative Pose Error} \\
     & & & AUC$\uparrow$ & Prec.$\uparrow$ & AUC$\uparrow$ & Prec.$\uparrow$ & Error (px)$\downarrow$ & Median Error$\downarrow$ & Prec.$\uparrow$ & AUC$\uparrow$ \\
     \cmidrule(lr){1-1} \cmidrule(lr){2-2} \cmidrule(lr){4-5} \cmidrule(lr){6-7} \cmidrule(lr){8-8} \cmidrule(lr){9-11}
    LoFTR~\cite{loftr} & CNN && 39.7 & 18.2 & 61.8 & 33.5 & 166.8 & 2.31m ~39.4\textdegree & 26.9 & ~~9.8 \\
    DUSt3R~\cite{dust3r} & ViT-Large && 45.9 & 28.7 & 69.8 & 50.4 & 115.8 & 0.99m ~~~7.1\textdegree & 39.4 & 21.4 \\
    MicKey~\cite{mickey} & ViT-Large && 57.2 & 31.2 & 74.8 & 49.3 & 129.5 & 1.59m ~26.0\textdegree & 28.3 & 12.0 \\
    \master~\cite{mast3r} & ViT-Large && 81.7 & 63.0 & 93.3 & 79.3 & ~~48.8 & \textbf{0.37m ~~~2.2\textdegree} & 74.0 & 54.7 \\ \midrule
    \textbf{\ours} & \textbf{ViT-Base} && \textbf{84.0} & \textbf{64.4} & \textbf{94.3} & \textbf{81.1} & \textbf{~~47.4} & 0.39m ~~~4.6\textdegree &  \textbf{76.8} & \textbf{55.9} \\
     \bottomrule
    \end{tabular}
    }
    \vspace{-0.2cm}
    \caption{\textbf{Results on the map-free visual relocalization official leaderboard.} AUC and Precision (Prec.) are reported in percentages. \master~\cite{mast3r} on the leaderboard is a private version which performs {slightly} better than the released model that we use as teacher. \label{tab:mapfree}}
\end{table*}

%% file: tex/07_conclusions.tex
\section{Conclusions}
\label{sec:conclusions}

\looseness=-1
In this paper, we describe and tackle co-distillation, the challenging multi-teacher distillation task that arises when the set of teachers is composed of models of a very different nature, including generic teachers, whose features generalize well across tasks and domains by design, and teachers specialized for a certain task.
We applied co-distillation to \dino~\cite{oquab2024dinov2}, \hmr~\cite{multihmr}, and \master~\cite{mast3r}, whose training tasks, training data and properties are highly different.
This distillation process produces a strong encoder, \ours, which, when combined with each teacher’s task-specific decoder, performs on par with or surpasses the teachers.
Notably, our encoder outperforms \master on the Map-free Visual Relocalization dataset while using significantly fewer parameters.

%% file: tex/99_supp.tex
\section{Training protocol details}\label{sec:eval_protocol}

\subsection{Datasets}

The 19 datasets that we use for co-distillation are listed in~\Cref{tab:datasets}, and a few examples per dataset are provided in~\cref{fig:datasets_1,fig:datasets_2}.
As can be seen from~\cref{tab:datasets}, the datasets are quite unbalanced in size.
During training, we construct a batch such that it contains an equal amount of randomly sampled images
{from the datasets associated with each teacher}, \ie \dino, \hmr and \master.

\input{tex/float/supplementary/tab_datasets}

\paragraph{Access to teacher training data.}
For presentation clarity and without loss of generality, in the main paper we assume that all the data used to train all the teachers is also available for distillation. This is in practice impossible at times, either because a subset of the dataset might not be public, or because of their size. In such cases, one can use only the subset of the datasets that is available, or source alternative data across all domains. This extends beyond distillation to the data used for finetuning.

\paragraph{Sample images from all datasets.}
In~\cref{fig:datasets_1,fig:datasets_2}, we visualize 10 randomly sampled images from each dataset listed in~\cref{tab:datasets}.

\subsection{Table of hyper-parameters}
\label{sec:imp_details}

A table with the set of hyper-parameters we use for training our \ours models are given in~\Cref{tab:hyperparameters}.
Further details can be found on github.

\input{tex/float/tab_hyperparameters}

\begingroup
\renewcommand{\vs}{\bm{s}}
\paragraph{Distillation loss.}
Following~\cite{sariyildiz2024unic}, given an image $x$, we minimize the combination of the cosine and smooth-$\ell_1$ losses between the outputs of student $\vs_i = h_i(f(x))$ and each teacher $\vt_i = t_i(x)$:

\begin{equation}
    \mathcal{L}_{\text{distil}} = \sum_{i=1}^N \mathcal{L}_{cos}(\vs_i, \vt_i) + \mathcal{L}_{s\ell_1}(\vs_i, \vt_i),
\end{equation}
where
\begin{align}
    \gL_{cos} (\vs, \vt) &= 1 - \frac{\vs \cdot \vt}{||\vs||_2 \times ||\vt||_2}, \\
    \gL_{sl1} (\vs, \vt) &=
\begin{cases}
    0.5 \times ||\vs - \vt||_2^2, & \text{for } ||\vs - \vt||_1 < 1, \\
    ||\vs - \vt||_1 - 0.5, & \text{otherwise}.
\end{cases}
\end{align}
\endgroup

\section{Details on decoder fine-tuning}\label{sec:dec_finetuning}

\paragraph{\master.}
\master relies on a binocular architecture with a Siamese ViT-encoder to encode the input images, followed by binocular decoders and prediction head. When finetuning this model, we simply replace the encoder and keep it frozen using the publicly available code of \master~\cite{mast3r}. Given the size of the decoders and heads, we initialize them with the released models, except for weights that have a mismatch of size, namely the fully-connected layer between the encoder and decoder, as our ViT-Base encoder has a smaller feature dimension than their ViT-Large one (768 \vs 1024) as well as the output layers that outputs a pixelwise prediction due the mismatch of patch sizes (14 \vs 16).
We finetune the model on 6.5M image pairs with AdamW on images at different resolutions.
For backbone distilled on $336{\times}336$ images, we use $\{ 448{\times}448$, $448{\times}336$, $448{\times}294$, $448{\times}252$, $448{\times}224$, $448{\times}140\}$, which corresponds to the same number of patches as \master's setting.
For backbone further distilled on $448{\times}448$ images, we use $\{518{\times}518$, $518{\times}392$, $518{\times}336$, $518{\times}294$, $518{\times}252$, $518{\times}168 \}$ which corresponds to the resolutions close to the ones from \master but that are multiple of $14$.

\paragraph{\hmr.}
To evaluate our model on the task of Human Mesh Recovery (HMR), we use the training framework and public code of \hmr~\cite{multihmr}.
We discard the projector modules and freeze the weights of the distilled student model.
The Human Perception Head (HPH) proposed in Multi-HMR is used to predict HMR from the outputs of the backbone, with two transformer blocks prepended to it.
This head is trained from scratch on the BEDLAM dataset, using images at a resolution of $672{\times}672$.
Training is done with a learning rate of $4e-5$, a batch size of 16, and a cosine decay schedule over 200k iterations.
After training, evaluation is performed on the BEDLAM validation set with a non-maximum suppression (NMS) kernel of size 3 and a detection threshold of 0.3, following the Multi-HMR protocol.

Notably, this evaluation procedure favors the teacher model, as its native resolution is $672{\times}672$, whereas the student model is distilled on images of resolution $448{\times}448$ only due to computational constraints.

\paragraph{Semantic segmentation and depth estimation evaluations.}
Semantic segmentation and depth estimation are dense prediction tasks, both formulated as classification tasks in this work, and solved following the {simple} setup proposed in \cite{oquab2024dinov2}, also followed by the most recent related works~\cite{ranzinger2024radio,sariyildiz2024unic}.
We extract the tokens from the last output layer of the student model and use as input to a linear prediction head.
For semantic segmentation, we additionally use the Transformer Projector of the \dino teacher as part of the frozen encoder, and train a linear head on top of the projector.
to predict class logits from a patch token.
This yields a $32{\times}32$ logit map that is upsampled via bilinear interpolation to the original image resolution of $512{\times}512$.

For depth estimation, we first upsample patch features by a factor of $4$ via bilinear interpolation, concatenate them along the feature dimension with the CLS token, and use these vectors as input to a linear layer. Depth prediction is treated as a soft classification task following~\cite{adabins}; we use 256 uniformly distributed bins.

\section{Attention map visualizations}\label{sec:attention}

In Fig.~2 of the main paper, we present a visualization of the encoder outputs from the teacher models and our student model using principal component analysis (PCA).
This analysis is conducted on three randomly selected images from the Map-free and BEDLAM datasets.
The visualization reveals that patch similarity patterns differ across the teacher models, while our student model appears to simultaneously attempt to capture and integrate multiple patterns from the different teachers.

To further investigate this phenomenon, we visualize in~\Cref{fig:attention} the attention probabilities obtained at the last encoder layer of the student model, as well as those of the three teacher-specific Transformer Projectors (TP) attached on top during distillation.
More concretely, given an image of size $448\times448$, we extract the $32\times32$ attention map for all the $1024$ patches (the patch size for the student model is 14).
In order to see the most prototypical attention patterns, we flatten all patch attentions and cluster them via $k$-Medoids ($k=9$), with the version available in Scikit-Learn.\footnote{\url{https://scikit-learn-extra.readthedocs.io/}}

We indeed observe different attention patterns for the last encoder block and the Transformer projectors.
For instance, the projector for \master yields much more localized attentions regardless of the input image compared to the projector for \dino, whose attentions have much wider spatial extent. We also notice that the projector for \hmr focuses mainly on the human, when there is one in the image (see~\Cref{fig:attention}).

Looking at the attentions of the last layer of the encoder, however, we observe once again that it seems to try to capture a mixture of the attentions of the three projectors: They exhibit a strong locality as in \master, a spatial extent similar to \dino, and also a strong preference for humans.

\section{Additional Results}

In this section, we provide additional evaluations for \ours models.
We report results for \master with a \ours encoder on multi-view depth estimation and camera pose regression tasks, as well as semantic segmentation performance on additional datasets and comparisons to 2D-to-3D distillation methods.
Furthermore, we evaluate our models on Feat2GS, a recently proposed benchmark for assessing models' 3D awareness in geometry and texture via novel view synthesis, and present extended qualitative results.

\subsection{Multi-view depth evaluation}
We follow the protocol of~\cite{mvd} and evaluate multi-depth stereo depth evaluation on KITTI~\cite{kitti}, DTU~\cite{dtu}, ETH3D~\cite{eth3d}, Tanks And Temples~\cite{tankstemples} and ScanNet~\cite{scannet}. We report the Absolute Relative Error (rel) and the Inlier Ratio ($\tau$) with a threshold of $1.03$ on each test set, as well as the averages over all test sets.
To extract depth prediction of one image, we follow DUSt3R~\cite{dust3r} and extract depthmaps as the z-coordinate of the predicted pointmaps; and when multiple pointmaps are available for one image from different image pairs, we simply rescale the predicted depthmaps and average them with weights given by the predicted confidence values. Results are reported in Table~\ref{tab:mvd}. \ours performs similarly to MASt3R and DUSt3R on this task overall, while using a smaller ViT-Base image encoder.

\begin{table*} [t]
    \centering
    \adjustbox{max width=\linewidth}{
    \begin{tabular}{ll|c@{~~~~}cc@{~~~~}cc@{~~~~}cc@{~~~~}cc@{~~~~}cc@{~~~~}c}
    \toprule
    \multirow{2}{*}{\textbf{Method}} & \multirow{2}{*}{\textbf{Encoder}} &  \multicolumn{2}{c}{\textbf{KITTI}}             & \multicolumn{2}{c}{\textbf{ScanNet}}           & \multicolumn{2}{c}{\textbf{ETH3D}}             & \multicolumn{2}{c}{\textbf{DTU}}               & \multicolumn{2}{c}{\textbf{T\&T}}              & \multicolumn{2}{c}{\textbf{Average}}           \\
                   && rel. $\downarrow$ & $\tau$ $\uparrow$ & rel. $\downarrow$ & $\tau$ $\uparrow$ & rel. $\downarrow$ & $\tau$ $\uparrow$ & rel. $\downarrow$ & $\tau$ $\uparrow$ & rel. $\downarrow$ & $\tau$ $\uparrow$ & rel. $\downarrow$ & $\tau$ $\uparrow$ \\ \midrule
    DeepV2D~\cite{deepv2d} & Hourglass  & 10.00              & 36.20             & 4.40              & 54.80             & 11.80              & 29.30             & 7.70              & 33.00             & 8.90             & 46.40            & 8.60              & 39.90             \\
    DUSt3R~\cite{dust3r} & ViT-Large & ~5.88 & 47.67 & \bf{3.01} & \bf{72.54} & 3.04 & 75.17 & \underline{2.92} & \underline{73.94} & 2.93 & 78.51 & 3.56 & \underline{69.56} \\
    MASt3R~\cite{mast3r} & ViT-Large & ~\bf{3.54} & \bf{65.68} & \underline{4.17} & \underline{65.22} & \bf{2.44} & \bf{82.77} & 3.46 & 66.89 & \bf{2.04} & \bf{87.88} & \bf{3.13} & \bf{73.69} \\
    \textbf{\ours} & ViT-Base & ~\underline{4.88} & \underline{50.76} & 4.24 & 59.68 & \underline{2.48} & \underline{77.97} & \bf{2.69} & \bf{75.63} & \underline{2.60} & \underline{79.19} & \underline{3.38} & 68.65 \\
    \bottomrule
    \end{tabular}
    }
    \vspace{-0.2cm}
    \caption{\textbf{Multi-view depth evaluation} with the absolute relative error (rel) and the inlier ratio ($\tau$) on several test sets, and the average across all test sets in the last column. DeepV2D uses ScanNet in the training set, explaining its better performance on this dataset.  \ours uses a ViT-Base encoder while DUSt3R and MASt3R a ViT-Large encoder.}
    \label{tab:mvd}
\end{table*}

\subsection{Multi-view camera pose regression evaluation}

Following the protocol of~\cite{posediffusion,mast3r}, we evaluate on the task of multi-view pose estimation on the CO3Dv2~\cite{co3d} and RealEstate10K~\cite{realestate10K} datasets using sequences of 10 images.
Matches obtained as output of the \master decoder and head for an image pair are used to estimate Essential Matrices and relative pose.
We report the Relative Rotation Accuracy (RRA) and Relative Translation Accuracy (RTA) on image pairs at a threshold of 15\textdegree, as well as the mean Average Accuracy (mAA30), \ie, the area under the accuracy curve of the angular differences (RRA@30, RTA@30). Results are reported in Table~\ref{tab:mvpose}. \ours performs on par with DUSt3R and MASt3R on the object-centric Co3Dv2 dataset, while it outperforms them on the more challenging RealEstate10K dataset. Once again, \ours uses a ViT-Base encoder while DUSt3R and MASt3R are based on a ViT-Large encoder.

\begin{table*}[t]
    \setlength{\tabcolsep}{4pt}
    \centering
    \adjustbox{max width=0.65\linewidth}{
        \begin{tabular}{ll|ccc|c}
        \toprule
        \multirow{2}{*}{\textbf{Method}} &  \multirow{2}{*}{\textbf{Encoder}} & \multicolumn{3}{c|}{\textbf{Co3Dv2} $\uparrow$} & \textbf{RealEstate10K} $\uparrow$ \\

        & & RRA@15 & RTA@15 & mAA(30) & mAA(30) \\
        \midrule
        DUSt3R~\cite{dust3r} & ViT-Large & \underline{93.3} & 88.4 & 77.2 & 61.2 \\
        \master~\cite{mast3r} & ViT-Large & {\bf 94.6} &  {\bf 91.9} & {\bf 81.8} & \underline{76.4} \\
        \textbf{\ours} & ViT-Base              & 92.2 & \underline{90.7} & \underline{78.8} & {\bf 79.9} \\
        \bottomrule
        \end{tabular}
    }
    \vspace{-0.2cm}
    \caption{\textbf{Multi-view pose regression evaluation} on the CO3Dv2~\cite{co3d}  and RealEstate10K~\cite{realestate10K} datasets with 10 random frames. \ours uses a ViT-Base encoder while DUSt3R and MASt3R a ViT-Large encoder.}
    \label{tab:mvpose}
\end{table*}

\begin{figure}[t]
    \centering
    \includegraphics[width=\linewidth]{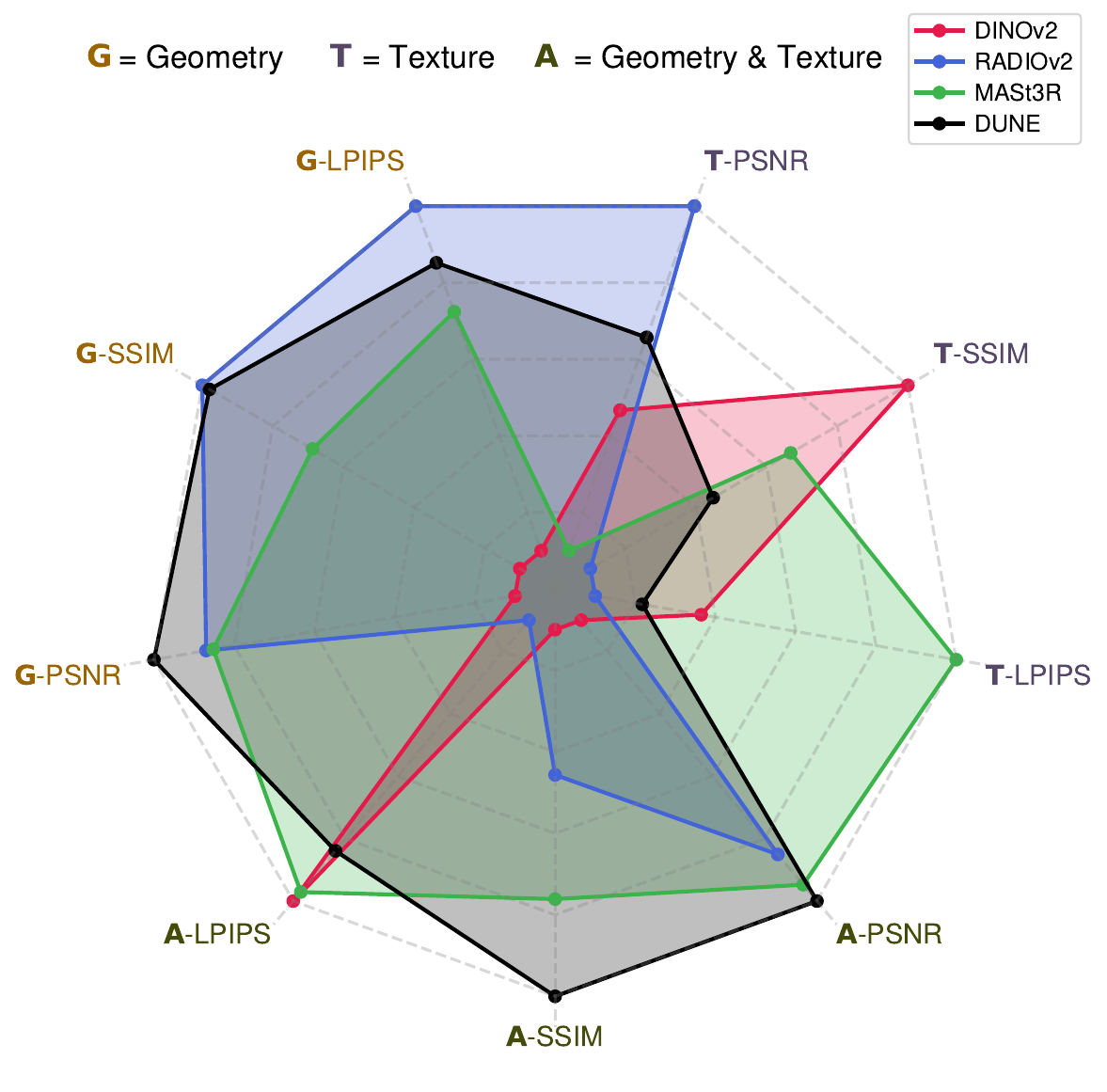}
    \vspace{-0.9cm}
    \caption{
        \textbf{Evaluating \ours on the Feat2GS benchmark.}
        The spider plot shows comparisson of different encoder models.
        The Feat2GS benchmark \cite{feat2gs} evaluates Novel View Synthesis as a proxy for 3D awareness.
        In all metrics, larger distance to the center indicates better performance.
        Note that models vary in size: RADIOv2 is a ViT-H, \master a ViT-L and \dino and \ours a ViT-B.}
    \label{fig:feat2gs}
\end{figure}

\subsection{Evaluating \ours on the Feat2GS benchmark}

In \cref{fig:feat2gs} we compare different encoder models in the Feat2GS benchmark \cite{feat2gs}.
The Feat2GS benchmark has three modalities, \textit{i)} \textbf{G}eometry: When only geometry parameters are predicted from features and texture is free-optimized for Novel View Synthesis.
\textit{ii)}  \textbf{T}exture: When only the texture is predicted from encoder features and the geometry is free-optimized.
And \textit{iii)} \textbf{A}ll: When both geometry and texture are predicted from features.
Our \ours encoder leads to the best performance when \textbf{A}ll the parameters are predicted from features (to our understanding the most challenging setting) and leads to the largest area over all settings and metrics.
For more detailed results, we also present \cref{tab:feat2gs} with per-dataset evaluations of all metrics and modalities.
While all encoders use a ViT architecture, they vary significantly in size, mainly due to the absence of ViT-B models for certain methods.
Namely, RADIOv2 only has a ViT-H model open-sourced and \master a ViT-L, \dino and our model \ours are ViT-B.
Thus, the fact \ours is obtaining the overall best performance compared to much larger models is even more remarkable.

\input{tex/float/tab_semseg_extra}

\input{tex/float/tab_depth_extra}

\input{tex/float/supplementary/table_feat2gs}

\subsection{Comparison to 3D-to-2D distillation and 3D-uplifting methods}

In~\Cref{tab:semseg_extra}, we report semantic segmentation evaluations on three datasets, comparing \ours to Pri3D~\cite{hou2021pri3d} (a 3D-to-2D distillation method) and \master.
For \ours, we present results using the encoder outputs directly, \ours (no proj.), and with the \dino projector applied after the encoder, \ours.
In all cases, only a linear layer is trained to predict patch labels.
\ours significantly outperforms both Pri3D and \master.

In~\Cref{tab:depth_extra}, we evaluate depth estimation performance on NYUv2, comparing \ours to FiT-3D~\cite{yue2024fit3d}, a recent method that enhances \dino features for 3D tasks.
\ours achieves substantially better performance than FiT-3D.

\subsection{Qualitative comparisons of teacher outputs to \ours}
\label{sec:extended_quals}

\paragraph{\master.}
In~\Cref{fig:mast3r_qualitative_1,fig:mast3r_qualitative_2,fig:mast3r_qualitative_many} we present qualitative results for \master and our student side-by-side.
We see that the student clearly improves over the teacher in some cases.

\paragraph{\hmr.}
In~\Cref{fig:hmr1,fig:hmr2} we present qualitative results for images randomly sampled from the bedlam validation set, comparing the outputs of the student and teacher.
Both models achieve results of comparable visual quality.

\input{tex/float/fig_attention}

\input{tex/float/supplementary/fig_mast3r}

\input{tex/float/supplementary/fig_hmr}

\input{tex/float/supplementary/fig_dataset_samples}

%% file: tex/float/supplementary/tab_datasets.tex
\begin{table}[ht]
    \centering
    \adjustbox{max width=\linewidth}{
    \begin{tabular}{lrlc}
        \toprule
        \textbf{Name} & \textbf{Size} & \textbf{Nature} & \textbf{Teacher} \\
        \toprule
        ImageNet-19K~\cite{deng2009imagenet,yang2020fda} & 13,153,480 & Real & \multirow{3}{*}{DINO-v2} \\
        Mapillary~\cite{mapillary} & 1,205,907 & Real & \\
        Google Landmarks v2~\cite{google-landmarks-v2} & 4,132,914 & Real & \\
        \midrule
        Habitat~\cite{habitat19iccv} & 284,968 & Rendered & \multirow{12}{*}{MAST3R} \\
        ARKitScenes~\cite{arkitscenes} & 456,108 & Rendered & \\
        Blended MVS~\cite{blendedMVS} & 98,937 & Rendered & \\
        MegaDepth~\cite{megadepth} & 36,949 & Real & \\
        ScanNet++~\cite{scannet++} & 60,188 & Rendered & \\
        CO3D-v2~\cite{co3d} & 185,100 & Real & \\
        Map-free~\cite{mapfree} & 41,300 & Real & \\
        WildRgb~\cite{wildrgb} & 224,400 & Real & \\
        VirtualKitti~\cite{vkitti2} & 1,200 & Synthetic &  \\
        Unreal4K~\cite{unreal4k} & 14,386 & Synthetic & \\
        TartanAir~\cite{tartanair2020iros} & 136,225 & Real & \\
        DL3DV~\cite{ling2024dl3dv} & 208,800 & Rendered & \\
        \midrule
        BEDLAM~\cite{bedlam} & 353,118 & Synthetic &  \multirow{4}{*}{Multi-HMR} \\
        AGORA~\cite{agora} & 14,314 & Synthetic & \\
        CUFFS~\cite{multihmr} & 54,944 & Synthetic & \\
        UBody~\cite{ubody} & 54,234 & Real & \\
        \midrule
        \multicolumn{1}{r}{\textbf{Total size:}} & \textbf{20,717,472} \\
        \bottomrule
    \end{tabular}
    }
    \vspace{-0.2cm}
    \caption{
        {\bf Datasets used for training \ours models.}
        The teacher column groups the datasets which are
        {associated with}
        each teacher.
    }
    \label{tab:datasets}
\end{table}

%% file: tex/float/tab_hyperparameters.tex
\begin{table}[h]
    \centering
    \adjustbox{max width=\linewidth}{
    \begin{tabular}{l|l}
        \toprule
        Hyper-parameter & Value \\
        \toprule
        \multirow{6}{*}{Encoder} & Architecture: ViT-Base \\
                                 & Patch size: 14 \\
                                 & Num. registers: 0 \\
                                 & QKV bias: True \\
                                 & LayerScale: True \\
                                 & Path drop rate: 0 \\ \hline
        \multirow{3}{*}{Projector} & Architecture: TP \\
                                   & Num. blocks: 1 \\
                                   & Block configuration follows encoder \\ \hline
        Image & Initial: $336 \times 336$ \\
        resolution & Fine-tuned: $448 \times 448$ \\  \hline
        Batch size & 128 per GPU \\ \hline
        Num. GPUs & 4 \\ \hline
        \multirow{3}{*}{Optimizer} & Type: AdamW \\
                                   & Weight decay: $3e-2$ \\
                                   & $(\beta_1, \beta_2)$: $(0.9, 0.99)$ \\ \hline
        \multirow{3}{*}{Learning rate} & Min: $1e-6$ \\
                                       & Max: $3e-4 \times \text{batch-size} / 256$ \\
                                       & Schedule: Cosine \\ \hline
        Data type & AMP with bloat16 \\ \hline
        Training data             & \multirow{2}{*}{All (\ours-20.7M, see~\Cref{tab:datasets})} \\
        (Tab~1 in the main paper) & \\ \hline
        Data sharing              & \multirow{2}{*}{Full data sharing} \\
        (Tab~2 in the main paper) & \\ \hline
        Training budget & $1,281,167 \times 100$ images \\
        \bottomrule
    \end{tabular}
    }
    \vspace{-0.2cm}
    \caption{\bf Hyper-parameters used for training \ours models.}
    \label{tab:hyperparameters}
\end{table}

%% file: tex/float/tab_semseg_extra.tex
{
\setlength{\tabcolsep}{2pt}
\begin{table}[t]
    \centering
    \adjustbox{max width=\linewidth}{
    \begin{tabular}{l c c c c}
        \toprule
         \multirow{2}{*}{Model} & Cityscapes        & NYUv2             & ScanNet             & Avg.  \\
               & (mIoU $\uparrow$) & (mIoU $\uparrow$) & (mIoU $\uparrow$)   & (mIoU $\uparrow$) \\
         \midrule
         Pri3D~\cite{hou2021pri3d} & 56.3 & 54.8 & 61.7 & 57.6 \\
         \master~\cite{mast3r} & 58.9 & 60.2 & 57.0 &  58.7 \\
         {\ours (no proj.)} & 65.6 &	66.1 & 61.2 &	64.3 \\
         {\ours} & {\bf 70.6} & {\bf 68.2} & {\bf 65.2} & {\bf 68.0} \\
         \bottomrule
    \end{tabular}
    }
    \vspace{-0.25cm}
    \caption{
        {\bf Additional semantic segmentation evaluations.}
        {As described in the paper, for improved segmentation performance we can use the DINO teacher projector as part of the frozen encoder, and learn a linear classifier on top. }
    }
    \label{tab:semseg_extra}
\end{table}
}

%% file: tex/float/tab_depth_extra.tex
\begin{table}[t]
    \centering
    \adjustbox{max width=\linewidth}{
    \begin{tabular}{l c c c}
        \toprule
        \multirow{2}{*}{Model} & NYUv2 \\
               & (RMSE $\downarrow$) \\
         \midrule
         FiT-3D~\cite{yue2024fit3d} & 0.380 \\
         \ours & 0.358 \\
         \bottomrule
    \end{tabular}
    }
    \vspace{-0.2cm}
    \caption{{\bf Comparisson to Fit-3D} on monocular depth.}
    \label{tab:depth_extra}
\end{table}

%% file: tex/float/supplementary/table_feat2gs.tex
\begin{table*}[h]
\setlength{\tabcolsep}{2pt}
\centering
\resizebox{\textwidth}{!}{
\begin{tabular}{l|>{\raggedleft\arraybackslash}p{0.9cm}>{\raggedleft\arraybackslash}p{0.9cm}>{\raggedleft\arraybackslash}p{0.9cm}|>{\raggedleft\arraybackslash}p{0.9cm}>{\raggedleft\arraybackslash}p{0.9cm}>{\raggedleft\arraybackslash}p{0.9cm}|>{\raggedleft\arraybackslash}p{0.9cm}>{\raggedleft\arraybackslash}p{0.9cm}>{\raggedleft\arraybackslash}p{0.9cm}|>{\raggedleft\arraybackslash}p{0.9cm}>{\raggedleft\arraybackslash}p{0.9cm}>{\raggedleft\arraybackslash}p{0.9cm}|>{\raggedleft\arraybackslash}p{0.9cm}>{\raggedleft\arraybackslash}p{0.9cm}>{\raggedleft\arraybackslash}p{0.9cm}|>{\raggedleft\arraybackslash}p{0.9cm}>{\raggedleft\arraybackslash}p{0.9cm}>{\raggedleft\arraybackslash}p{0.9cm}|>{\raggedleft\arraybackslash}p{0.9cm}>{\raggedleft\arraybackslash}p{0.9cm}>{\raggedleft\arraybackslash}p{0.9cm}|>{\raggedleft\arraybackslash}p{0.9cm}>{\raggedleft\arraybackslash}p{0.9cm}>{\raggedleft\arraybackslash}p{0.9cm}|>{\raggedleft\arraybackslash}p{0.9cm}>{\raggedleft\arraybackslash}p{0.9cm}>{\raggedleft\arraybackslash}p{0.9cm}}
\toprule
\multicolumn{1}{c|}{} & \multicolumn{9}{c|}{LLFF} & \multicolumn{9}{c|}{DL3DV} & \multicolumn{9}{c}{Casual} \\
\midrule
\multicolumn{1}{c|}{} & \multicolumn{3}{c|}{\textbf{G}eometry} & \multicolumn{3}{c|}{\textbf{T}exture} & \multicolumn{3}{c|}{\textbf{A}ll} & \multicolumn{3}{c|}{\textbf{G}eometry} & \multicolumn{3}{c|}{\textbf{T}exture} & \multicolumn{3}{c|}{\textbf{A}ll} & \multicolumn{3}{c|}{\textbf{G}eometry} & \multicolumn{3}{c|}{\textbf{T}exture} & \multicolumn{3}{c}{\textbf{A}ll} \\
\midrule
Feature & \fontsize{8.5pt}{9pt}\selectfont{PSNR$\uparrow$} & \fontsize{8.5pt}{9pt}\selectfont{SSIM$\uparrow$} & \fontsize{8.5pt}{9pt}\selectfont{LPIPS$\downarrow$} & \fontsize{8.5pt}{9pt}\selectfont{PSNR$\uparrow$} & \fontsize{8.5pt}{9pt}\selectfont{SSIM$\uparrow$} & \fontsize{8.5pt}{9pt}\selectfont{LPIPS$\downarrow$} & \fontsize{8.5pt}{9pt}\selectfont{PSNR$\uparrow$} & \fontsize{8.5pt}{9pt}\selectfont{SSIM$\uparrow$} & \fontsize{8.5pt}{9pt}\selectfont{LPIPS$\downarrow$} & \fontsize{8.5pt}{9pt}\selectfont{PSNR$\uparrow$} & \fontsize{8.5pt}{9pt}\selectfont{SSIM$\uparrow$} & \fontsize{8.5pt}{9pt}\selectfont{LPIPS$\downarrow$} & \fontsize{8.5pt}{9pt}\selectfont{PSNR$\uparrow$} & \fontsize{8.5pt}{9pt}\selectfont{SSIM$\uparrow$} & \fontsize{8.5pt}{9pt}\selectfont{LPIPS$\downarrow$} & \fontsize{8.5pt}{9pt}\selectfont{PSNR$\uparrow$} & \fontsize{8.5pt}{9pt}\selectfont{SSIM$\uparrow$} & \fontsize{8.5pt}{9pt}\selectfont{LPIPS$\downarrow$} & \fontsize{8.5pt}{9pt}\selectfont{PSNR$\uparrow$} & \fontsize{8.5pt}{9pt}\selectfont{SSIM$\uparrow$} & \fontsize{8.5pt}{9pt}\selectfont{LPIPS$\downarrow$} & \fontsize{8.5pt}{9pt}\selectfont{PSNR$\uparrow$} & \fontsize{8.5pt}{9pt}\selectfont{SSIM$\uparrow$} & \fontsize{8.5pt}{9pt}\selectfont{LPIPS$\downarrow$} & \fontsize{8.5pt}{9pt}\selectfont{PSNR$\uparrow$} & \fontsize{8.5pt}{9pt}\selectfont{SSIM$\uparrow$} & \fontsize{8.5pt}{9pt}\selectfont{LPIPS$\downarrow$} \\
\midrule
\dino  &             \cellcolor[rgb]{0.85,0.88,0.67}19.69 &             \cellcolor[rgb]{1.00,0.82,0.82}.7405 &                \cellcolor[rgb]{1.00,0.82,0.82}.2148 &             \cellcolor[rgb]{1.00,0.90,0.82}18.79 &             \cellcolor[rgb]{0.56,0.75,0.38}.7173 &                \cellcolor[rgb]{0.56,0.75,0.38}.2179 &             \cellcolor[rgb]{0.56,0.75,0.38}19.90 &             \cellcolor[rgb]{1.00,0.90,0.82}.7257 &                \cellcolor[rgb]{0.56,0.75,0.38}.2521 &             \cellcolor[rgb]{1.00,0.82,0.82}18.24 &             \cellcolor[rgb]{1.00,0.82,0.82}.7042 &                \cellcolor[rgb]{1.00,0.82,0.82}.3427 &             \cellcolor[rgb]{1.00,0.82,0.82}17.00 &             \cellcolor[rgb]{0.56,0.75,0.38}.6605 &                \cellcolor[rgb]{0.85,0.88,0.67}.3382 &             \cellcolor[rgb]{0.85,0.88,0.67}18.15 &             \cellcolor[rgb]{1.00,0.82,0.82}.7138 &                \cellcolor[rgb]{0.85,0.88,0.67}.3556 &             \cellcolor[rgb]{1.00,0.82,0.82}19.28 &             \cellcolor[rgb]{1.00,0.82,0.82}.6557 &                \cellcolor[rgb]{1.00,0.82,0.82}.3613 &             \cellcolor[rgb]{0.56,0.75,0.38}17.86 &             \cellcolor[rgb]{0.85,0.88,0.67}.5871 &                \cellcolor[rgb]{1.00,0.90,0.82}.3571 &             \cellcolor[rgb]{1.00,0.82,0.82}19.15 &             \cellcolor[rgb]{1.00,0.82,0.82}.6654 &                \cellcolor[rgb]{0.56,0.75,0.38}.3877 \\
RADIOv2 &             \cellcolor[rgb]{1.00,0.82,0.82}19.66 &             \cellcolor[rgb]{1.00,0.90,0.82}.7454 &                \cellcolor[rgb]{1.00,0.90,0.82}.2121 &             \cellcolor[rgb]{0.56,0.75,0.38}18.85 &             \cellcolor[rgb]{1.00,0.82,0.82}.7137 &                \cellcolor[rgb]{1.00,0.82,0.82}.2215 &             \cellcolor[rgb]{1.00,0.90,0.82}19.83 &             \cellcolor[rgb]{1.00,0.82,0.82}.7139 &                \cellcolor[rgb]{1.00,0.82,0.82}.3048 &             \cellcolor[rgb]{1.00,0.90,0.82}18.27 &             \cellcolor[rgb]{0.85,0.88,0.67}.7092 &                \cellcolor[rgb]{0.56,0.75,0.38}.3296 &             \cellcolor[rgb]{0.56,0.75,0.38}17.04 &             \cellcolor[rgb]{1.00,0.82,0.82}.6582 &                \cellcolor[rgb]{1.00,0.82,0.82}.3400 &             \cellcolor[rgb]{1.00,0.82,0.82}18.00 &             \cellcolor[rgb]{1.00,0.90,0.82}.7159 &                \cellcolor[rgb]{1.00,0.82,0.82}.3687 &             \cellcolor[rgb]{1.00,0.90,0.82}19.50 &             \cellcolor[rgb]{0.85,0.88,0.67}.6646 &                \cellcolor[rgb]{0.56,0.75,0.38}.3372 &             \cellcolor[rgb]{1.00,0.90,0.82}17.76 &             \cellcolor[rgb]{1.00,0.82,0.82}.5826 &                \cellcolor[rgb]{1.00,0.82,0.82}.3580 &             \cellcolor[rgb]{1.00,0.90,0.82}19.43 &             \cellcolor[rgb]{0.85,0.88,0.67}.6698 &                \cellcolor[rgb]{1.00,0.82,0.82}.4034 \\
\master  &             \cellcolor[rgb]{0.56,0.75,0.38}19.74 &             \cellcolor[rgb]{0.85,0.88,0.67}.7477 &                \cellcolor[rgb]{0.85,0.88,0.67}.2061 &             \cellcolor[rgb]{0.56,0.75,0.38}18.85 &             \cellcolor[rgb]{0.85,0.88,0.67}.7169 &                \cellcolor[rgb]{0.85,0.88,0.67}.2181 &             \cellcolor[rgb]{0.85,0.88,0.67}19.84 &             \cellcolor[rgb]{0.56,0.75,0.38}.7269 &                \cellcolor[rgb]{0.85,0.88,0.67}.2588 &             \cellcolor[rgb]{0.85,0.88,0.67}18.30 &             \cellcolor[rgb]{0.56,0.75,0.38}.7102 &                \cellcolor[rgb]{1.00,0.90,0.82}.3347 &             \cellcolor[rgb]{0.56,0.75,0.38}17.04 &             \cellcolor[rgb]{0.85,0.88,0.67}.6602 &                \cellcolor[rgb]{0.56,0.75,0.38}.3373 &             \cellcolor[rgb]{1.00,0.90,0.82}18.05 &             \cellcolor[rgb]{0.85,0.88,0.67}.7161 &                \cellcolor[rgb]{0.56,0.75,0.38}.3538 &             \cellcolor[rgb]{0.56,0.75,0.38}19.65 &             \cellcolor[rgb]{1.00,0.90,0.82}.6594 &                \cellcolor[rgb]{1.00,0.90,0.82}.3459 &             \cellcolor[rgb]{1.00,0.82,0.82}17.71 &             \cellcolor[rgb]{0.56,0.75,0.38}.5968 &                \cellcolor[rgb]{0.56,0.75,0.38}.3369 &             \cellcolor[rgb]{0.56,0.75,0.38}19.60 &             \cellcolor[rgb]{1.00,0.90,0.82}.6691 &                \cellcolor[rgb]{0.85,0.88,0.67}.3882 \\
\ours    &             \cellcolor[rgb]{0.85,0.88,0.67}19.69 &             \cellcolor[rgb]{0.56,0.75,0.38}.7499 &                \cellcolor[rgb]{0.56,0.75,0.38}.2041 &             \cellcolor[rgb]{1.00,0.82,0.82}18.75 &             \cellcolor[rgb]{1.00,0.90,0.82}.7147 &                \cellcolor[rgb]{1.00,0.90,0.82}.2200 &             \cellcolor[rgb]{1.00,0.82,0.82}19.70 &             \cellcolor[rgb]{0.85,0.88,0.67}.7261 &                \cellcolor[rgb]{1.00,0.90,0.82}.2689 &             \cellcolor[rgb]{0.56,0.75,0.38}18.33 &             \cellcolor[rgb]{1.00,0.90,0.82}.7088 &                \cellcolor[rgb]{0.85,0.88,0.67}.3337 &             \cellcolor[rgb]{1.00,0.90,0.82}17.02 &             \cellcolor[rgb]{1.00,0.90,0.82}.6595 &                \cellcolor[rgb]{1.00,0.90,0.82}.3392 &             \cellcolor[rgb]{0.56,0.75,0.38}18.18 &             \cellcolor[rgb]{0.56,0.75,0.38}.7185 &                \cellcolor[rgb]{1.00,0.90,0.82}.3571 &             \cellcolor[rgb]{0.85,0.88,0.67}19.51 &             \cellcolor[rgb]{0.56,0.75,0.38}.6665 &                \cellcolor[rgb]{0.85,0.88,0.67}.3445 &             \cellcolor[rgb]{0.85,0.88,0.67}17.81 &             \cellcolor[rgb]{1.00,0.90,0.82}.5835 &                \cellcolor[rgb]{0.85,0.88,0.67}.3569 &             \cellcolor[rgb]{0.85,0.88,0.67}19.56 &             \cellcolor[rgb]{0.56,0.75,0.38}.6728 &                \cellcolor[rgb]{1.00,0.90,0.82}.3894 \\
\midrule
\multicolumn{1}{c|}{} & \multicolumn{9}{c|}{MipNeRF 360} & \multicolumn{9}{c|}{MVImgNet} & \multicolumn{9}{c}{Tanks and Temples} \\
\midrule
\multicolumn{1}{c|}{} & \multicolumn{3}{c|}{\textbf{G}eometry} & \multicolumn{3}{c|}{\textbf{T}exture} & \multicolumn{3}{c|}{\textbf{A}ll} & \multicolumn{3}{c|}{\textbf{G}eometry} & \multicolumn{3}{c|}{\textbf{T}exture} & \multicolumn{3}{c|}{\textbf{A}ll} & \multicolumn{3}{c|}{\textbf{G}eometry} & \multicolumn{3}{c|}{\textbf{T}exture} & \multicolumn{3}{c}{\textbf{A}ll} \\
\midrule
Feature & \fontsize{8.5pt}{9pt}\selectfont{PSNR$\uparrow$} & \fontsize{8.5pt}{9pt}\selectfont{SSIM$\uparrow$} & \fontsize{8.5pt}{9pt}\selectfont{LPIPS$\downarrow$} & \fontsize{8.5pt}{9pt}\selectfont{PSNR$\uparrow$} & \fontsize{8.5pt}{9pt}\selectfont{SSIM$\uparrow$} & \fontsize{8.5pt}{9pt}\selectfont{LPIPS$\downarrow$} & \fontsize{8.5pt}{9pt}\selectfont{PSNR$\uparrow$} & \fontsize{8.5pt}{9pt}\selectfont{SSIM$\uparrow$} & \fontsize{8.5pt}{9pt}\selectfont{LPIPS$\downarrow$} & \fontsize{8.5pt}{9pt}\selectfont{PSNR$\uparrow$} & \fontsize{8.5pt}{9pt}\selectfont{SSIM$\uparrow$} & \fontsize{8.5pt}{9pt}\selectfont{LPIPS$\downarrow$} & \fontsize{8.5pt}{9pt}\selectfont{PSNR$\uparrow$} & \fontsize{8.5pt}{9pt}\selectfont{SSIM$\uparrow$} & \fontsize{8.5pt}{9pt}\selectfont{LPIPS$\downarrow$} & \fontsize{8.5pt}{9pt}\selectfont{PSNR$\uparrow$} & \fontsize{8.5pt}{9pt}\selectfont{SSIM$\uparrow$} & \fontsize{8.5pt}{9pt}\selectfont{LPIPS$\downarrow$} & \fontsize{8.5pt}{9pt}\selectfont{PSNR$\uparrow$} & \fontsize{8.5pt}{9pt}\selectfont{SSIM$\uparrow$} & \fontsize{8.5pt}{9pt}\selectfont{LPIPS$\downarrow$} & \fontsize{8.5pt}{9pt}\selectfont{PSNR$\uparrow$} & \fontsize{8.5pt}{9pt}\selectfont{SSIM$\uparrow$} & \fontsize{8.5pt}{9pt}\selectfont{LPIPS$\downarrow$} & \fontsize{8.5pt}{9pt}\selectfont{PSNR$\uparrow$} & \fontsize{8.5pt}{9pt}\selectfont{SSIM$\uparrow$} & \fontsize{8.5pt}{9pt}\selectfont{LPIPS$\downarrow$} \\
\midrule
\dino  &             \cellcolor[rgb]{1.00,0.82,0.82}21.15 &             \cellcolor[rgb]{1.00,0.82,0.82}.5154 &                \cellcolor[rgb]{1.00,0.82,0.82}.3794 &             \cellcolor[rgb]{1.00,0.90,0.82}19.60 &             \cellcolor[rgb]{1.00,0.90,0.82}.4746 &                \cellcolor[rgb]{0.56,0.75,0.38}.3625 &             \cellcolor[rgb]{1.00,0.82,0.82}21.12 &             \cellcolor[rgb]{1.00,0.82,0.82}.5136 &                \cellcolor[rgb]{0.56,0.75,0.38}.4533 &             \cellcolor[rgb]{1.00,0.82,0.82}19.44 &             \cellcolor[rgb]{1.00,0.82,0.82}.5973 &                \cellcolor[rgb]{1.00,0.82,0.82}.3152 &             \cellcolor[rgb]{1.00,0.82,0.82}16.84 &             \cellcolor[rgb]{0.56,0.75,0.38}.5362 &                \cellcolor[rgb]{0.56,0.75,0.38}.3323 &             \cellcolor[rgb]{1.00,0.82,0.82}19.41 &             \cellcolor[rgb]{1.00,0.90,0.82}.5956 &                \cellcolor[rgb]{0.56,0.75,0.38}.3651 &             \cellcolor[rgb]{1.00,0.82,0.82}18.29 &             \cellcolor[rgb]{1.00,0.82,0.82}.6334 &                \cellcolor[rgb]{1.00,0.82,0.82}.3818 &             \cellcolor[rgb]{0.56,0.75,0.38}18.18 &             \cellcolor[rgb]{0.85,0.88,0.67}.6368 &                \cellcolor[rgb]{0.85,0.88,0.67}.3202 &             \cellcolor[rgb]{1.00,0.82,0.82}18.82 &             \cellcolor[rgb]{1.00,0.82,0.82}.6503 &                \cellcolor[rgb]{1.00,0.90,0.82}.3976 \\
RADIOv2 &             \cellcolor[rgb]{1.00,0.90,0.82}21.21 &             \cellcolor[rgb]{0.56,0.75,0.38}.5341 &                \cellcolor[rgb]{0.56,0.75,0.38}.3438 &             \cellcolor[rgb]{0.85,0.88,0.67}19.71 &             \cellcolor[rgb]{0.85,0.88,0.67}.4760 &                \cellcolor[rgb]{1.00,0.82,0.82}.3656 &             \cellcolor[rgb]{1.00,0.90,0.82}21.21 &             \cellcolor[rgb]{0.85,0.88,0.67}.5228 &                \cellcolor[rgb]{1.00,0.82,0.82}.4930 &             \cellcolor[rgb]{0.85,0.88,0.67}19.55 &             \cellcolor[rgb]{0.56,0.75,0.38}.6121 &                \cellcolor[rgb]{0.85,0.88,0.67}.2934 &             \cellcolor[rgb]{0.56,0.75,0.38}16.97 &             \cellcolor[rgb]{1.00,0.82,0.82}.5327 &                \cellcolor[rgb]{1.00,0.82,0.82}.3348 &             \cellcolor[rgb]{0.56,0.75,0.38}19.57 &             \cellcolor[rgb]{1.00,0.82,0.82}.5952 &                \cellcolor[rgb]{1.00,0.82,0.82}.3940 &             \cellcolor[rgb]{0.56,0.75,0.38}19.43 &             \cellcolor[rgb]{0.56,0.75,0.38}.6695 &                \cellcolor[rgb]{0.56,0.75,0.38}.3422 &             \cellcolor[rgb]{0.85,0.88,0.67}18.13 &             \cellcolor[rgb]{1.00,0.90,0.82}.6311 &                \cellcolor[rgb]{1.00,0.82,0.82}.3220 &             \cellcolor[rgb]{0.56,0.75,0.38}19.07 &             \cellcolor[rgb]{0.56,0.75,0.38}.6609 &                \cellcolor[rgb]{1.00,0.82,0.82}.4067 \\
\master  &             \cellcolor[rgb]{0.85,0.88,0.67}21.27 &             \cellcolor[rgb]{1.00,0.90,0.82}.5272 &                \cellcolor[rgb]{1.00,0.90,0.82}.3568 &             \cellcolor[rgb]{1.00,0.82,0.82}19.55 &             \cellcolor[rgb]{1.00,0.82,0.82}.4722 &                \cellcolor[rgb]{0.85,0.88,0.67}.3633 &             \cellcolor[rgb]{0.85,0.88,0.67}21.26 &             \cellcolor[rgb]{1.00,0.90,0.82}.5217 &                \cellcolor[rgb]{0.85,0.88,0.67}.4572 &             \cellcolor[rgb]{1.00,0.90,0.82}19.50 &             \cellcolor[rgb]{1.00,0.90,0.82}.6055 &                \cellcolor[rgb]{1.00,0.90,0.82}.2971 &             \cellcolor[rgb]{1.00,0.90,0.82}16.92 &             \cellcolor[rgb]{0.85,0.88,0.67}.5354 &                \cellcolor[rgb]{0.56,0.75,0.38}.3323 &             \cellcolor[rgb]{0.85,0.88,0.67}19.53 &             \cellcolor[rgb]{0.56,0.75,0.38}.5998 &                \cellcolor[rgb]{0.85,0.88,0.67}.3654 &             \cellcolor[rgb]{1.00,0.90,0.82}19.11 &             \cellcolor[rgb]{1.00,0.90,0.82}.6542 &                \cellcolor[rgb]{1.00,0.90,0.82}.3596 &             \cellcolor[rgb]{1.00,0.82,0.82}18.02 &             \cellcolor[rgb]{0.56,0.75,0.38}.6385 &                \cellcolor[rgb]{0.56,0.75,0.38}.3094 &             \cellcolor[rgb]{1.00,0.90,0.82}18.90 &             \cellcolor[rgb]{1.00,0.90,0.82}.6569 &                \cellcolor[rgb]{0.56,0.75,0.38}.3930 \\
\ours    &             \cellcolor[rgb]{0.56,0.75,0.38}21.38 &             \cellcolor[rgb]{0.85,0.88,0.67}.5340 &                \cellcolor[rgb]{0.85,0.88,0.67}.3527 &             \cellcolor[rgb]{0.56,0.75,0.38}19.72 &             \cellcolor[rgb]{0.56,0.75,0.38}.4791 &                \cellcolor[rgb]{1.00,0.90,0.82}.3636 &             \cellcolor[rgb]{0.56,0.75,0.38}21.27 &             \cellcolor[rgb]{0.56,0.75,0.38}.5254 &                \cellcolor[rgb]{1.00,0.90,0.82}.4609 &             \cellcolor[rgb]{0.56,0.75,0.38}19.59 &             \cellcolor[rgb]{0.85,0.88,0.67}.6115 &                \cellcolor[rgb]{0.56,0.75,0.38}.2912 &             \cellcolor[rgb]{0.85,0.88,0.67}16.93 &             \cellcolor[rgb]{1.00,0.90,0.82}.5346 &                \cellcolor[rgb]{1.00,0.90,0.82}.3342 &             \cellcolor[rgb]{1.00,0.90,0.82}19.48 &             \cellcolor[rgb]{0.85,0.88,0.67}.5980 &                \cellcolor[rgb]{1.00,0.90,0.82}.3685 &             \cellcolor[rgb]{0.85,0.88,0.67}19.36 &             \cellcolor[rgb]{0.85,0.88,0.67}.6621 &                \cellcolor[rgb]{0.85,0.88,0.67}.3546 &             \cellcolor[rgb]{1.00,0.90,0.82}18.11 &             \cellcolor[rgb]{1.00,0.82,0.82}.6300 &                \cellcolor[rgb]{1.00,0.90,0.82}.3219 &             \cellcolor[rgb]{0.85,0.88,0.67}19.02 &             \cellcolor[rgb]{0.85,0.88,0.67}.6592 &                \cellcolor[rgb]{0.85,0.88,0.67}.3951 \\
\bottomrule
\end{tabular}
}
\vspace{-0.2cm}
\caption{
    \textbf{Per-dataset results of Novel View Synthesis metrics in the Feat2GS benchmark~\cite{feat2gs}}.
    { Note that models vary in size: RADIOv2 is a ViT-H, \master a ViT-L and \dino and \ours a ViT-B.}
}
\label{tab:feat2gs}
\end{table*}

%% file: tex/float/fig_attention.tex
\begin{figure*}[t]
\vspace{-0.75cm}
\adjustbox{max width=\linewidth}{
\begin{tabular}{ c | ccc ccc ccc}
\multicolumn{10}{c}{The last encoder block} \\
\includegraphics[width=2cm]{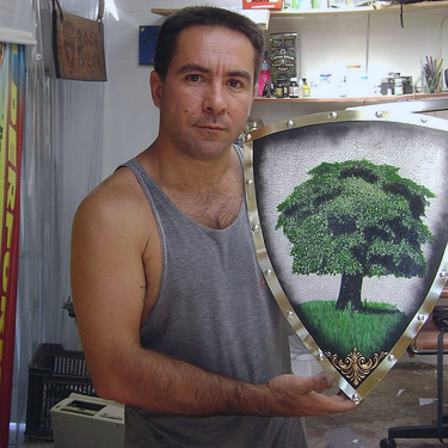} &
    \includegraphics[width=2cm]{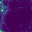} &
    \includegraphics[width=2cm]{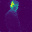} &
    \includegraphics[width=2cm]{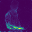} &
    \includegraphics[width=2cm]{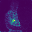} &
    \includegraphics[width=2cm]{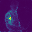} &
    \includegraphics[width=2cm]{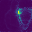} &
    \includegraphics[width=2cm]{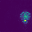} &
    \includegraphics[width=2cm]{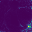} &
    \includegraphics[width=2cm]{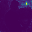} \\
\multicolumn{10}{c}{Transformer projector for \dino} \\
\includegraphics[width=2cm]{tex/res/attention/im-0084_0orig.png} &
    \includegraphics[width=2cm]{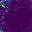} &
    \includegraphics[width=2cm]{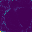} &
    \includegraphics[width=2cm]{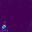} &
    \includegraphics[width=2cm]{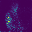} &
    \includegraphics[width=2cm]{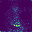} &
    \includegraphics[width=2cm]{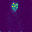} &
    \includegraphics[width=2cm]{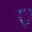} &
    \includegraphics[width=2cm]{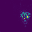} &
    \includegraphics[width=2cm]{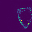} \\
\multicolumn{10}{c}{Transformer projector for \master} \\
\includegraphics[width=2cm]{tex/res/attention/im-0084_0orig.png} &
    \includegraphics[width=2cm]{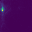} &
    \includegraphics[width=2cm]{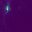} &
    \includegraphics[width=2cm]{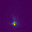} &
    \includegraphics[width=2cm]{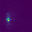} &
    \includegraphics[width=2cm]{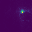} &
    \includegraphics[width=2cm]{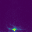} &
    \includegraphics[width=2cm]{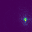} &
    \includegraphics[width=2cm]{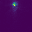} &
    \includegraphics[width=2cm]{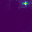} \\
\multicolumn{10}{c}{Transformer projector for \hmr} \\
\includegraphics[width=2cm]{tex/res/attention/im-0084_0orig.png} &
    \includegraphics[width=2cm]{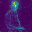} &
    \includegraphics[width=2cm]{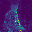} &
    \includegraphics[width=2cm]{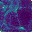} &
    \includegraphics[width=2cm]{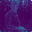} &
    \includegraphics[width=2cm]{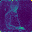} &
    \includegraphics[width=2cm]{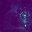} &
    \includegraphics[width=2cm]{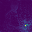} &
    \includegraphics[width=2cm]{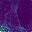} &
    \includegraphics[width=2cm]{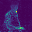} \\
\midrule
\multicolumn{10}{c}{The last encoder block} \\
\includegraphics[width=2cm]{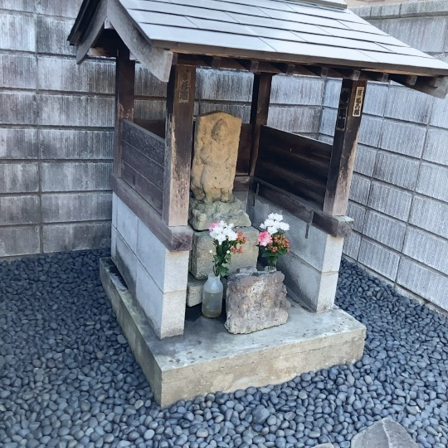} &
    \includegraphics[width=2cm]{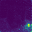} &
    \includegraphics[width=2cm]{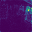} &
    \includegraphics[width=2cm]{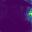} &
    \includegraphics[width=2cm]{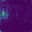} &
    \includegraphics[width=2cm]{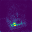} &
    \includegraphics[width=2cm]{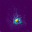} &
    \includegraphics[width=2cm]{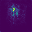} &
    \includegraphics[width=2cm]{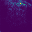} &
    \includegraphics[width=2cm]{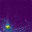} \\
\multicolumn{10}{c}{Transformer projector for \dino} \\
\includegraphics[width=2cm]{tex/res/attention/im-0123_0orig.png} &
    \includegraphics[width=2cm]{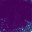} &
    \includegraphics[width=2cm]{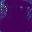} &
    \includegraphics[width=2cm]{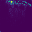} &
    \includegraphics[width=2cm]{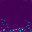} &
    \includegraphics[width=2cm]{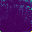} &
    \includegraphics[width=2cm]{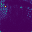} &
    \includegraphics[width=2cm]{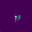} &
    \includegraphics[width=2cm]{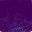} &
    \includegraphics[width=2cm]{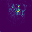} \\
\multicolumn{10}{c}{Transformer projector for \master} \\
\includegraphics[width=2cm]{tex/res/attention/im-0123_0orig.png} &
    \includegraphics[width=2cm]{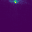} &
    \includegraphics[width=2cm]{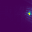} &
    \includegraphics[width=2cm]{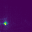} &
    \includegraphics[width=2cm]{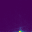} &
    \includegraphics[width=2cm]{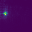} &
    \includegraphics[width=2cm]{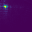} &
    \includegraphics[width=2cm]{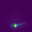} &
    \includegraphics[width=2cm]{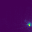} &
    \includegraphics[width=2cm]{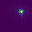} \\
\multicolumn{10}{c}{Transformer projector for \hmr} \\
\includegraphics[width=2cm]{tex/res/attention/im-0123_0orig.png} &
    \includegraphics[width=2cm]{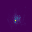} &
    \includegraphics[width=2cm]{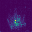} &
    \includegraphics[width=2cm]{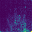} &
    \includegraphics[width=2cm]{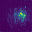} &
    \includegraphics[width=2cm]{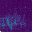} &
    \includegraphics[width=2cm]{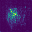} &
    \includegraphics[width=2cm]{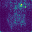} &
    \includegraphics[width=2cm]{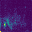} &
    \includegraphics[width=2cm]{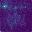} \\
\midrule
\multicolumn{10}{c}{The last encoder block} \\
\includegraphics[width=2cm]{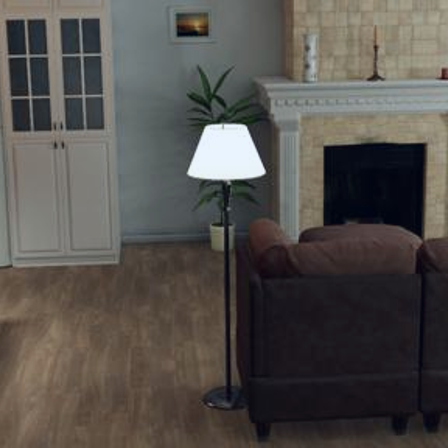} &
    \includegraphics[width=2cm]{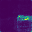} &
    \includegraphics[width=2cm]{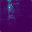} &
    \includegraphics[width=2cm]{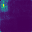} &
    \includegraphics[width=2cm]{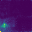} &
    \includegraphics[width=2cm]{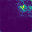} &
    \includegraphics[width=2cm]{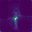} &
    \includegraphics[width=2cm]{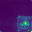} &
    \includegraphics[width=2cm]{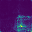} &
    \includegraphics[width=2cm]{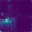} \\
\multicolumn{10}{c}{Transformer projector for \dino} \\
\includegraphics[width=2cm]{tex/res/attention/im-0217_0orig.png} &
    \includegraphics[width=2cm]{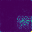} &
    \includegraphics[width=2cm]{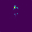} &
    \includegraphics[width=2cm]{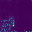} &
    \includegraphics[width=2cm]{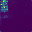} &
    \includegraphics[width=2cm]{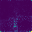} &
    \includegraphics[width=2cm]{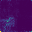} &
    \includegraphics[width=2cm]{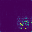} &
    \includegraphics[width=2cm]{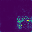} &
    \includegraphics[width=2cm]{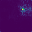} \\
\multicolumn{10}{c}{Transformer projector for \master} \\
\includegraphics[width=2cm]{tex/res/attention/im-0217_0orig.png} &
    \includegraphics[width=2cm]{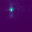} &
    \includegraphics[width=2cm]{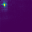} &
    \includegraphics[width=2cm]{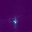} &
    \includegraphics[width=2cm]{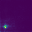} &
    \includegraphics[width=2cm]{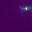} &
    \includegraphics[width=2cm]{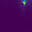} &
    \includegraphics[width=2cm]{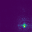} &
    \includegraphics[width=2cm]{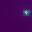} &
    \includegraphics[width=2cm]{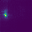} \\
\multicolumn{10}{c}{Transformer projector for \hmr} \\
\includegraphics[width=2cm]{tex/res/attention/im-0217_0orig.png} &
    \includegraphics[width=2cm]{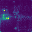} &
    \includegraphics[width=2cm]{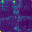} &
    \includegraphics[width=2cm]{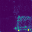} &
    \includegraphics[width=2cm]{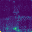} &
    \includegraphics[width=2cm]{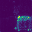} &
    \includegraphics[width=2cm]{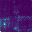} &
    \includegraphics[width=2cm]{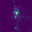} &
    \includegraphics[width=2cm]{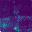} &
    \includegraphics[width=2cm]{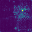} \\
\end{tabular}
}
\vspace{-0.3cm}
\caption{
    {\bf Visualization of attention maps.}
    Given an image of resolution $448\times448$ (1st column), we extract using our student model the attention probability map (of size $32\times32$) for each patch from either
    the last encoder layer or the Transformer projector for each teacher.
    Then, we flatten each map and run $k$-medoids clustering with $k=9$, and visualize centroids.
}
\label{fig:attention}
\end{figure*}

%% file: tex/float/supplementary/fig_mast3r.tex
\begin{figure*}
    \centering
    \includegraphics[width=0.8\linewidth]{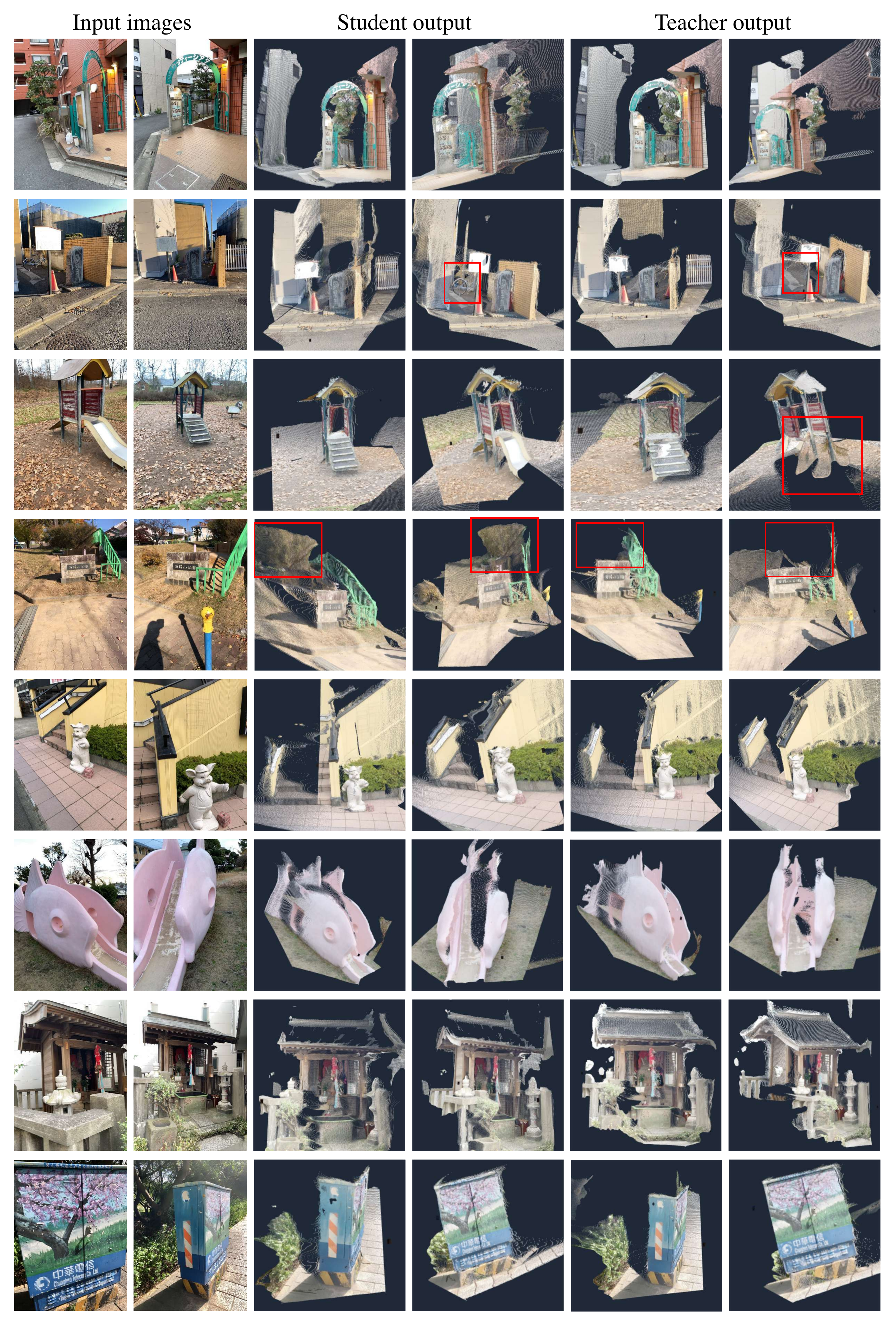}
    \vspace{-0.3cm}
    \caption{\textbf{Qualitative results for the \master teacher and our student.} Each row presents two input images and corresponding 3D reconstructions. Images were sampled from the Niantic dataset. With a red square, we highlight regions where our student seems to outperform the teacher.}
    \label{fig:mast3r_qualitative_1}
\end{figure*}

\newpage

\begin{figure*}
    \centering
    \includegraphics[width=0.8\linewidth]{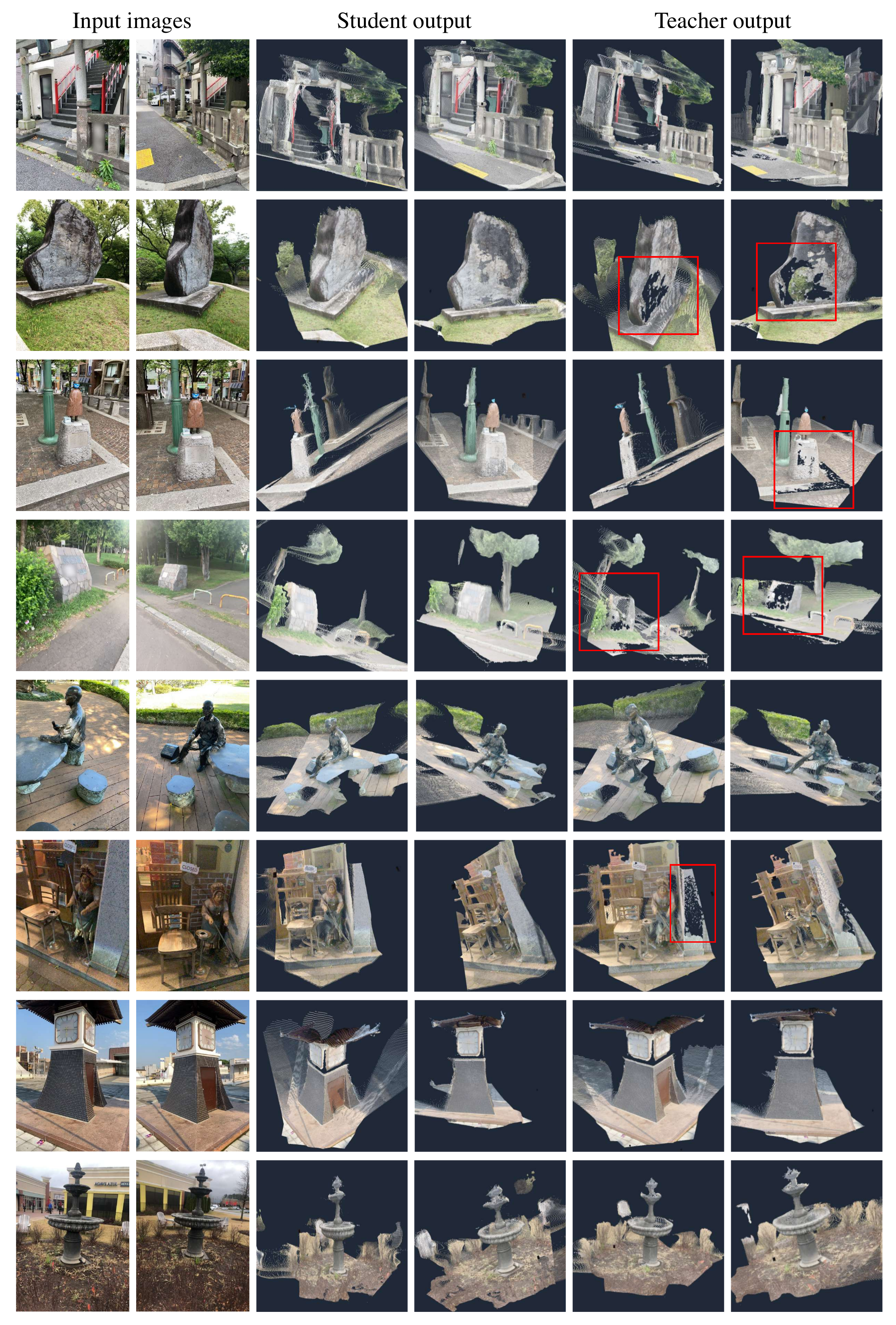}
    \vspace{-0.3cm}
    \caption{\textbf{Qualitative results for the \master teacher and our student.} Each row presents two input images and corresponding 3D reconstructions. Images were sampled from the Niantic dataset. With a red square, we highlight regions where our student seems to outperform the teacher.}
    \label{fig:mast3r_qualitative_2}
\end{figure*}

\newpage

\begin{figure*}
    \centering
    \includegraphics[width=0.73\linewidth]{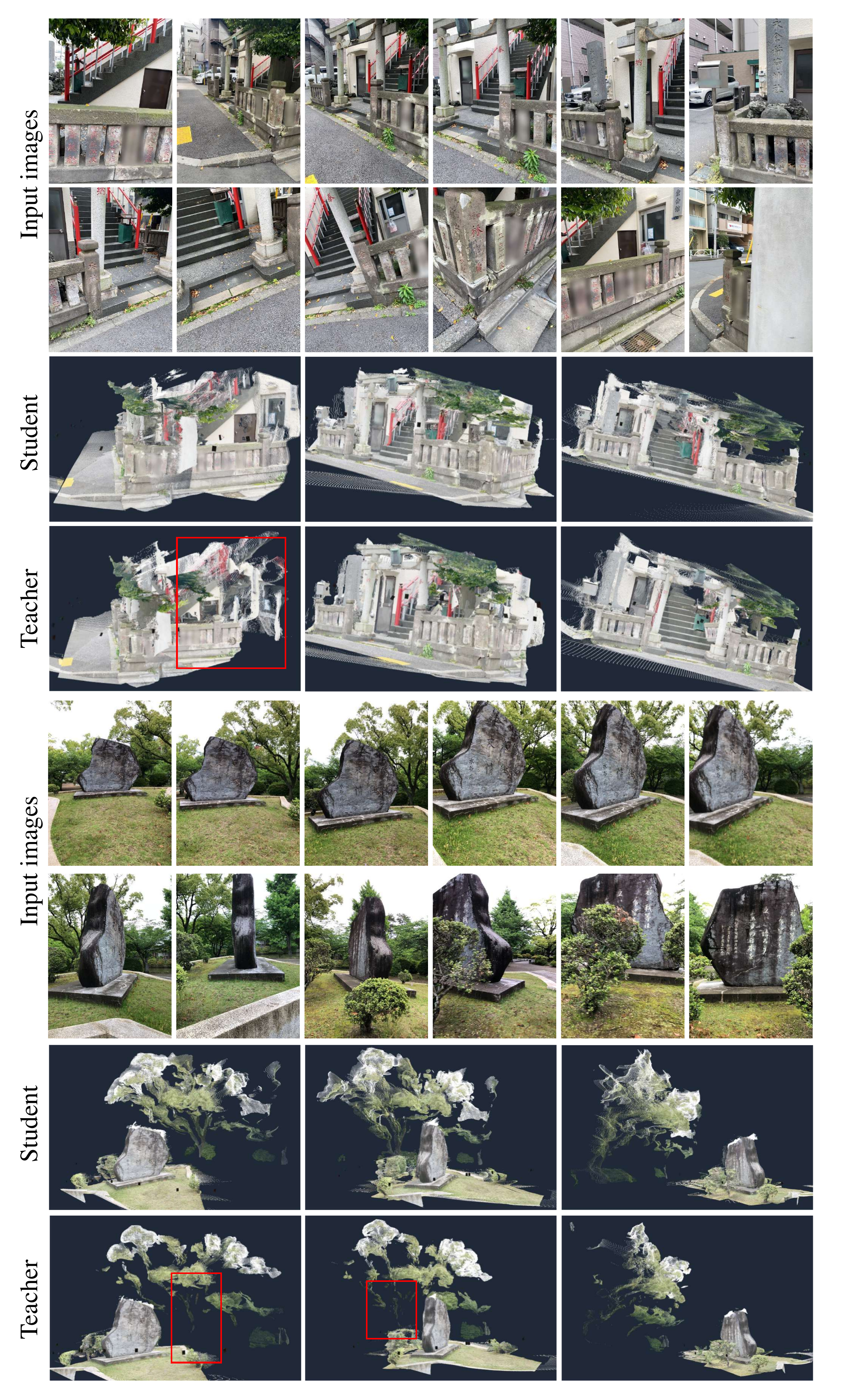}
    \vspace{-0.3cm}
    \caption{\textbf{Scene reconstructions from longer input sequences for the \master teacher and our student.} With a red square, we highlight regions where our student seems to outperform the teacher.
    }
    \label{fig:mast3r_qualitative_many}
\end{figure*}

%% file: tex/float/supplementary/fig_hmr.tex
\begin{figure*}
\begin{center}

\renewcommand{\arraystretch}{0.5}
    \begin{tabular}{ccc|ccc}

Input image & Student output & Teacher output &  Input image & Student output & Teacher output  \\
\includegraphics[width=0.15\linewidth]{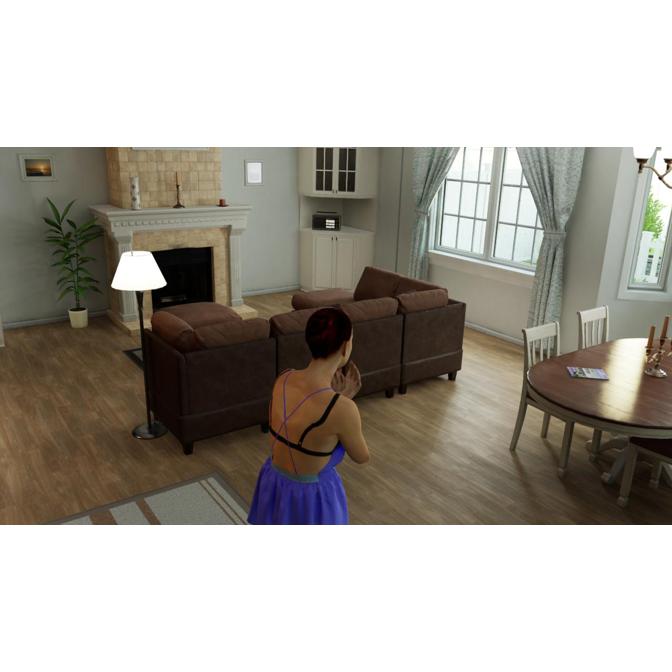} &
\includegraphics[width=0.15\linewidth]{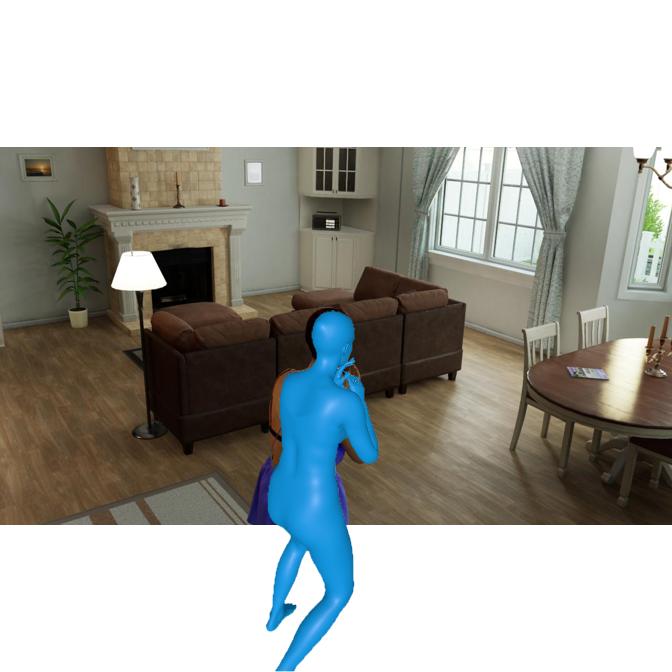} &
\includegraphics[width=0.15\linewidth]{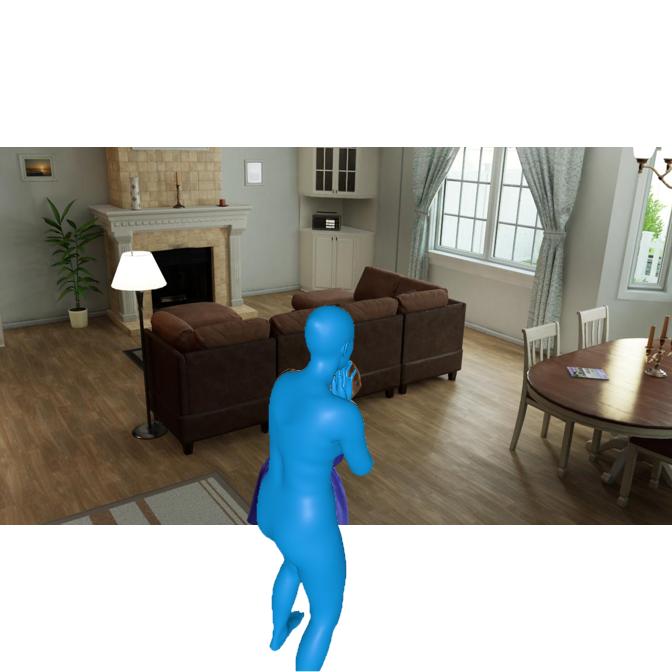} &

\includegraphics[width=0.15\linewidth]{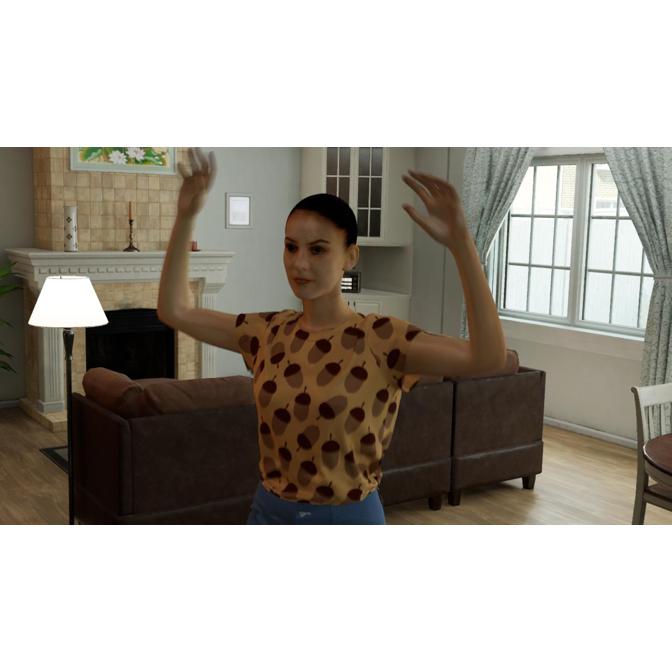} &
\includegraphics[width=0.15\linewidth]{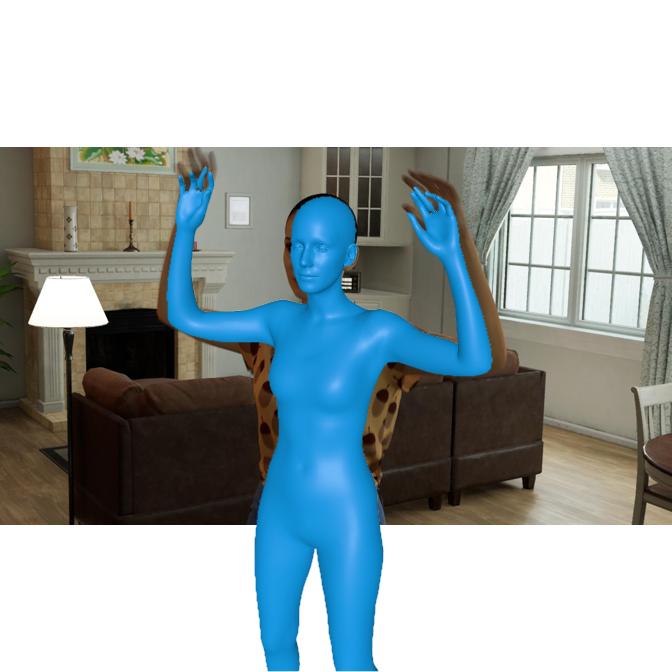} &
\includegraphics[width=0.15\linewidth]{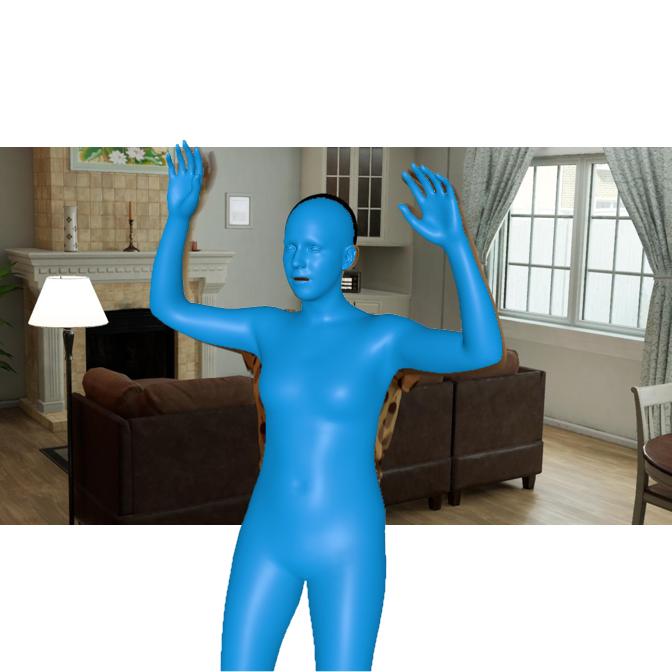} \\

\includegraphics[width=0.15\linewidth]{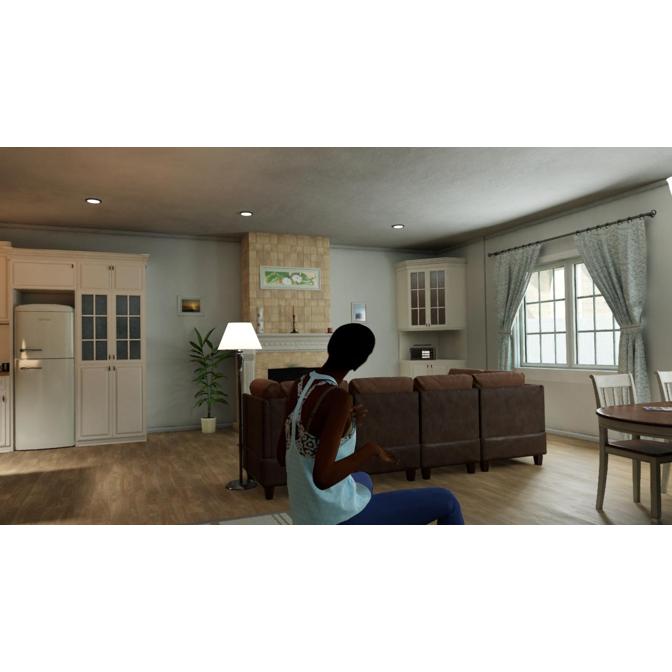} &
\includegraphics[width=0.15\linewidth]{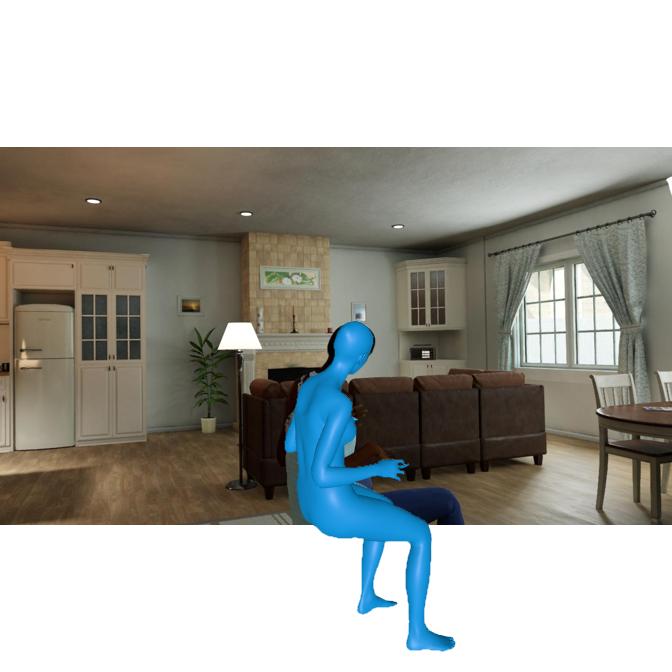} &
\includegraphics[width=0.15\linewidth]{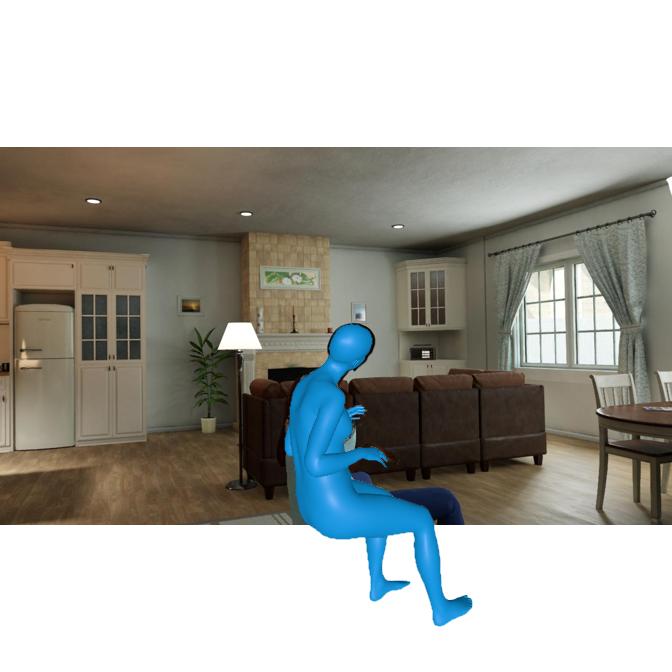} &

\includegraphics[width=0.15\linewidth]{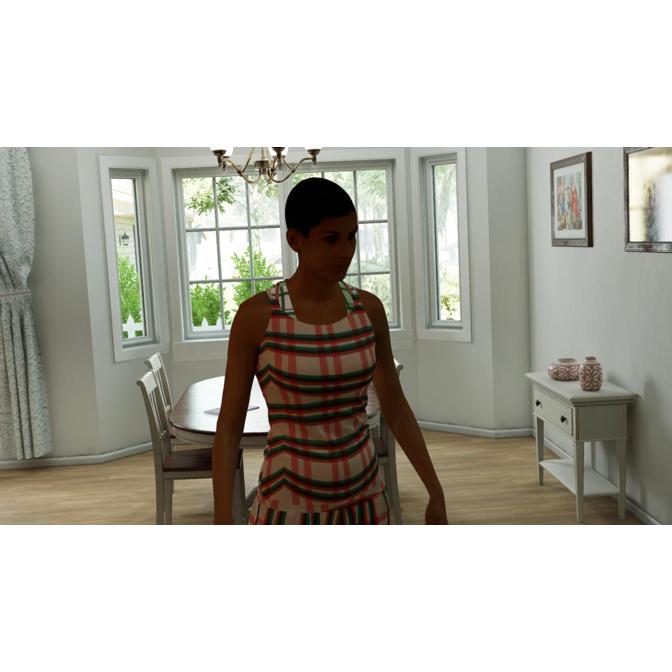} &
\includegraphics[width=0.15\linewidth]{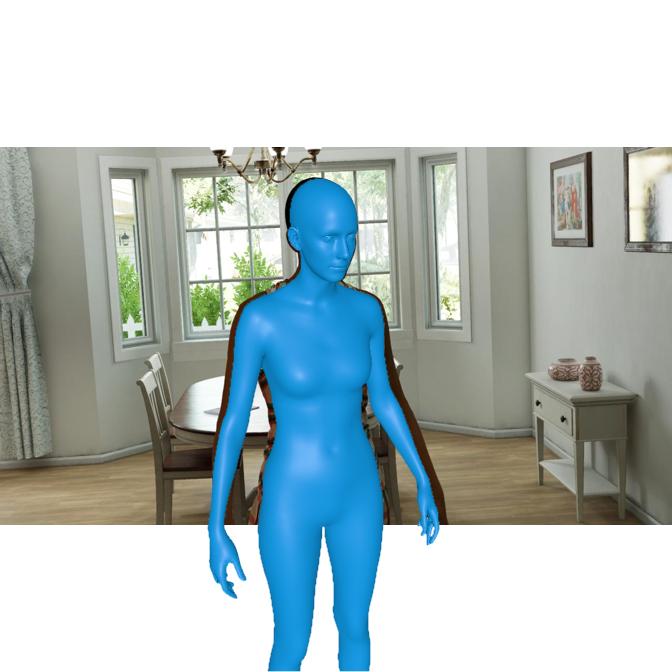} &
\includegraphics[width=0.15\linewidth]{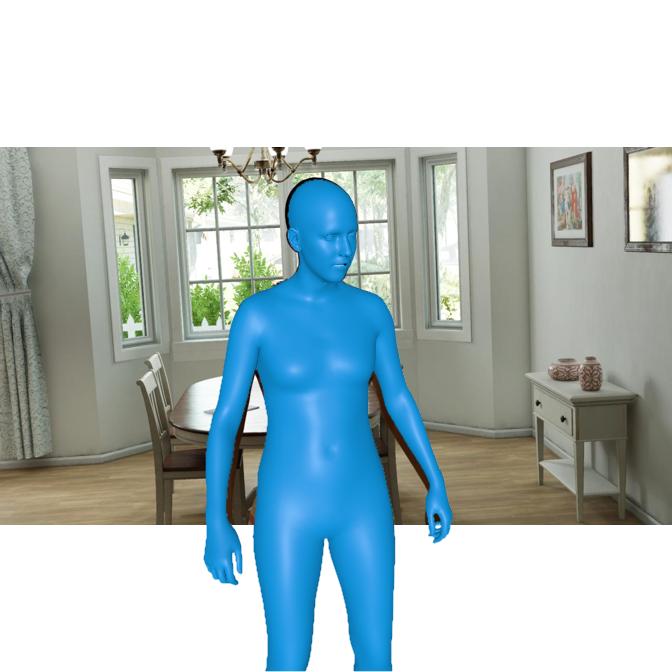} \\

\includegraphics[width=0.15\linewidth]{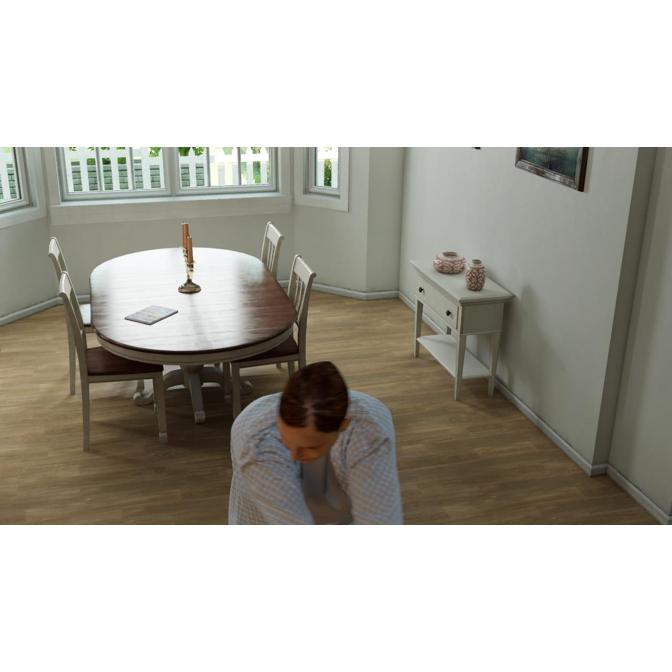} &
\includegraphics[width=0.15\linewidth]{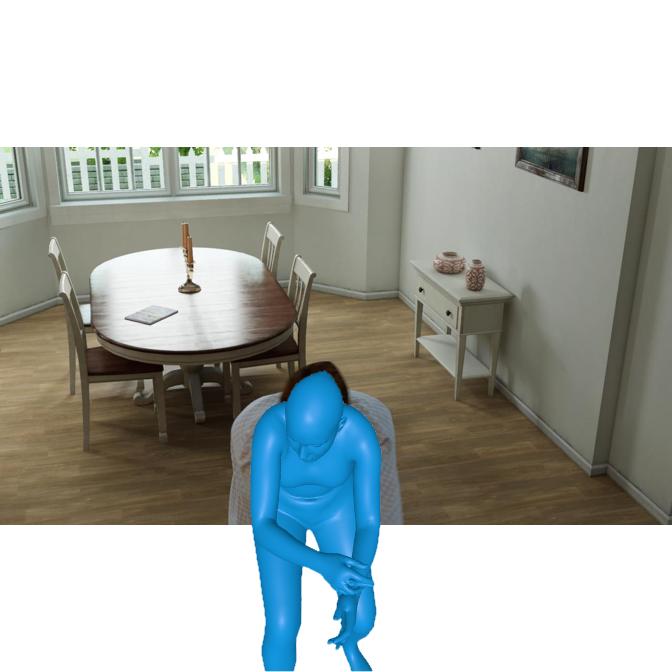} &
\includegraphics[width=0.15\linewidth]{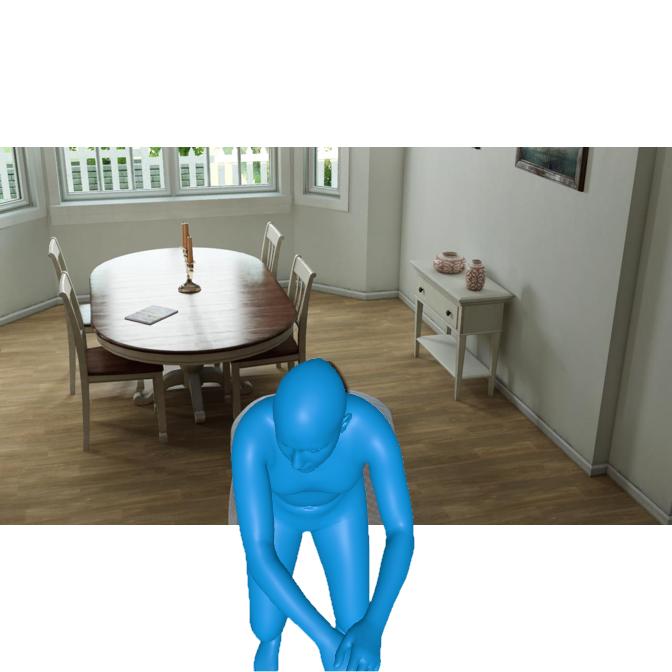} &

\includegraphics[width=0.15\linewidth]{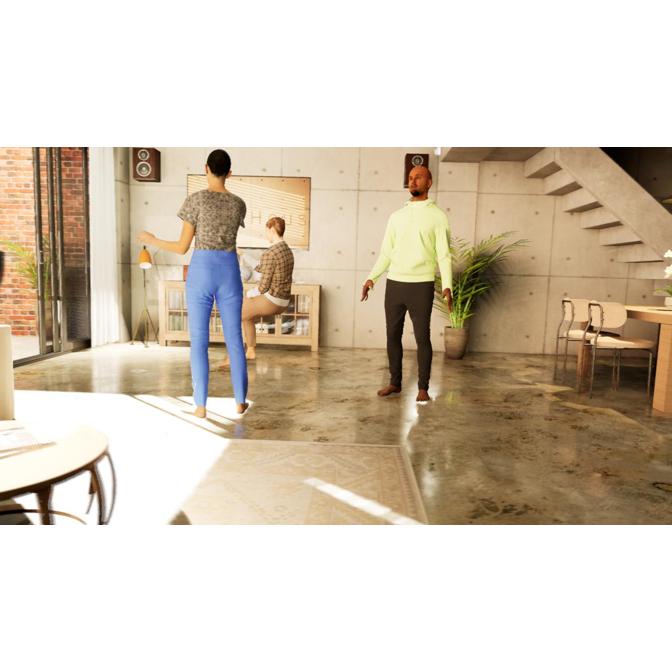} &
\includegraphics[width=0.15\linewidth]{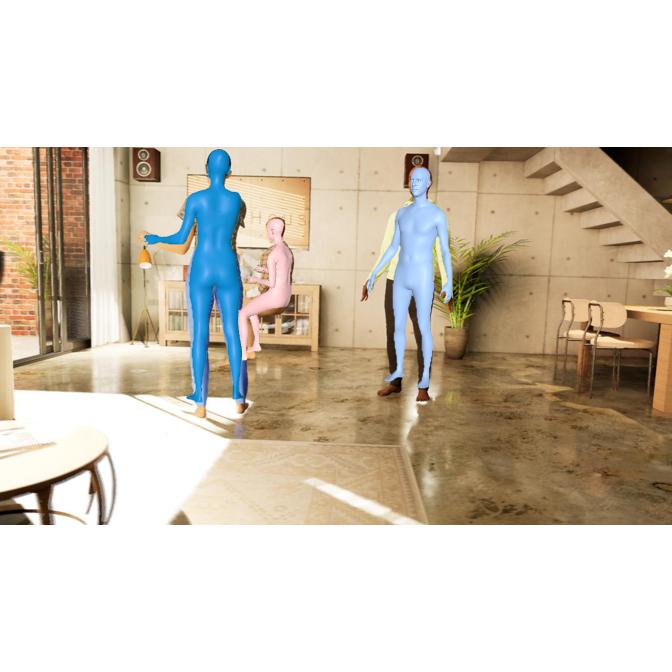} &
\includegraphics[width=0.15\linewidth]{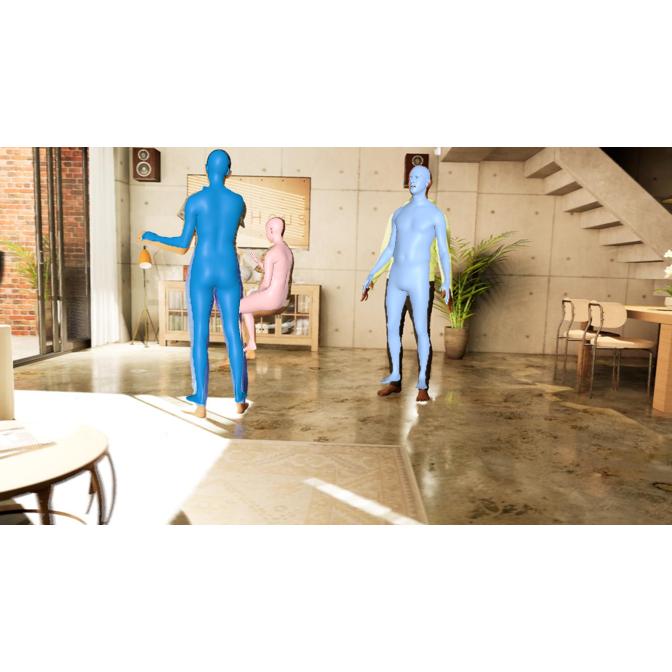} \\

\includegraphics[width=0.15\linewidth]{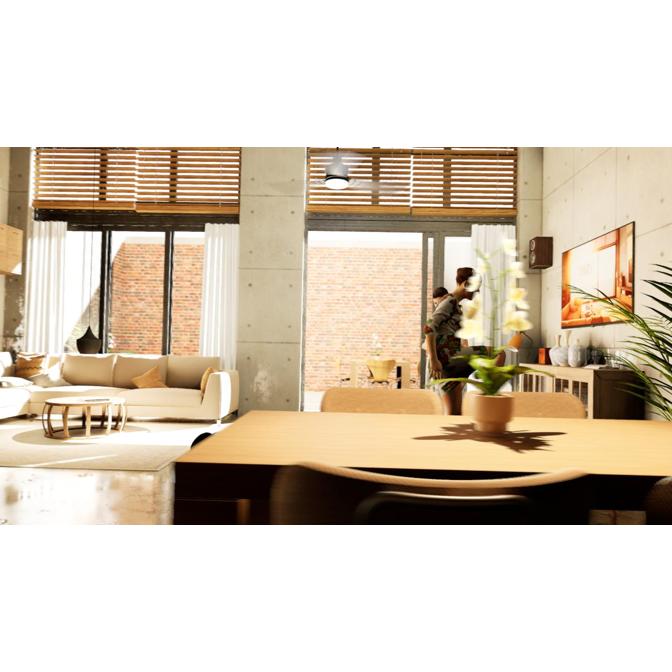} &
\includegraphics[width=0.15\linewidth]{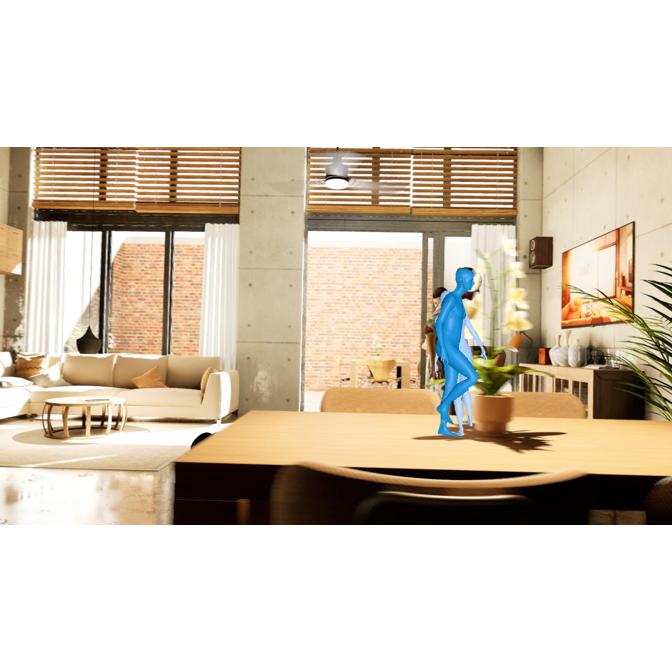} &
\includegraphics[width=0.15\linewidth]{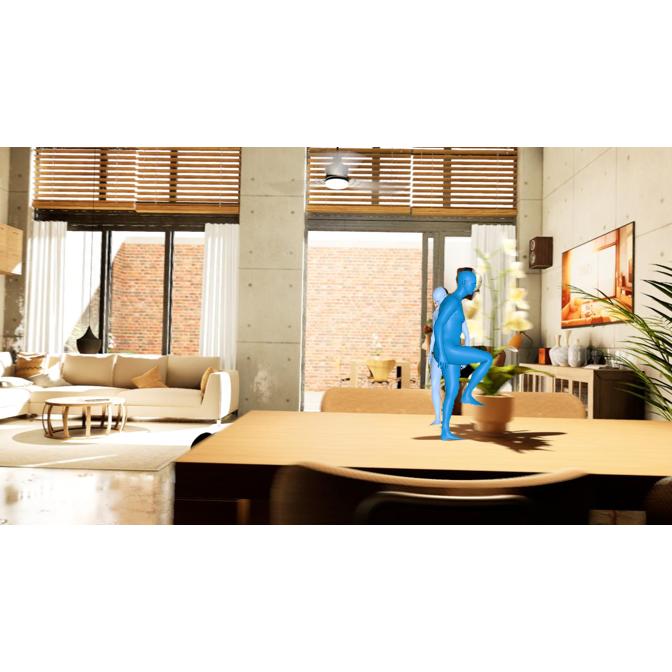} &

\includegraphics[width=0.15\linewidth]{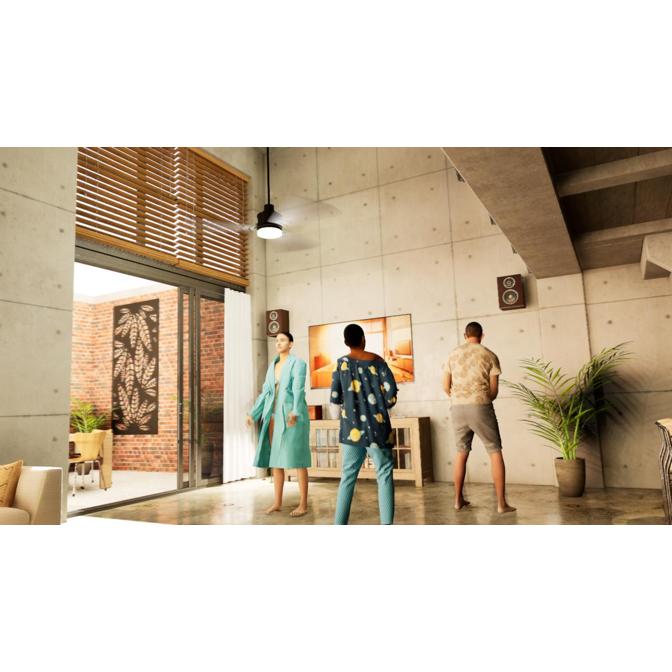} &
\includegraphics[width=0.15\linewidth]{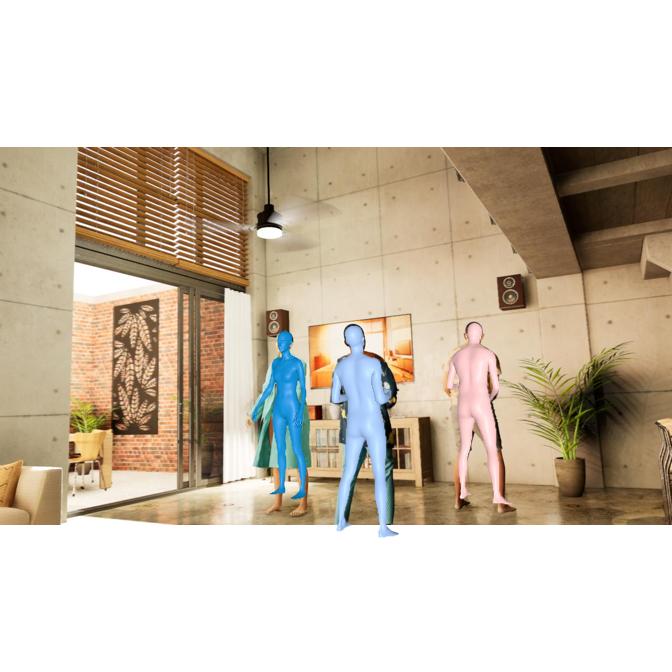} &
\includegraphics[width=0.15\linewidth]{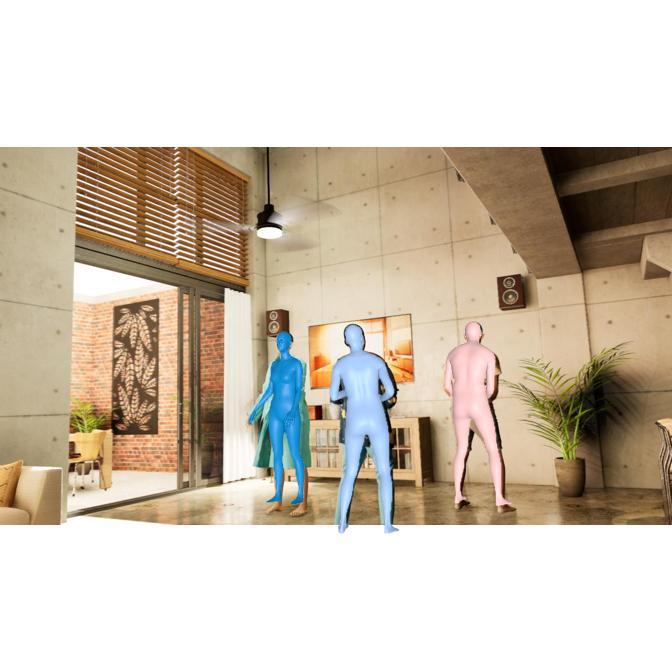} \\

\includegraphics[width=0.15\linewidth]{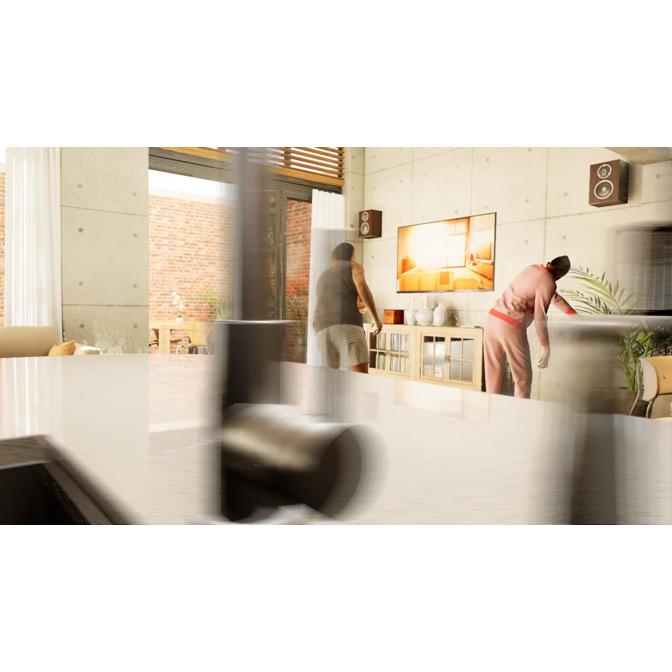} &
\includegraphics[width=0.15\linewidth]{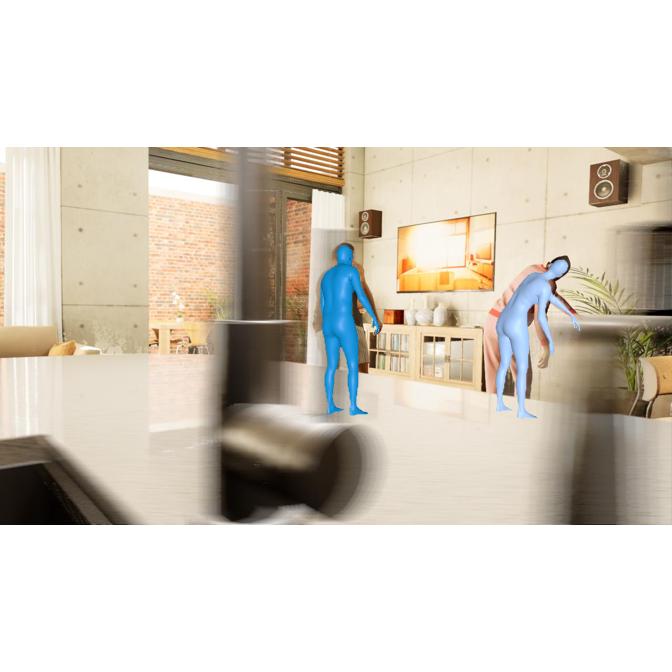} &
\includegraphics[width=0.15\linewidth]{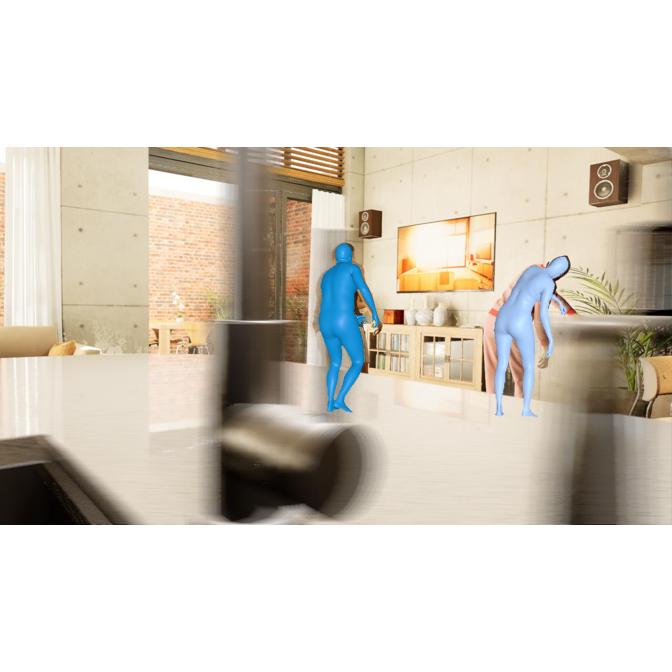} &

\includegraphics[width=0.15\linewidth]{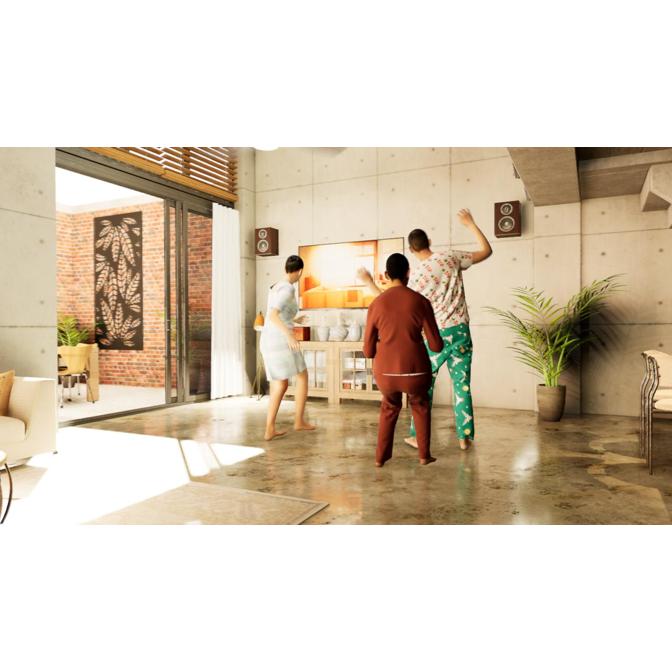} &
\includegraphics[width=0.15\linewidth]{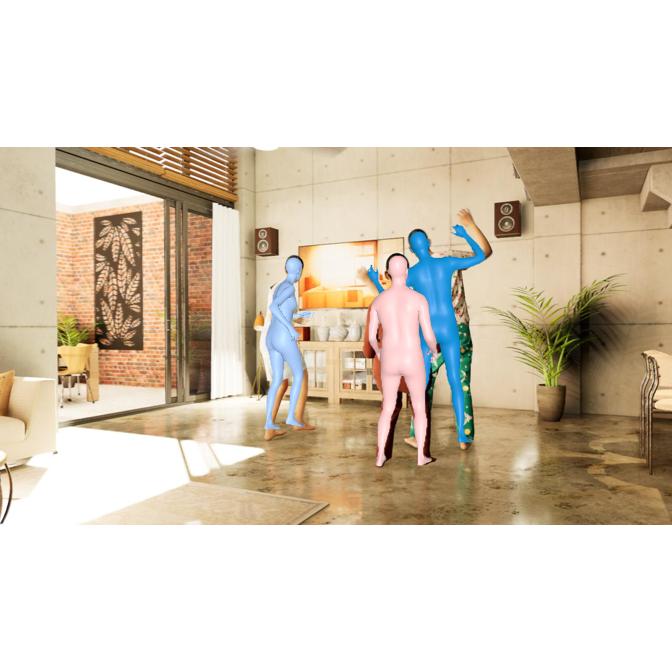} &
\includegraphics[width=0.15\linewidth]{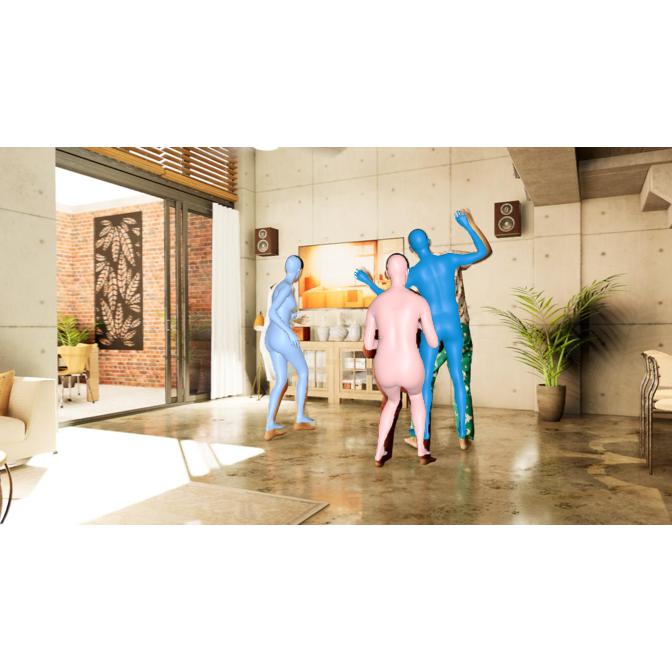} \\

\includegraphics[width=0.15\linewidth]{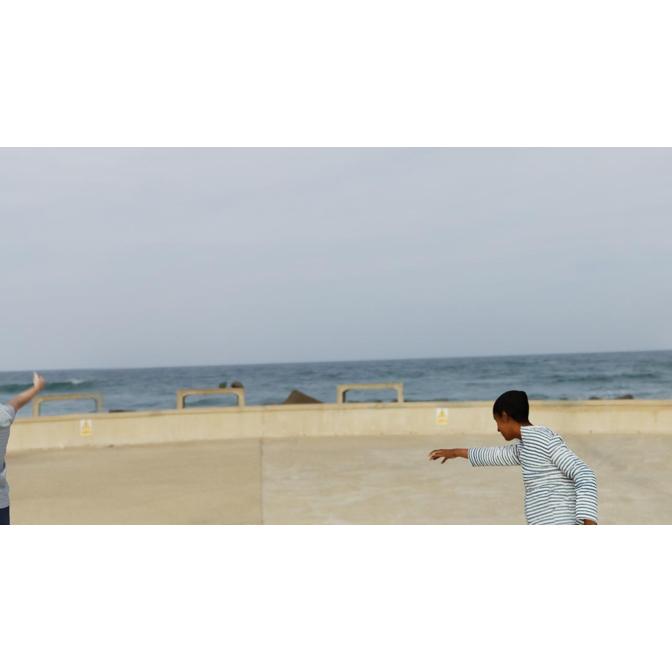} &
\includegraphics[width=0.15\linewidth]{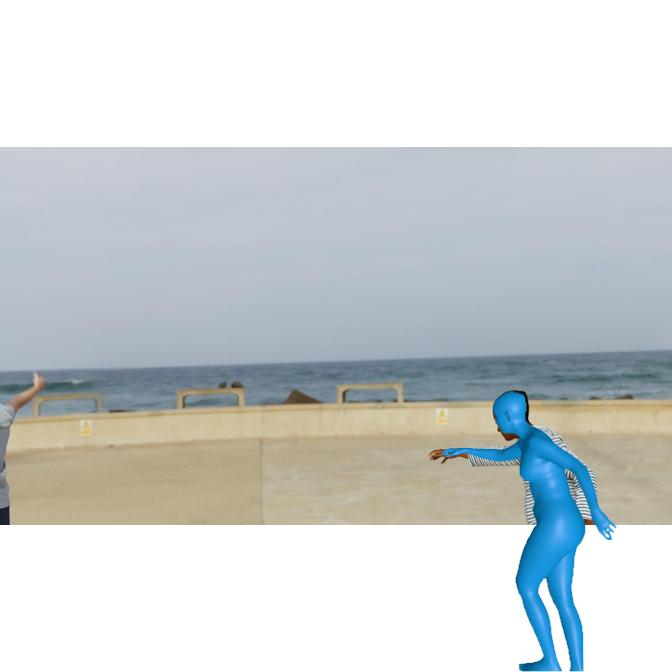} &
\includegraphics[width=0.15\linewidth]{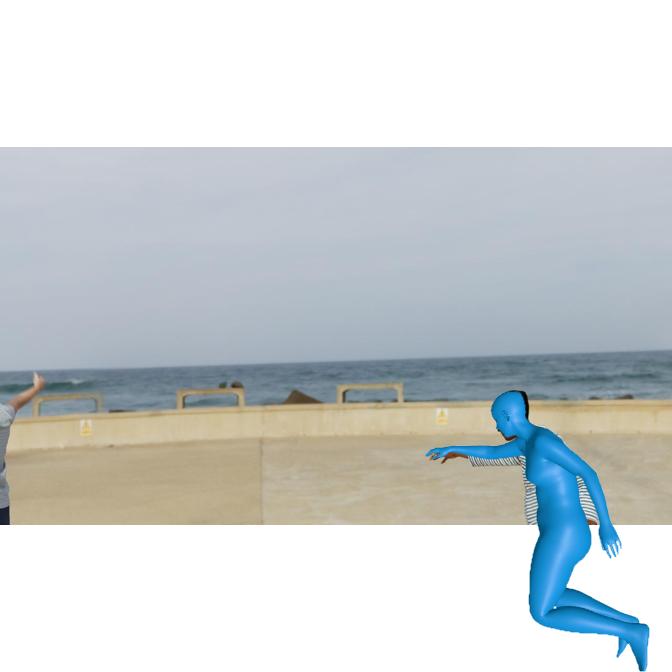} &

\includegraphics[width=0.15\linewidth]{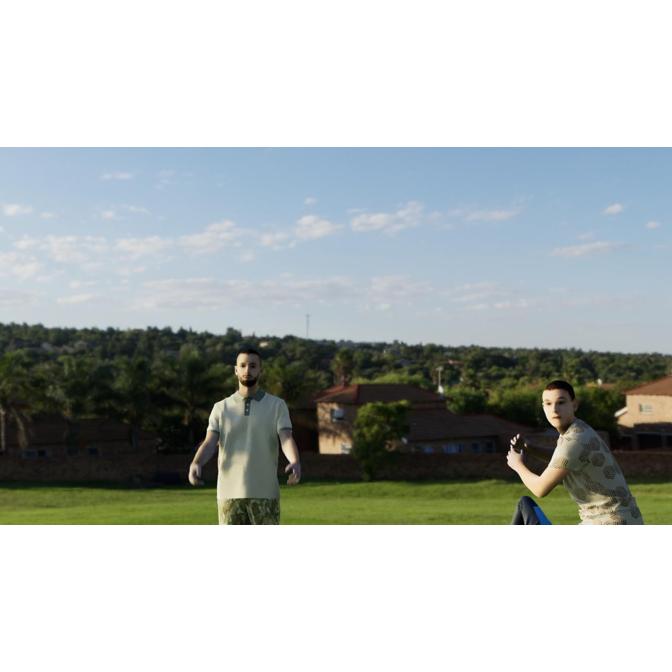} &
\includegraphics[width=0.15\linewidth]{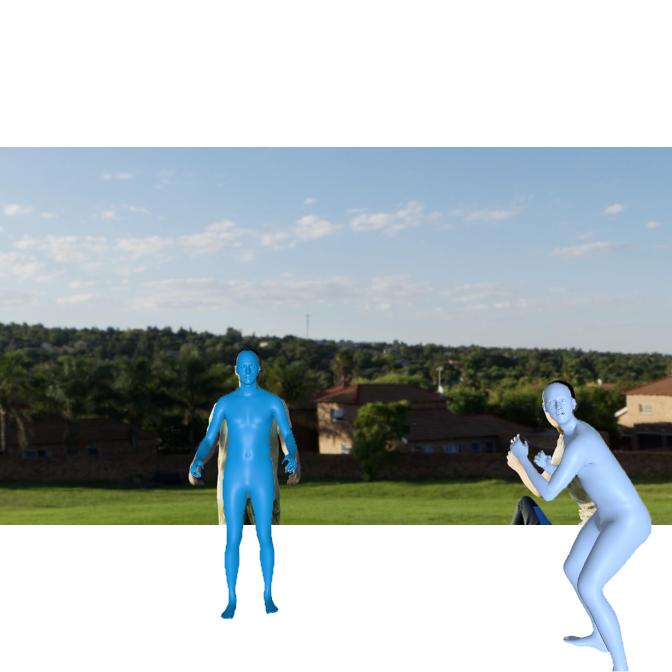} &
\includegraphics[width=0.15\linewidth]{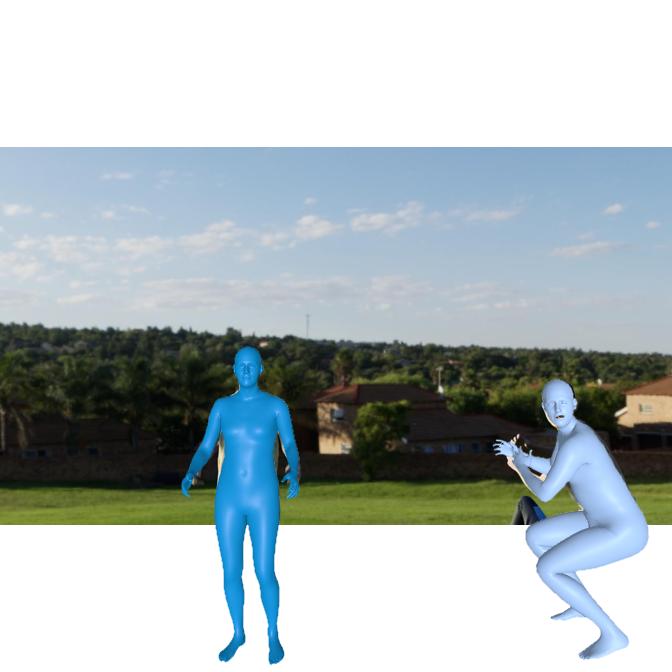} \\

\includegraphics[width=0.15\linewidth]{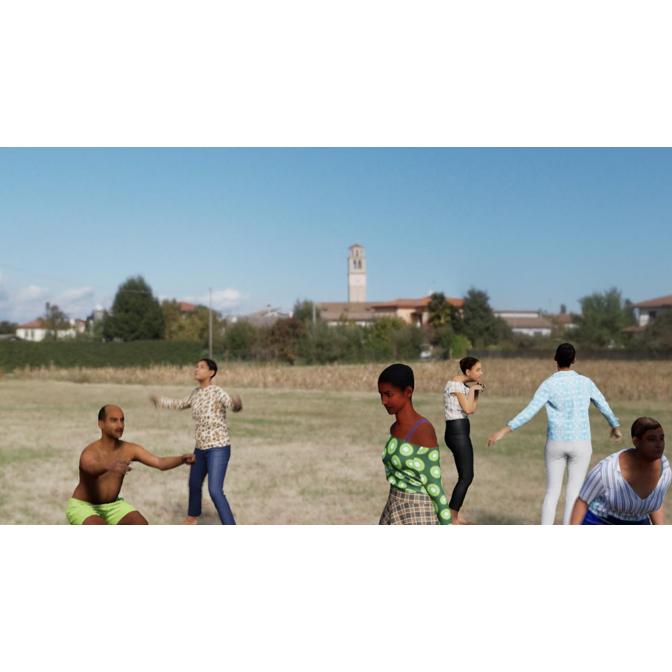} &
\includegraphics[width=0.15\linewidth]{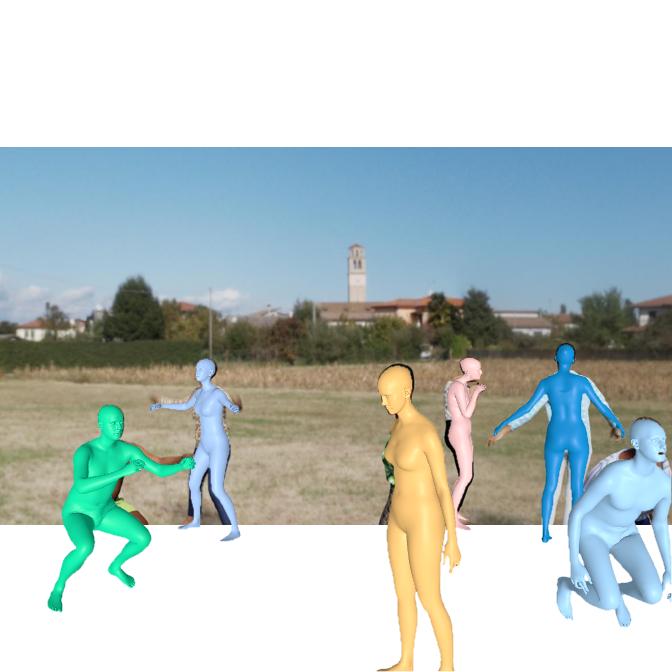} &
\includegraphics[width=0.15\linewidth]{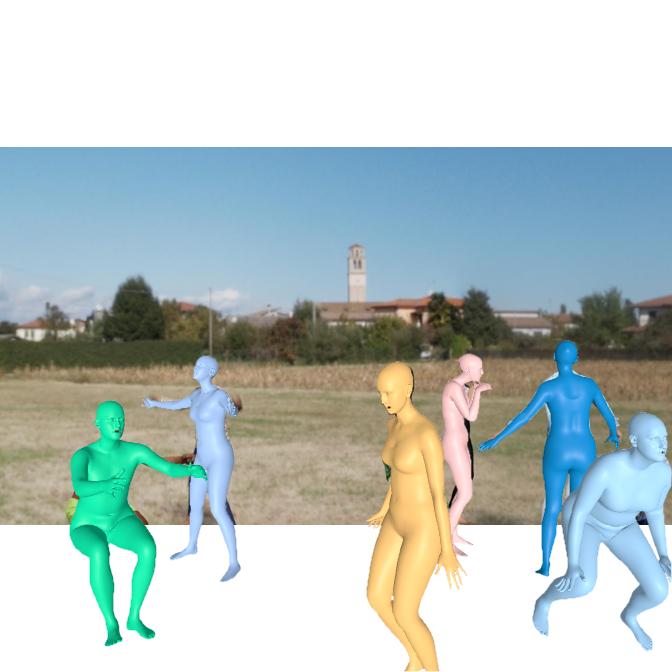} &

\includegraphics[width=0.15\linewidth]{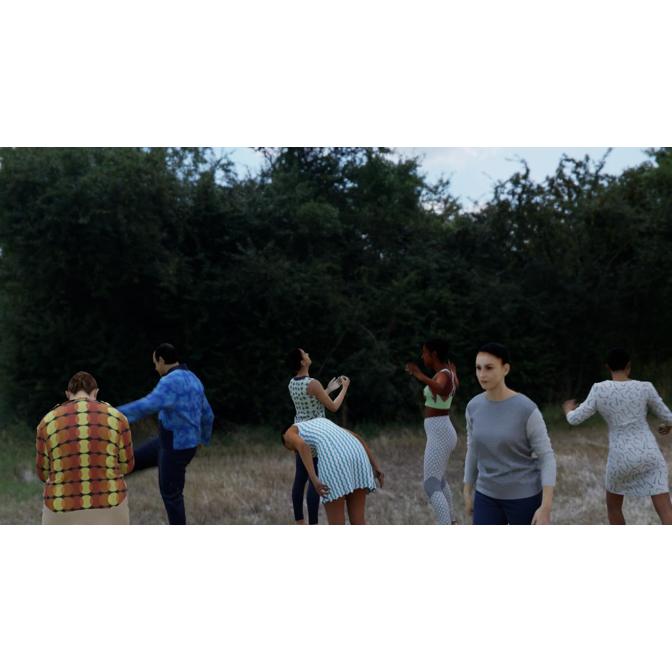} &
\includegraphics[width=0.15\linewidth]{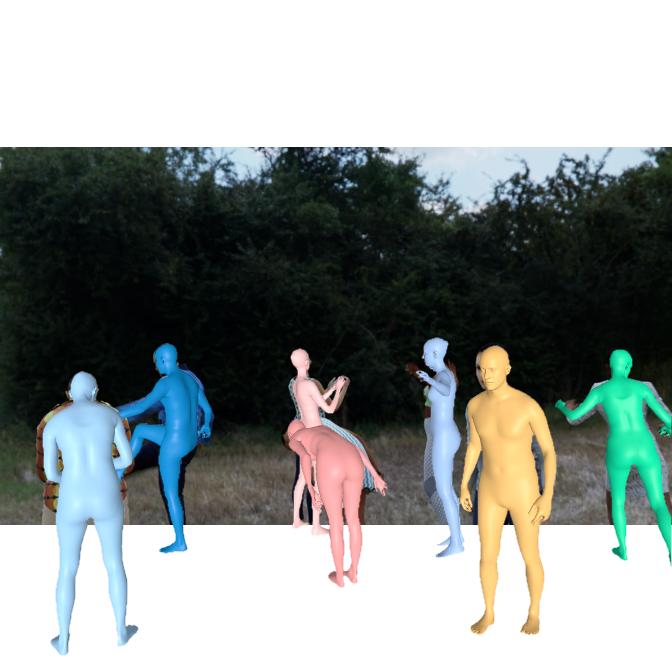} &
\includegraphics[width=0.15\linewidth]{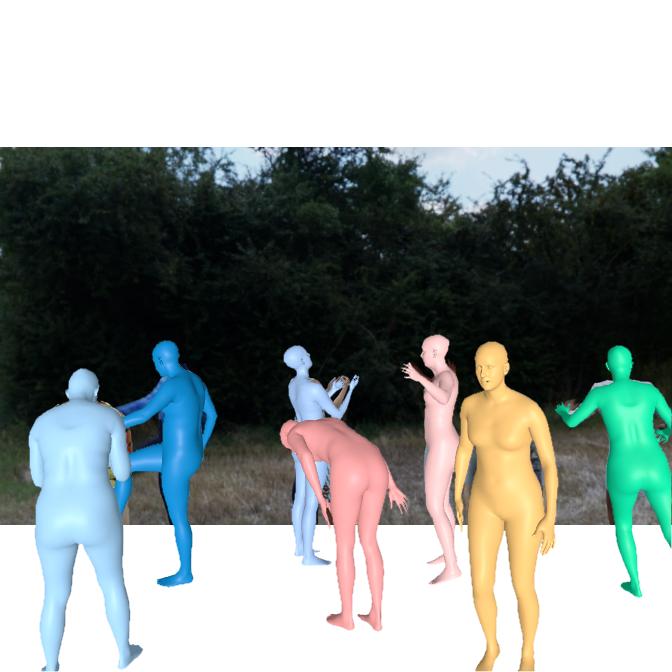} \\

\end{tabular}
\end{center}
\vspace{-0.3cm}
\caption{\label{fig:hmr1}\textbf{Qualitative Human Mesh Recovery results.} Qualitative comparison of outputs between teacher and student. Images sampled in the validation set and sorted by alphabetical order. The two models produce outputs of comparable visual quality.}
\end{figure*}

\begin{figure*}
\begin{center}
\begin{tabular}{ccc|ccc}

Input image & Student output & Teacher output &  Input image & Student output & Teacher output  \\

\includegraphics[width=0.15\linewidth]{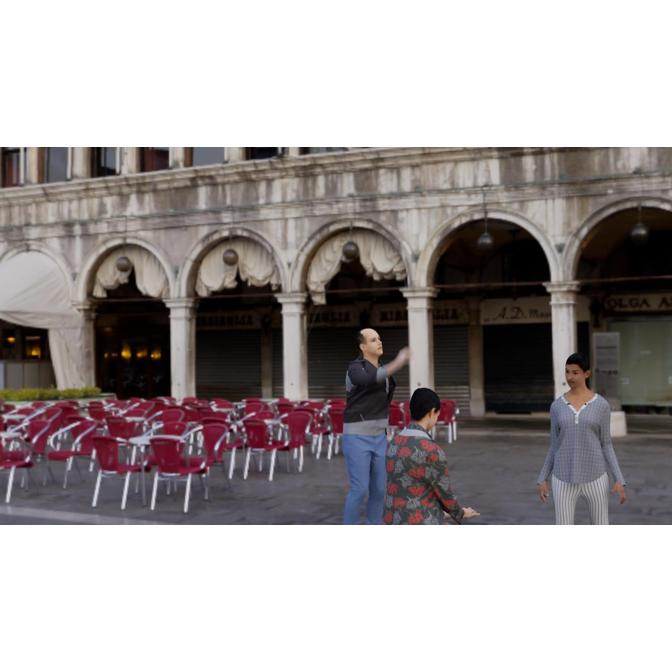} &
\includegraphics[width=0.15\linewidth]{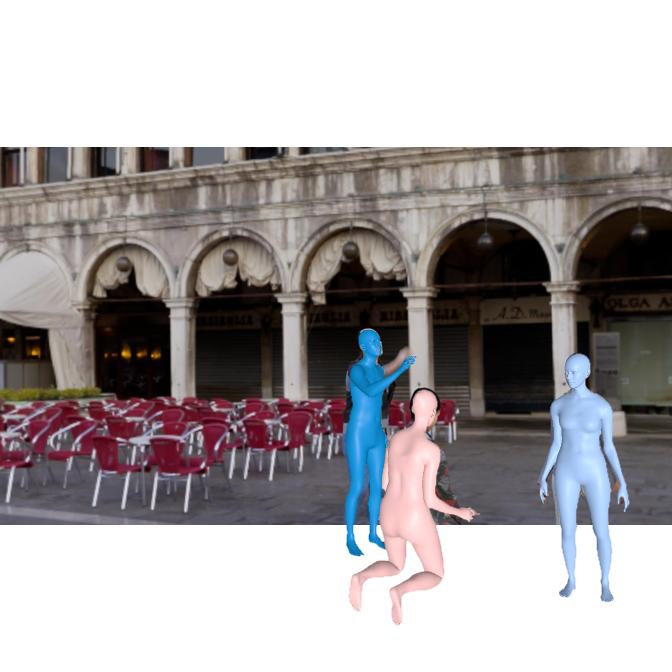} &
\includegraphics[width=0.15\linewidth]{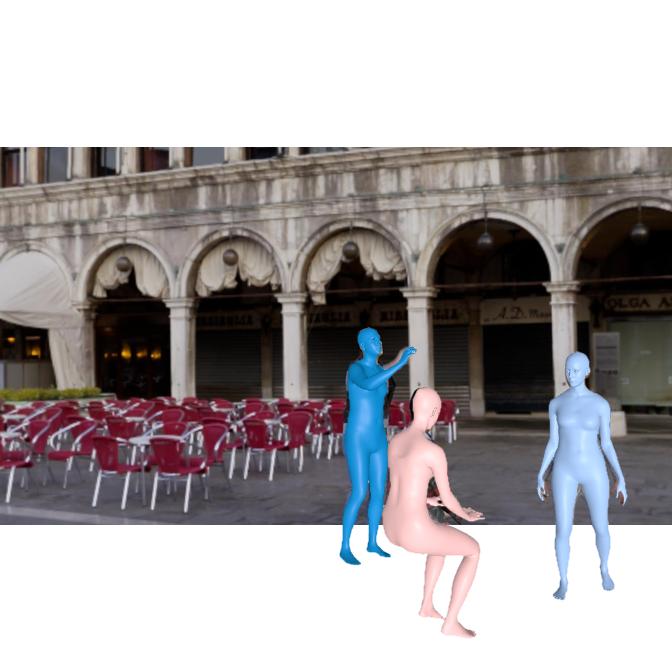} &

\includegraphics[width=0.15\linewidth]{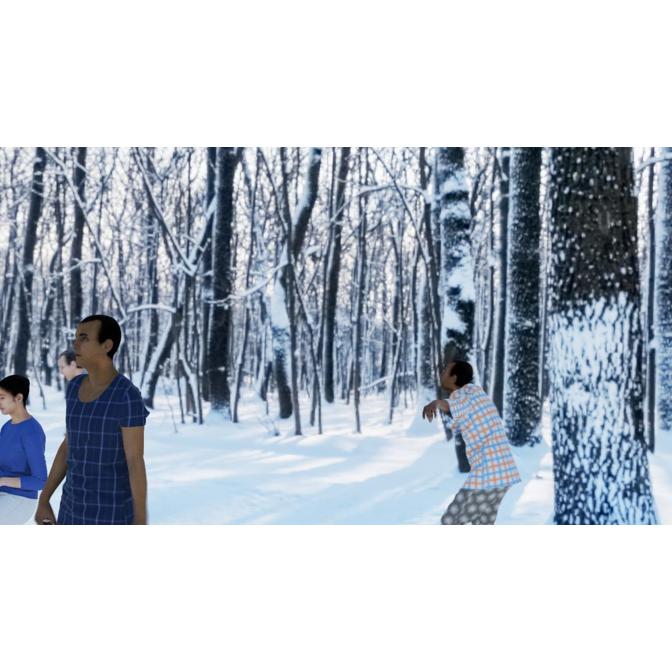} &
\includegraphics[width=0.15\linewidth]{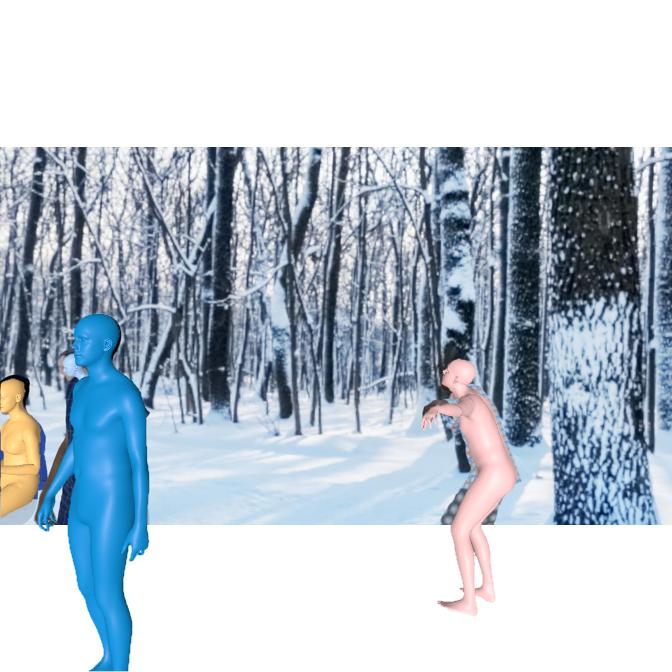} &
\includegraphics[width=0.15\linewidth]{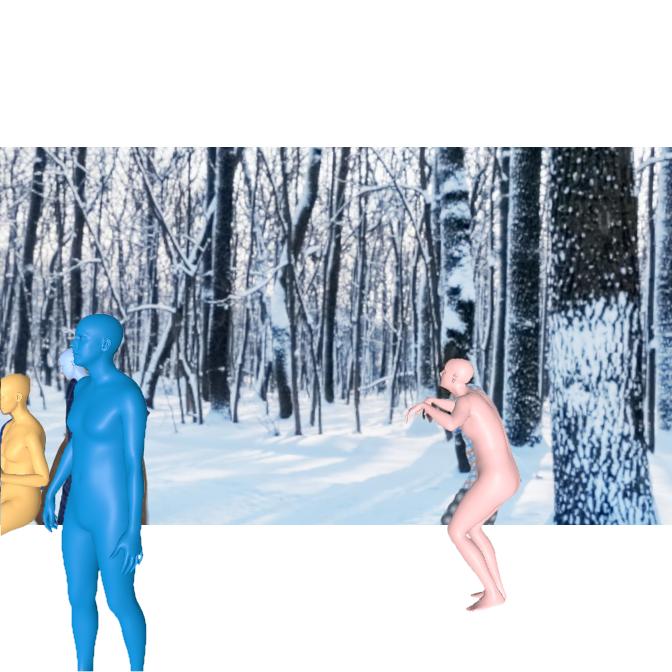} \\

\includegraphics[width=0.15\linewidth]{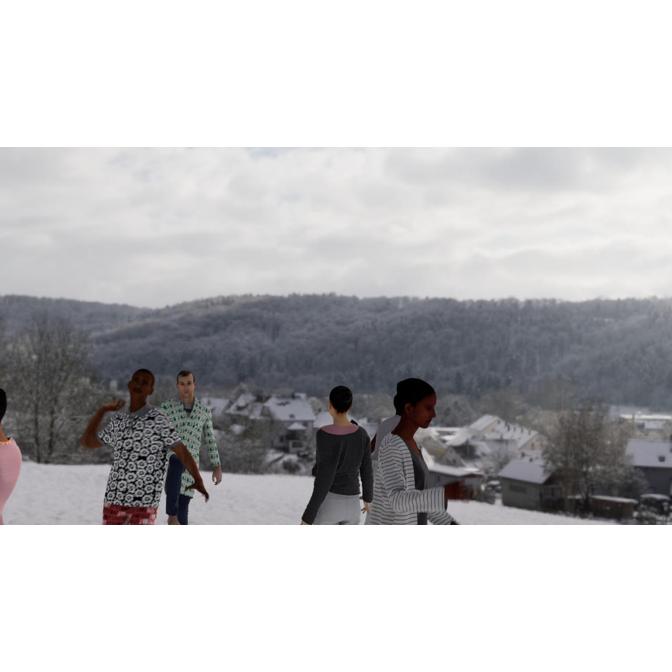} &
\includegraphics[width=0.15\linewidth]{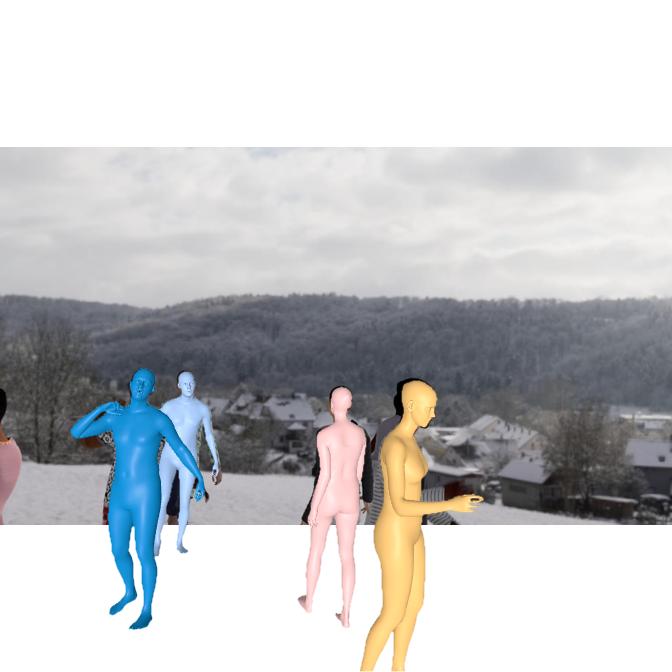} &
\includegraphics[width=0.15\linewidth]{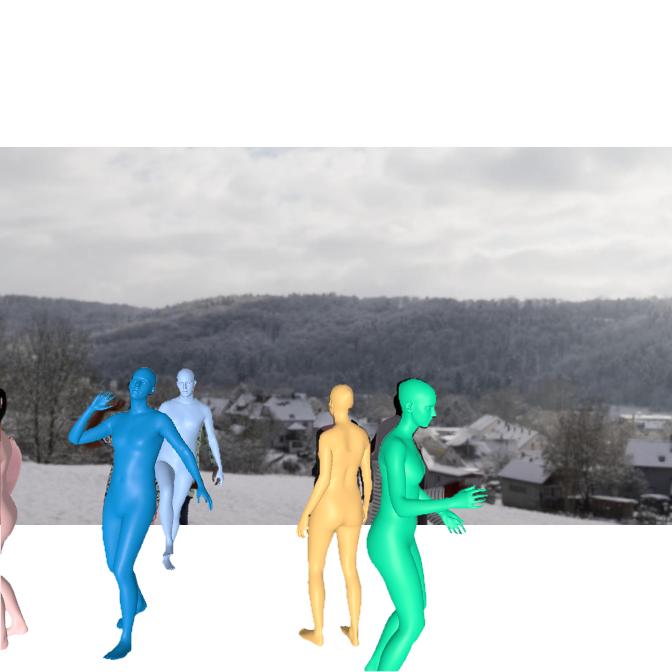} &

\includegraphics[width=0.15\linewidth]{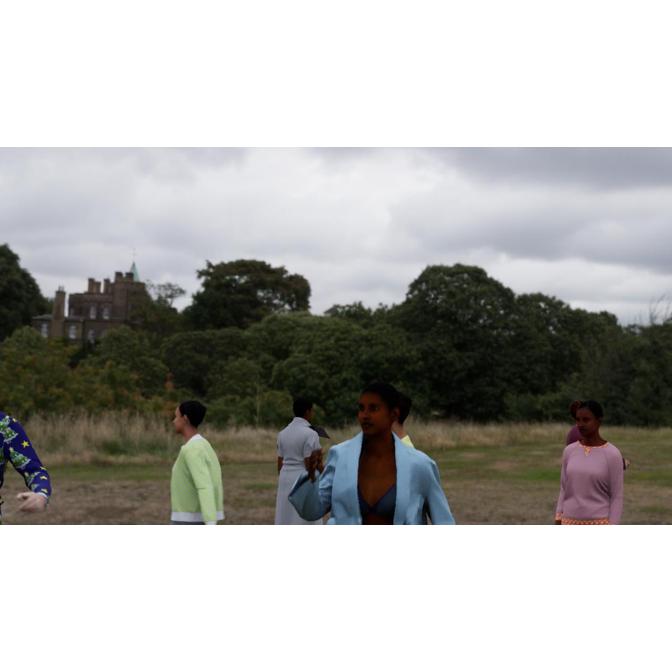} &
\includegraphics[width=0.15\linewidth]{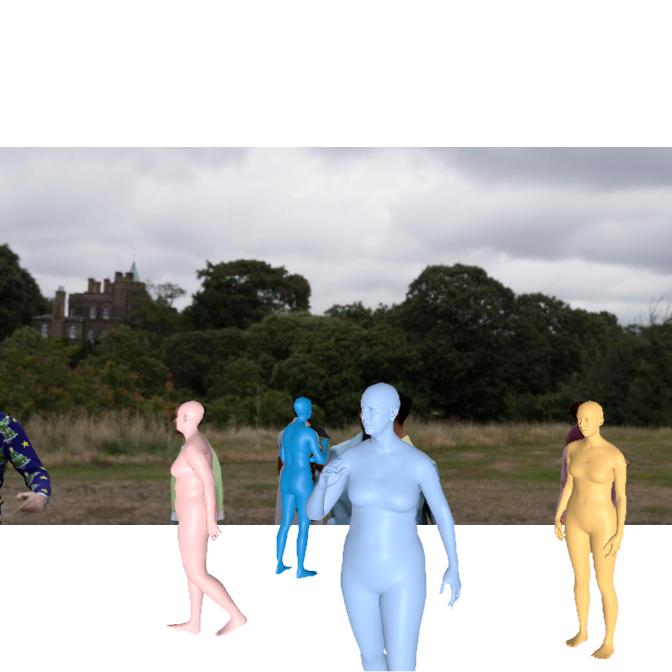} &
\includegraphics[width=0.15\linewidth]{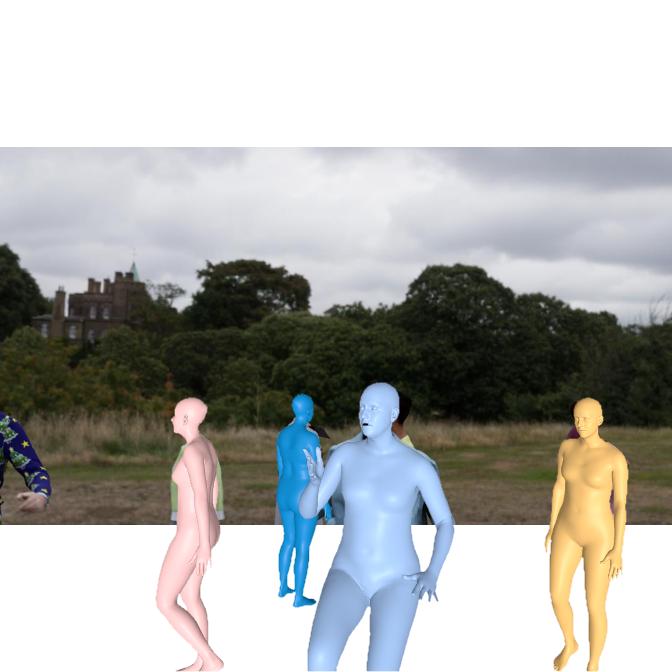} \\

\includegraphics[width=0.15\linewidth]{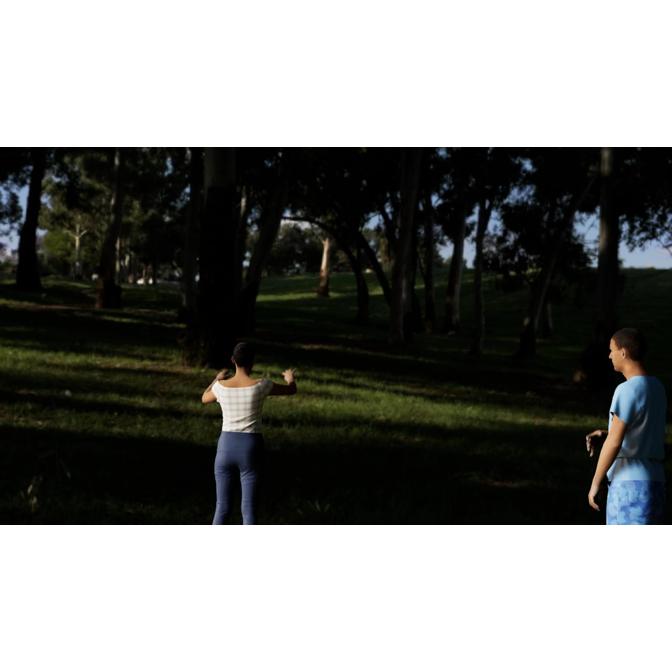} &
\includegraphics[width=0.15\linewidth]{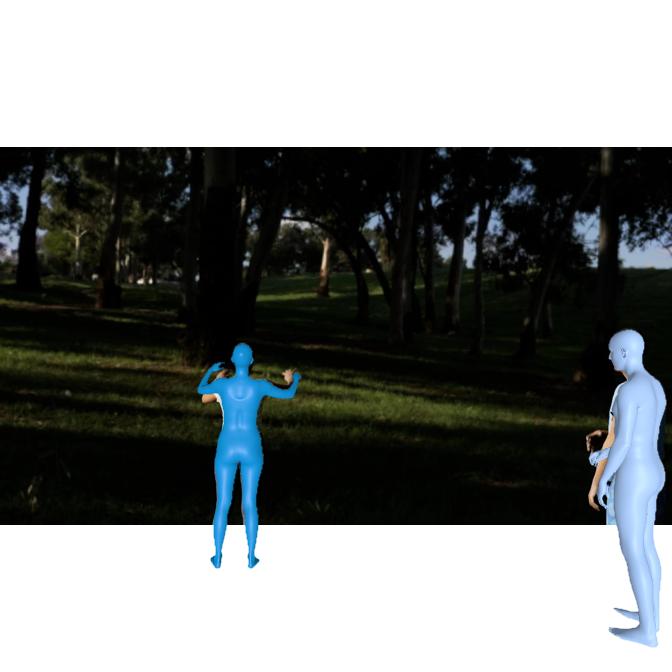} &
\includegraphics[width=0.15\linewidth]{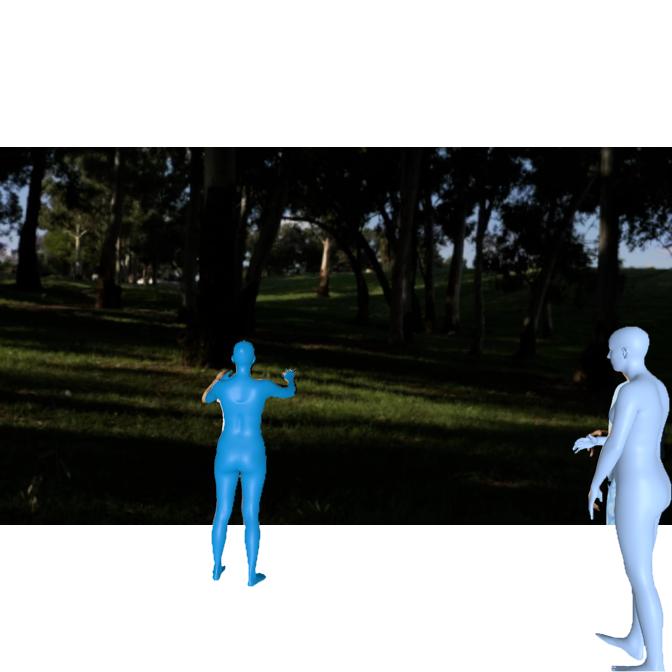} &

\includegraphics[width=0.15\linewidth]{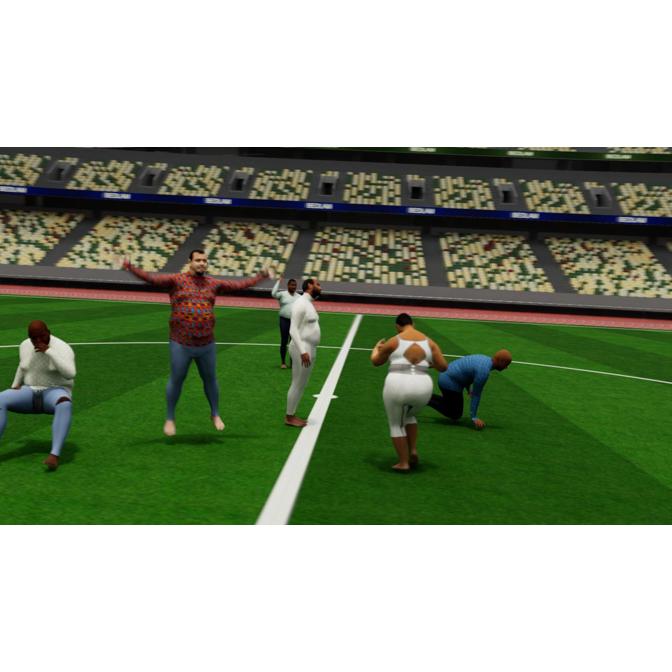} &
\includegraphics[width=0.15\linewidth]{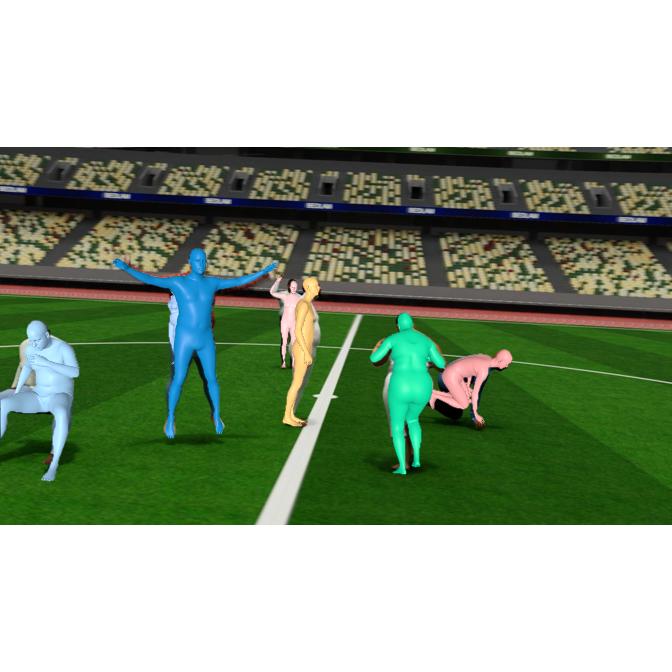} &
\includegraphics[width=0.15\linewidth]{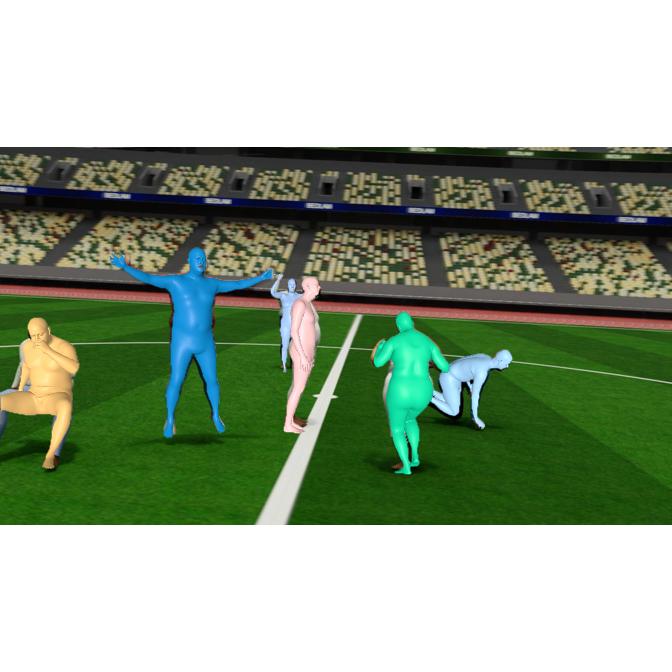} \\

\includegraphics[width=0.15\linewidth]{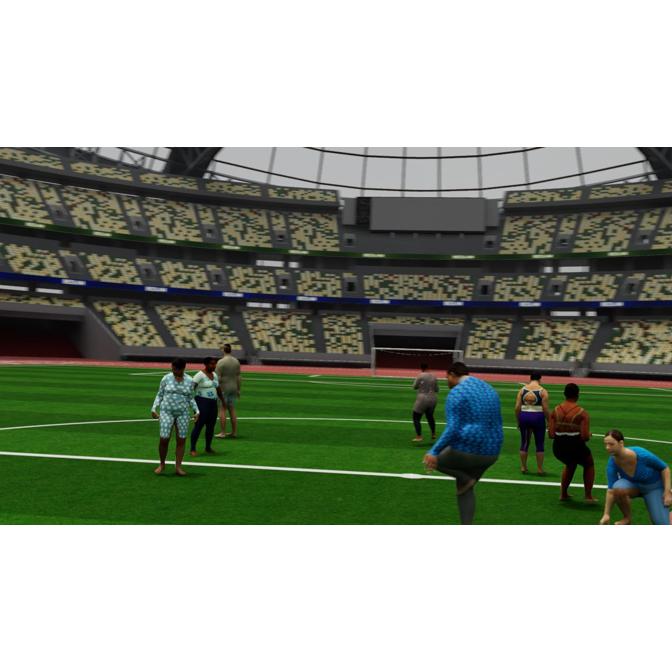} &
\includegraphics[width=0.15\linewidth]{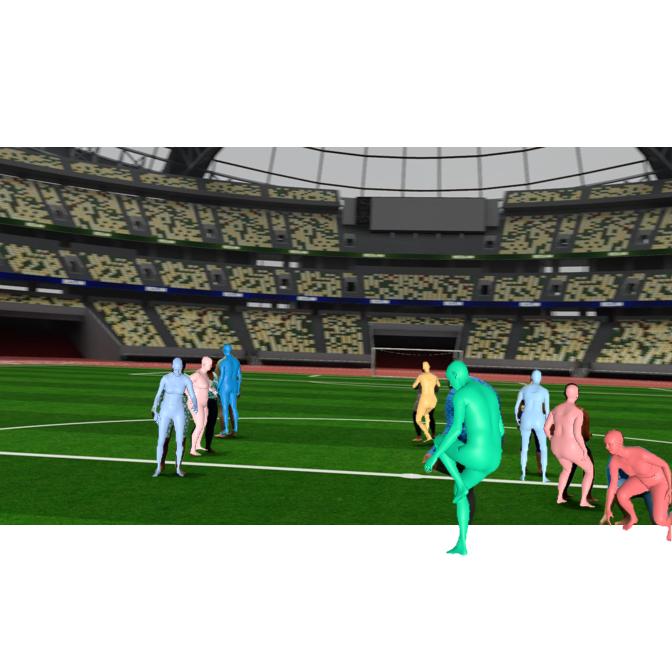} &
\includegraphics[width=0.15\linewidth]{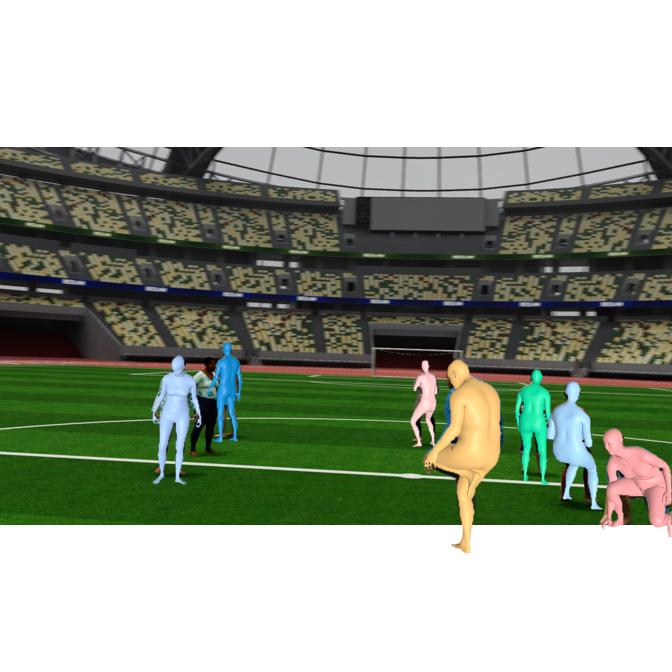} &

\includegraphics[width=0.15\linewidth]{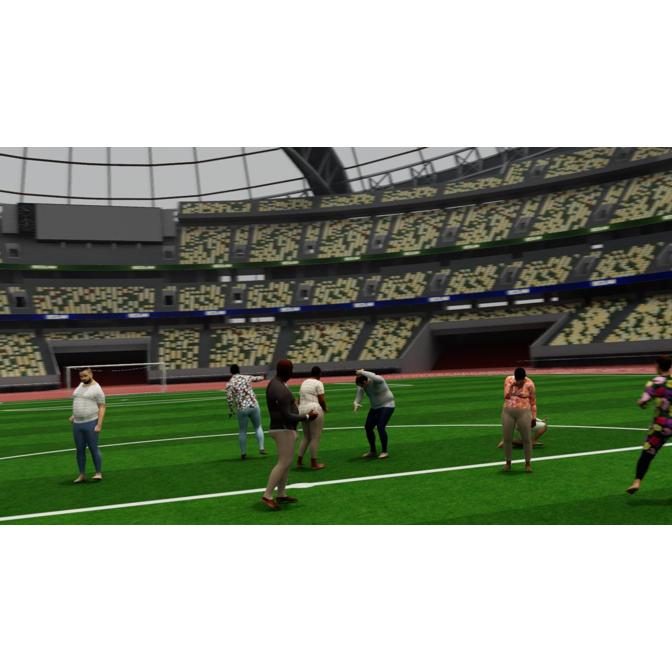} &
\includegraphics[width=0.15\linewidth]{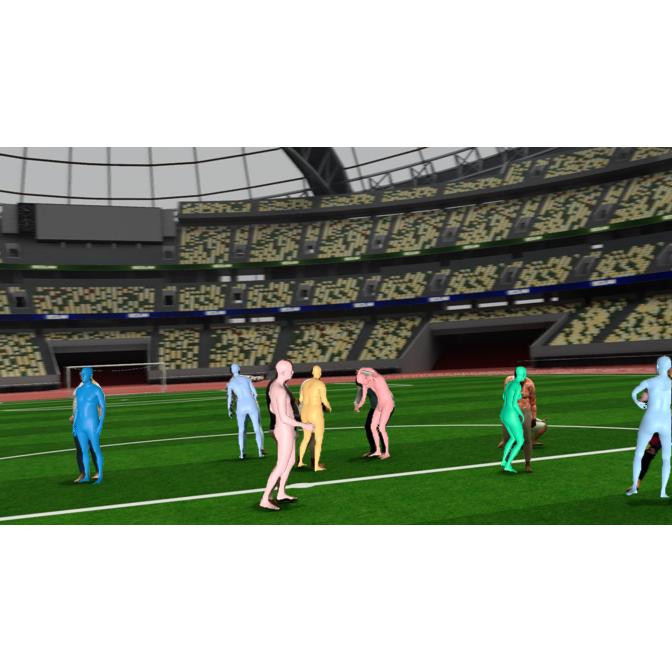} &
\includegraphics[width=0.15\linewidth]{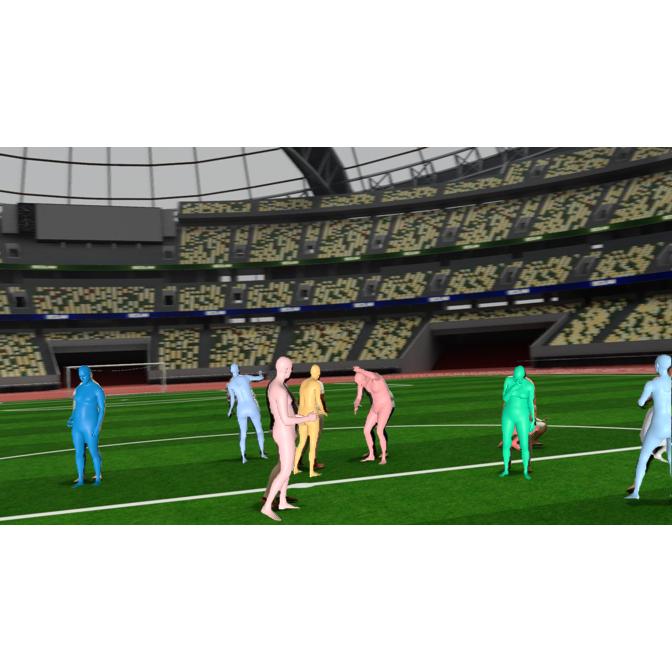} \\

\includegraphics[width=0.15\linewidth]{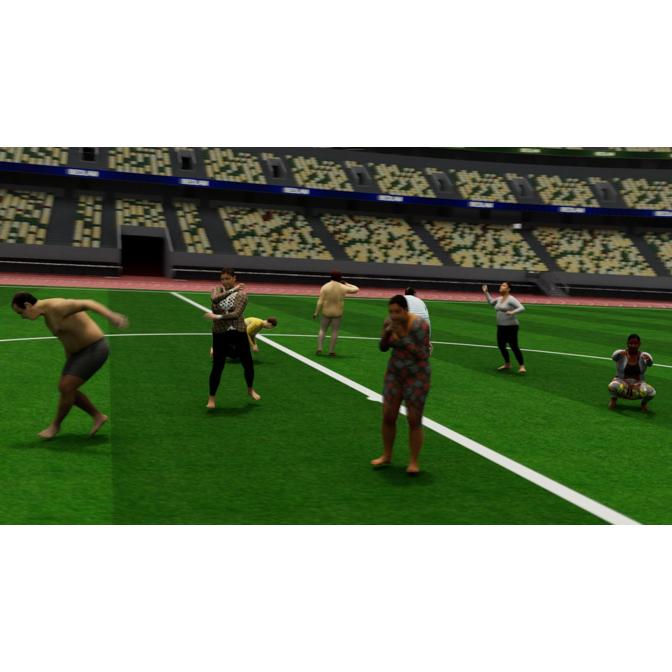} &
\includegraphics[width=0.15\linewidth]{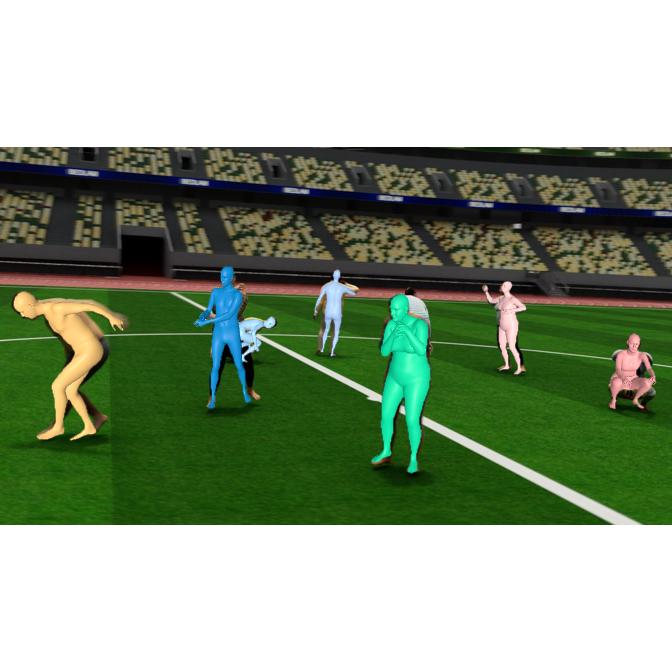} &
\includegraphics[width=0.15\linewidth]{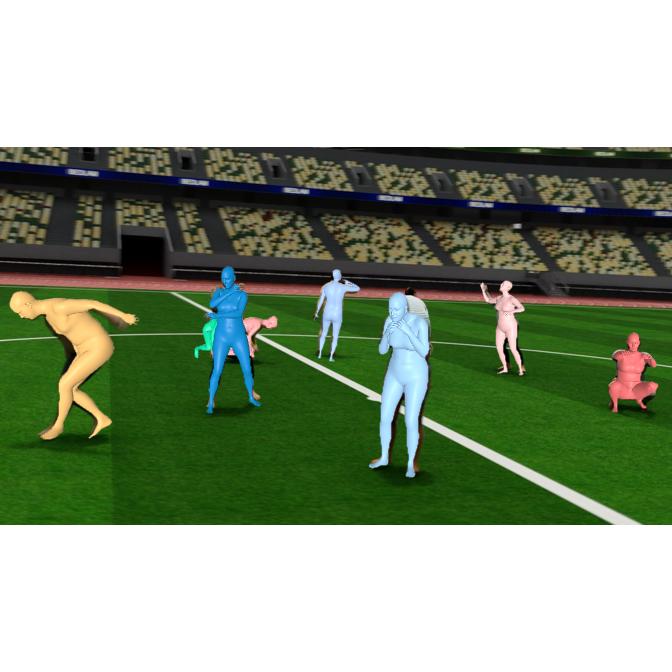} &

\includegraphics[width=0.15\linewidth]{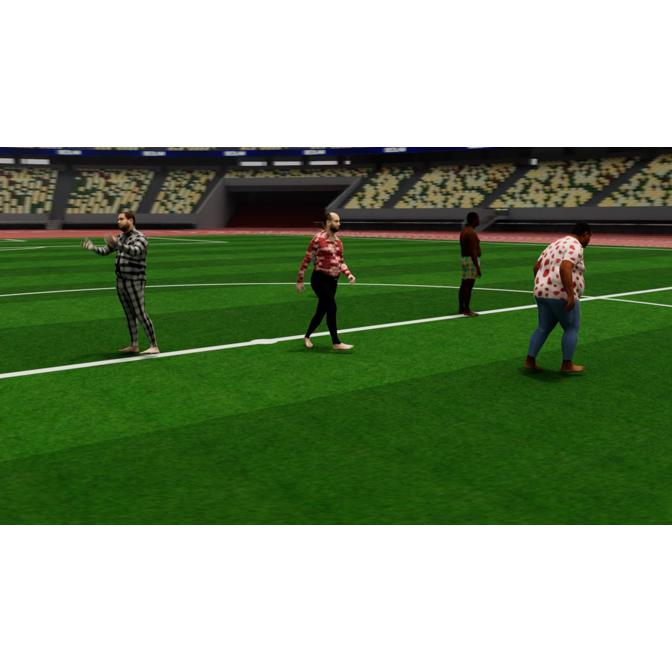} &
\includegraphics[width=0.15\linewidth]{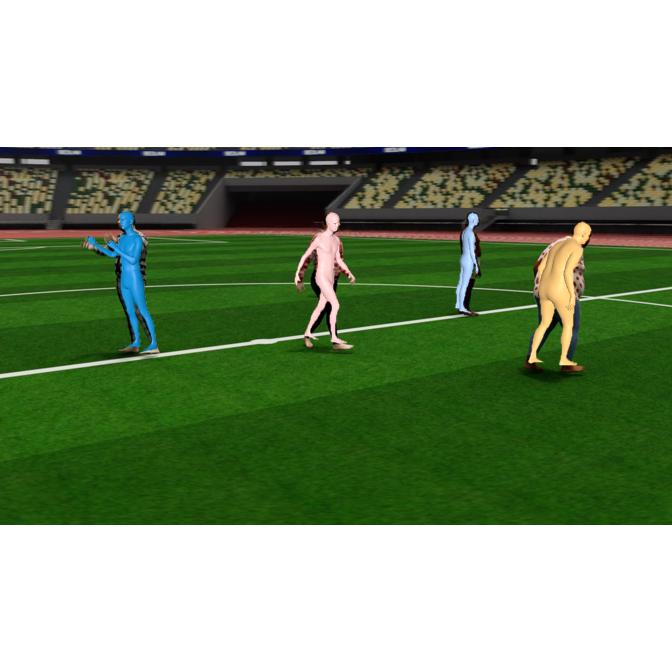} &
\includegraphics[width=0.15\linewidth]{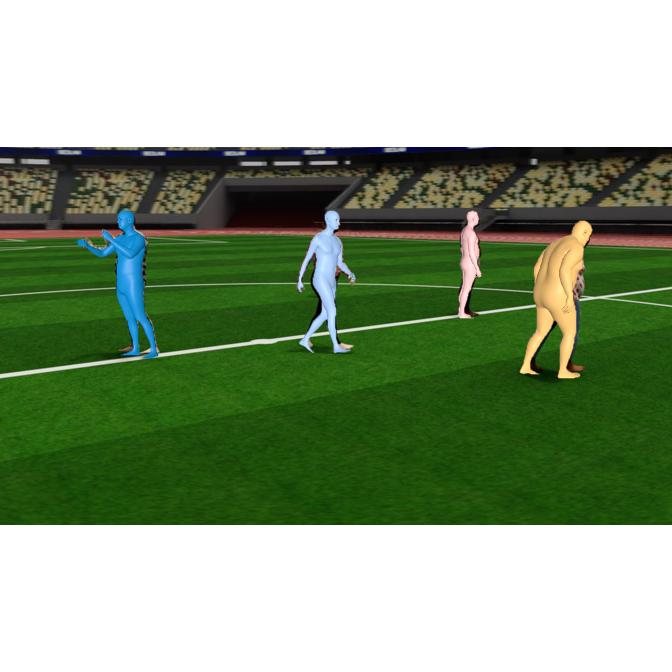} \\

\includegraphics[width=0.15\linewidth]{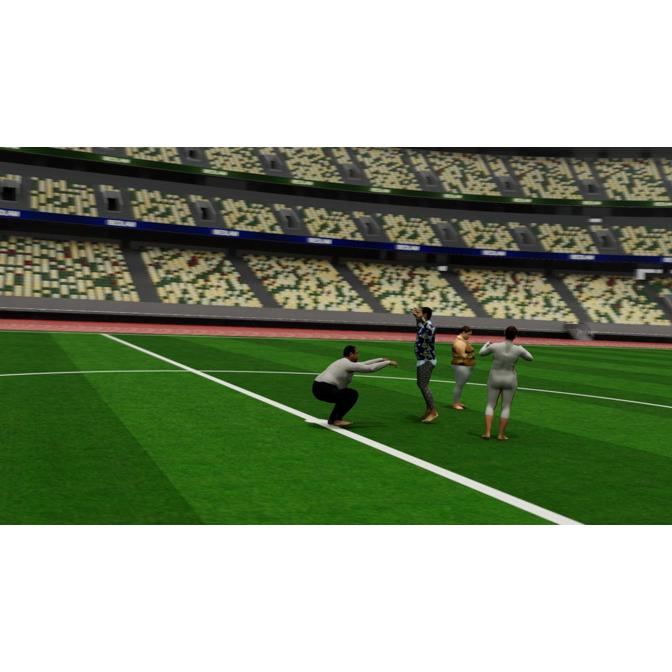} &
\includegraphics[width=0.15\linewidth]{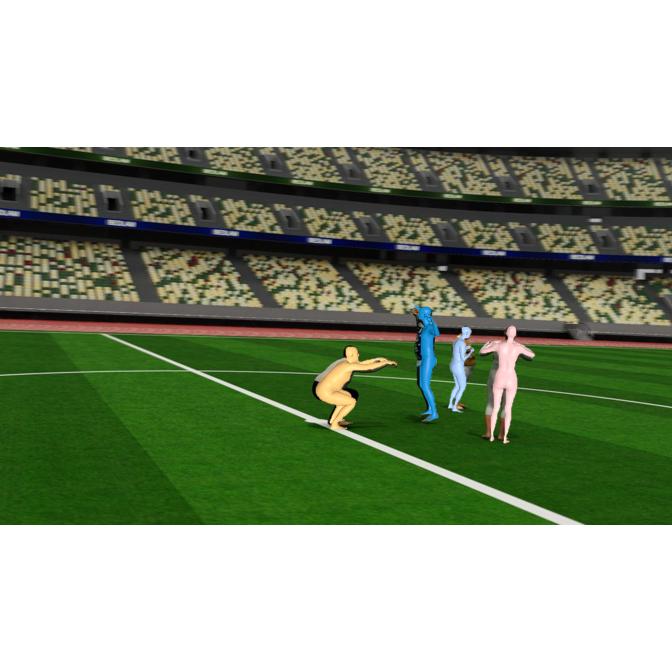} &
\includegraphics[width=0.15\linewidth]{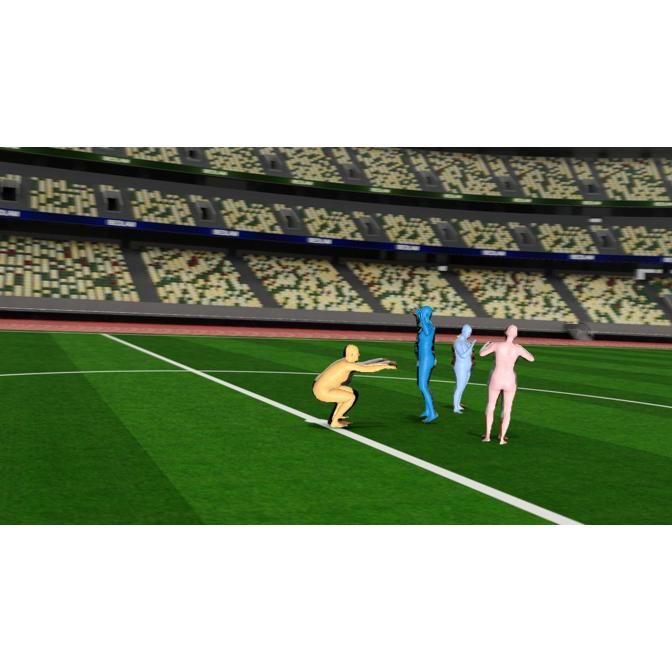} &

\includegraphics[width=0.15\linewidth]{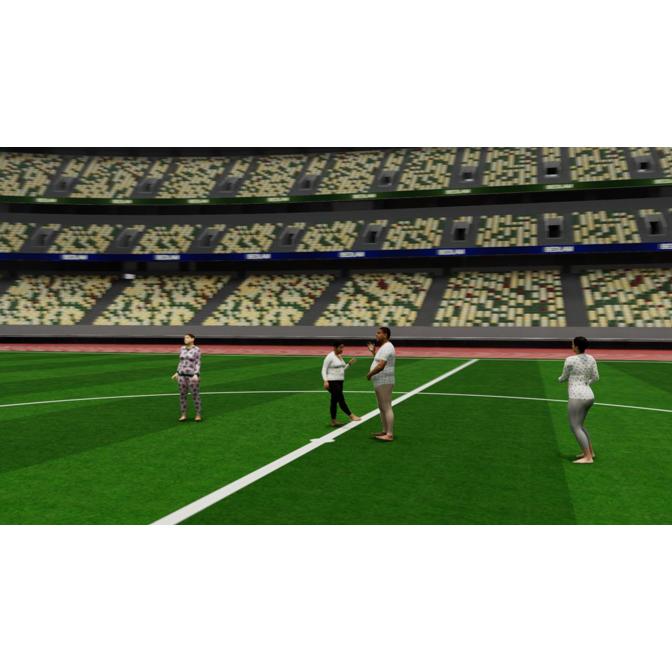} &
\includegraphics[width=0.15\linewidth]{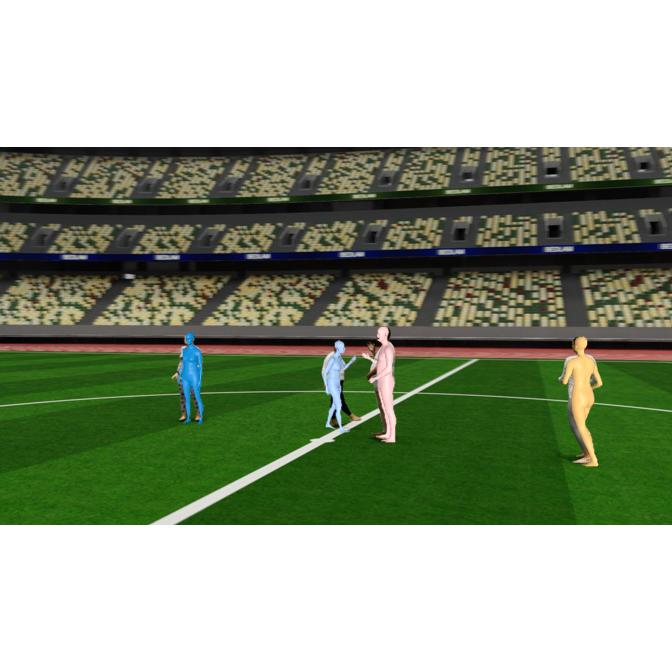} &
\includegraphics[width=0.15\linewidth]{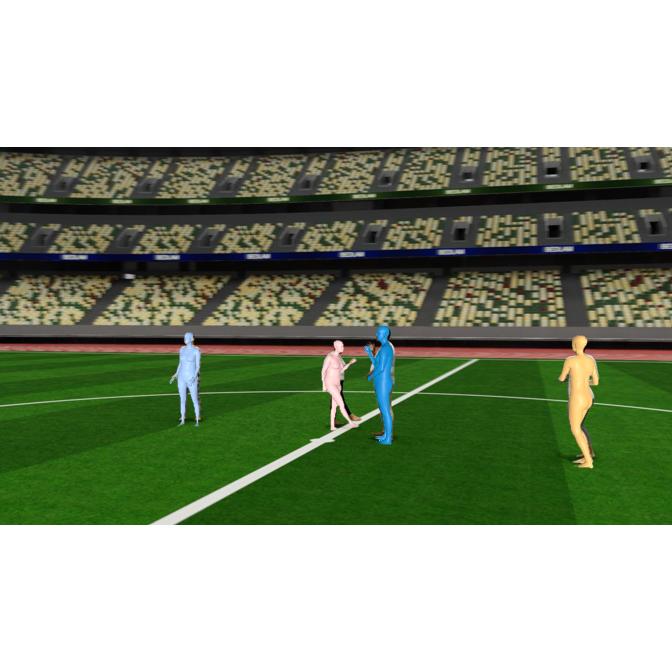} \\

\end{tabular}
\end{center}
\vspace{-0.3cm}
\caption{\label{fig:hmr2}\textbf{Qualitative Human Mesh Recovery results (continued).}  Qualitative comparison of outputs between teacher and student. Images sampled in the validation set and sorted by alphabetical order. The two models produce outputs of comparable visual quality.}
\end{figure*}

%% file: tex/float/supplementary/fig_dataset_samples.tex
\begin{figure*}[t]
\centering
\adjustbox{max width=\textwidth}{
\begin{tabular}{c}
    ImageNet-19K \\
    \includegraphics[height=2cm]{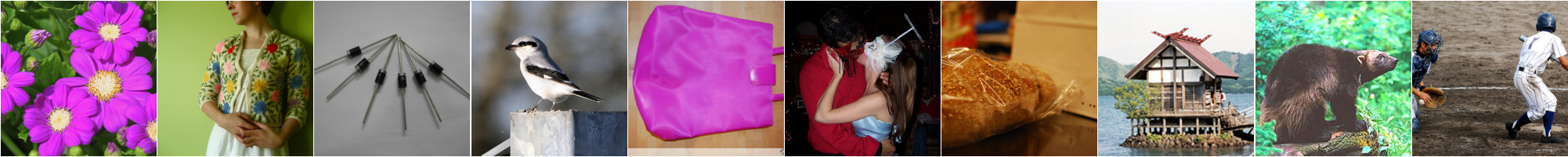} \\
    Mapillary \\
    \includegraphics[height=2cm]{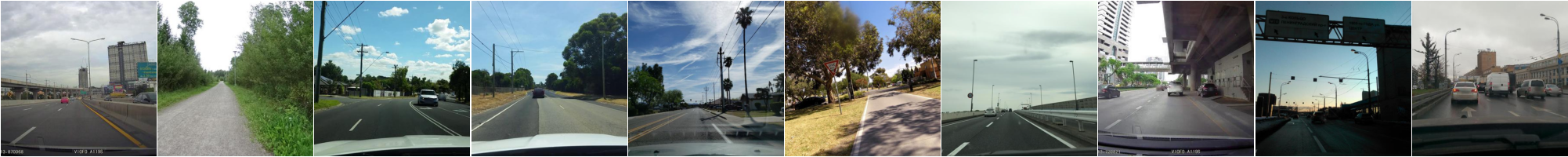} \\
    Google Landmarks v2 \\
    \includegraphics[height=2cm]{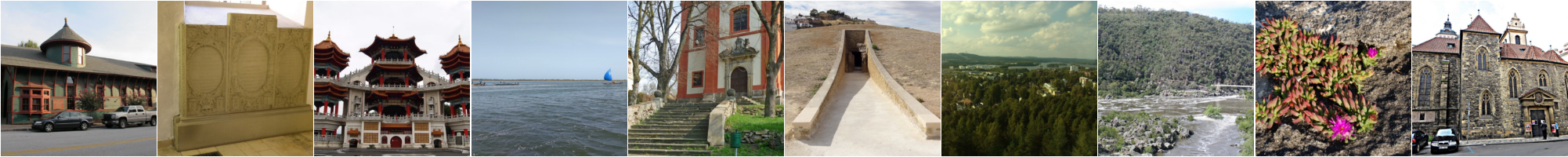} \\
    Habitat \\
    \includegraphics[height=2cm]{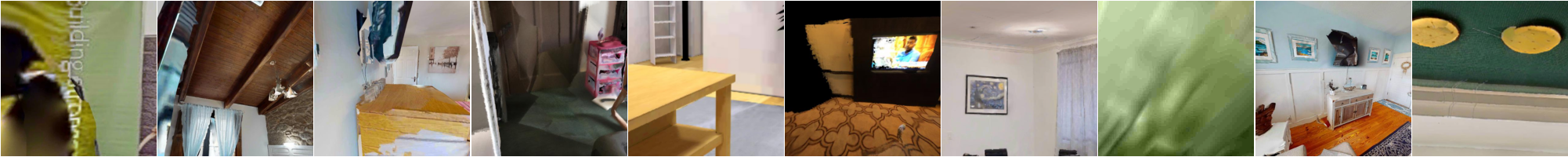} \\
    ARKitScenes \\
    \includegraphics[height=2cm]{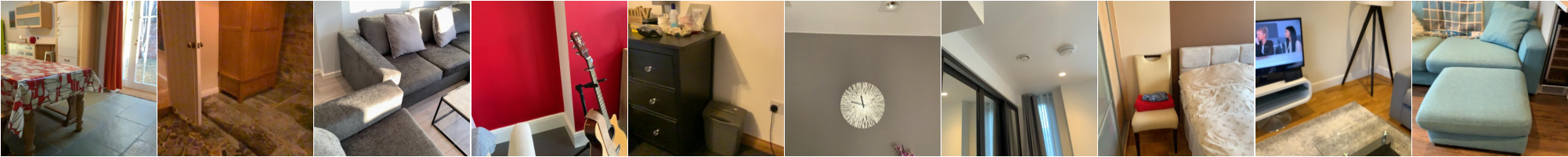} \\
    Blended MVS \\
    \includegraphics[height=2cm]{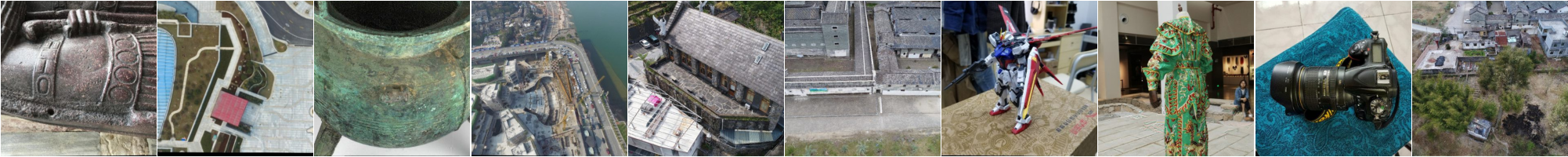} \\
    MegaDepth \\
    \includegraphics[height=2cm]{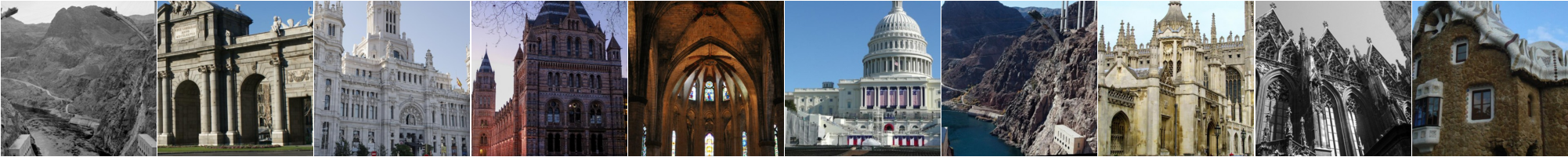} \\
    ScanNet++ \\
    \includegraphics[height=2cm]{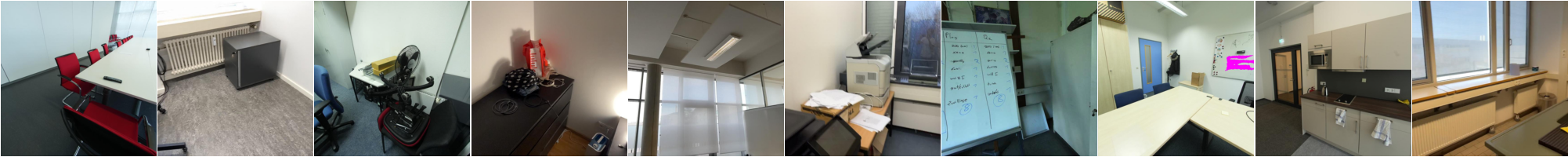} \\
    CO3D-v2 \\
    \includegraphics[height=2cm]{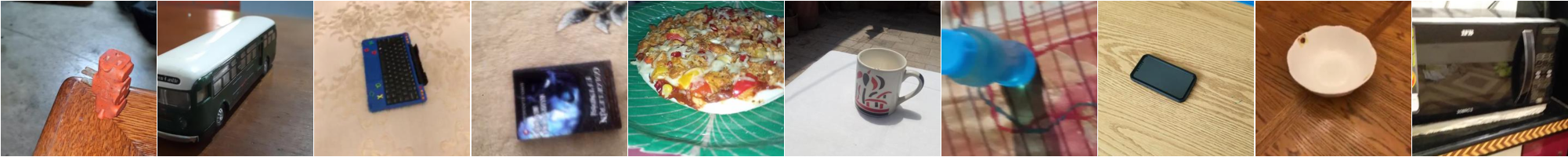} \\
    Map-free \\
    \includegraphics[height=2cm]{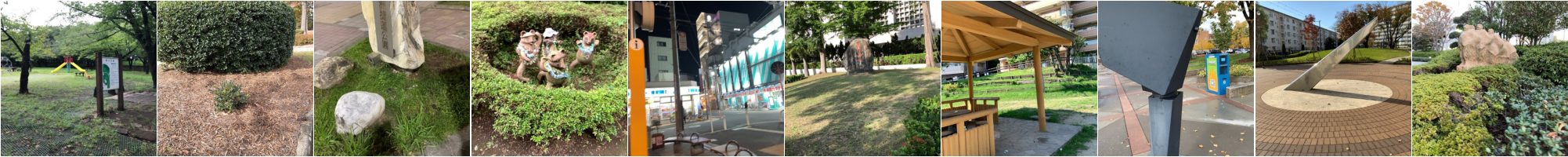} \\
\end{tabular}
}
\vspace{-0.3cm}
\caption{
    {\bf Visualization of random samples from datasets.}
    We visualize 10 randomly sampled images from each dataset listed in~\Cref{tab:datasets}.
    See~\Cref{fig:datasets_2} for the visualization of the remaining datasets.
    }
\label{fig:datasets_1}
\end{figure*}

\begin{figure*}[t]
\centering
\adjustbox{max width=\textwidth}{
\begin{tabular}{c}
    WildRgbd \\
    \includegraphics[height=2cm]{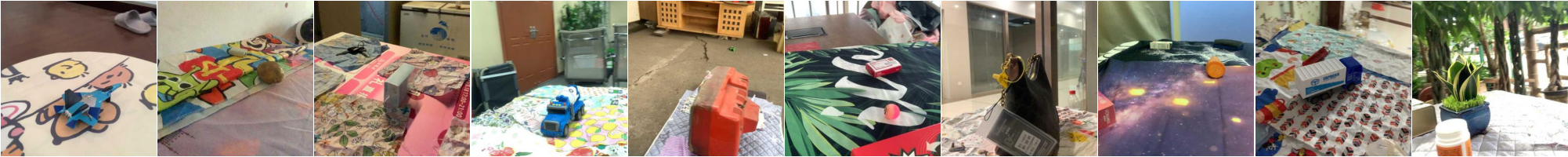} \\
    VirtualKitti \\
    \includegraphics[height=2cm]{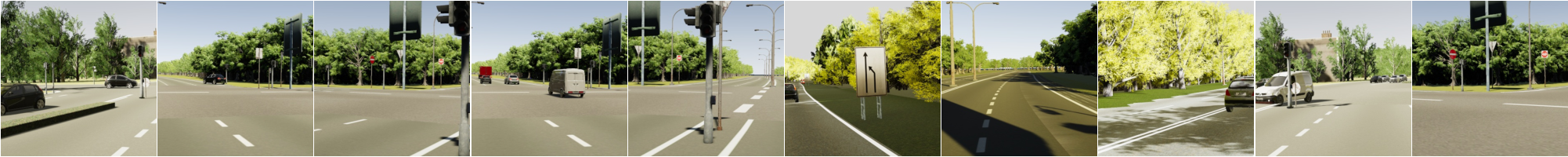} \\
    Unreal4K \\
    \includegraphics[height=2cm]{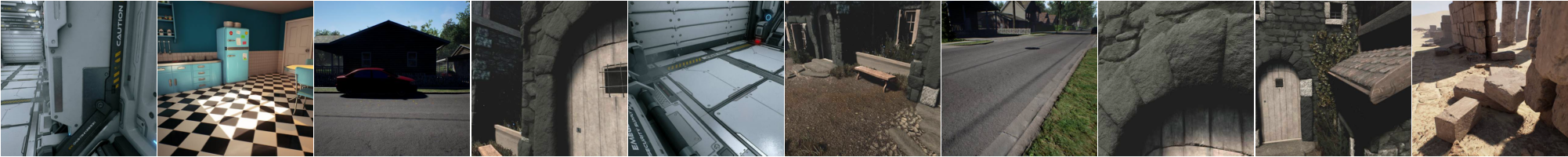} \\
    TartanAir \\
    \includegraphics[height=2cm]{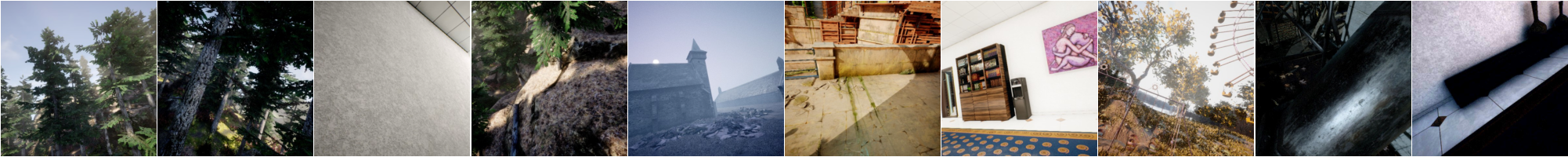} \\
    DL3DV \\
    \includegraphics[height=2cm]{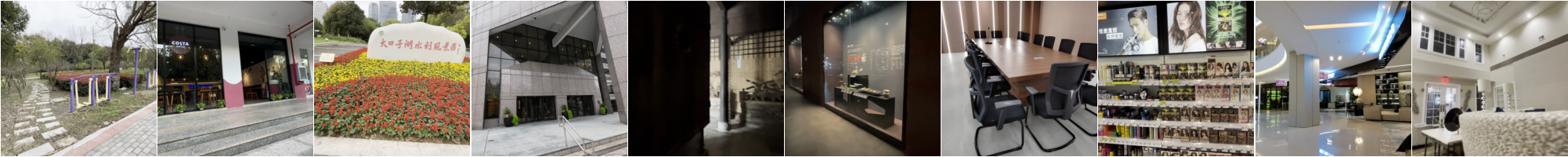} \\
    BEDLAM \\
    \includegraphics[height=2cm]{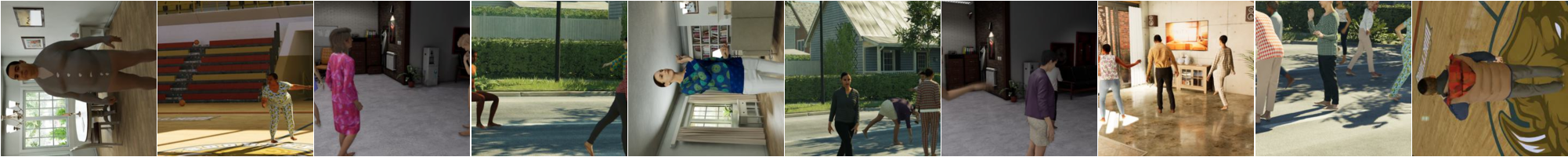} \\
    AGORA \\
    \includegraphics[height=2cm]{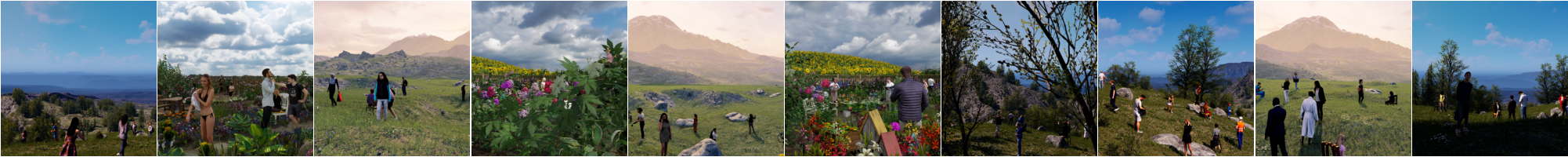} \\
    CUFFS \\
    \includegraphics[height=2cm]{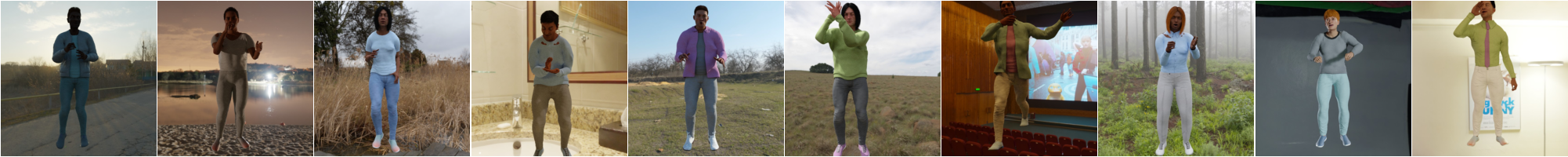} \\
    UBody \\
    \includegraphics[height=2cm]{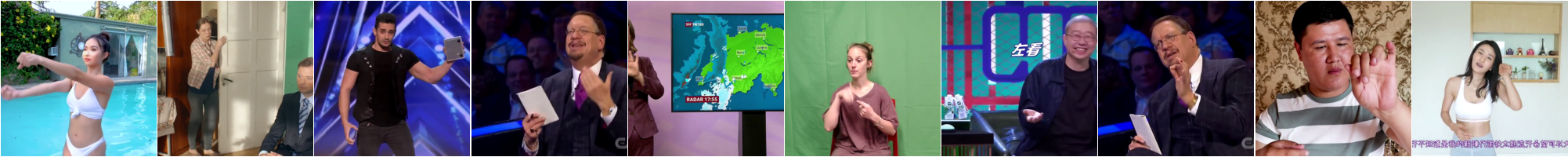} \\
\end{tabular}
}
\vspace{-0.3cm}
\caption{    {\bf Visualization of random samples from datasets (continuation of~\Cref{fig:datasets_1}).}
    We visualize 10 randomly sampled images from each dataset listed in~\Cref{tab:datasets}.
    See~\Cref{fig:datasets_2} for the visualization of the remaining datasets.}
\label{fig:datasets_2}
\end{figure*}